\documentclass[runningheads]{llncs}
\usepackage{graphicx}
\usepackage{comment}
\usepackage{amsmath,amssymb} 
\usepackage{color}

\usepackage[pagebackref=true,breaklinks=true,letterpaper=true,colorlinks,bookmarks=false]{hyperref}

\usepackage{enumitem}
\usepackage{mmstyle}
\usepackage{makecell}
\usepackage{subfig}
\usepackage{dsfont}
\usepackage{pifont}
\usepackage{multirow}
\usepackage{capt-of}
\usepackage{listing}

\usepackage{algorithm, algpseudocode} 
\algnewcommand\algorithmicinput{\textbf{INPUT:}}
\algnewcommand\INPUT{\item[\algorithmicinput]}
\algnewcommand\algorithmicoutput{\textbf{OUTPUT:}}
\algnewcommand\OUTPUT{\item[\algorithmicoutput]}

\newcommand{\etc}{\emph{etc}.}
\newcommand{\cmark}{\ding{51}}%

\graphicspath{{figures/}}

\begin{document}
\pagestyle{headings}
\mainmatter
\def\ECCVSubNumber{2710}  

\title{MovieNet: A Holistic Dataset for Movie Understanding} 

\titlerunning{MovieNet: A Holistic Dataset for Movie Understanding}
%
\author{Qingqiu Huang\thanks{Equal contribution} \and
Yu Xiong$^\star$ \and
Anyi Rao \and
Jiaze Wang \and
Dahua Lin}
\authorrunning{Q. Huang, Y. Xiong, A. Rao, J. Wang and D. Lin}
%
\institute{CUHK-SenseTime Joint Lab, The Chinese University of Hong Kong\\
\email{\{hq016, xy017, ra018, dhlin\}@ie.cuhk.edu.hk}\\ \email{jzwang@link.cuhk.edu.hk}}
\maketitle


\begin{abstract}

Recent years have seen remarkable advances in visual understanding.
However, how to understand a story-based long video with artistic styles, \eg movie, remains challenging.
In this paper, we introduce MovieNet -- a holistic dataset for movie understanding.
MovieNet contains $1,100$ movies with a large amount of multi-modal data, \eg trailers, photos, plot descriptions, \etc.
Besides, different aspects of manual annotations are provided in MovieNet, including $1.1M$ characters with bounding boxes and identities, $42K$ scene boundaries, $2.5K$ aligned description sentences, $65K$ tags of place and action, and $92K$ tags of cinematic style.
To the best of our knowledge, MovieNet is the largest dataset with richest annotations for comprehensive movie understanding.
Based on MovieNet, we set up several benchmarks for movie understanding from different angles.
Extensive experiments are executed on these benchmarks to show the immeasurable value of MovieNet and the gap of current approaches towards comprehensive movie understanding. We believe that such a holistic dataset would promote the researches on story-based long video understanding and beyond.
%
%
MovieNet will be published in compliance with regulations at \url{https://movienet.github.io}.
\end{abstract}

\section{Introduction}
\label{sec:introduction}

\begin{figure}[t]
	\centering
	\includegraphics[width=0.93\linewidth]{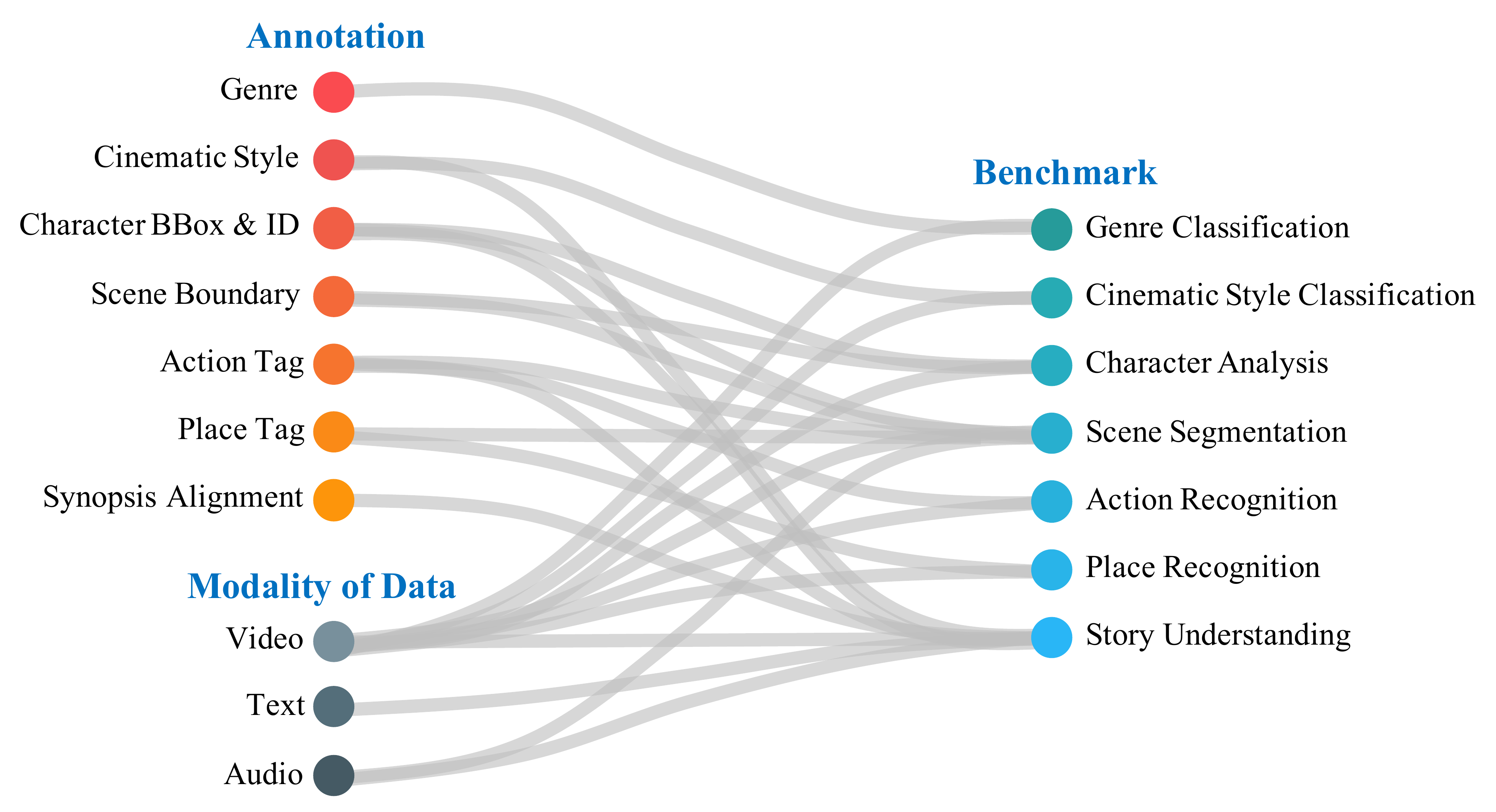}
	\caption{\small
		The data, annotation, benchmark and their relations in MovieNet, which together build a holistic dataset for comprehensive movie understanding.
	}
	\label{fig:teaser}
\end{figure}

\begin{figure}[!ht]
	\centering
	\includegraphics[width=\linewidth]{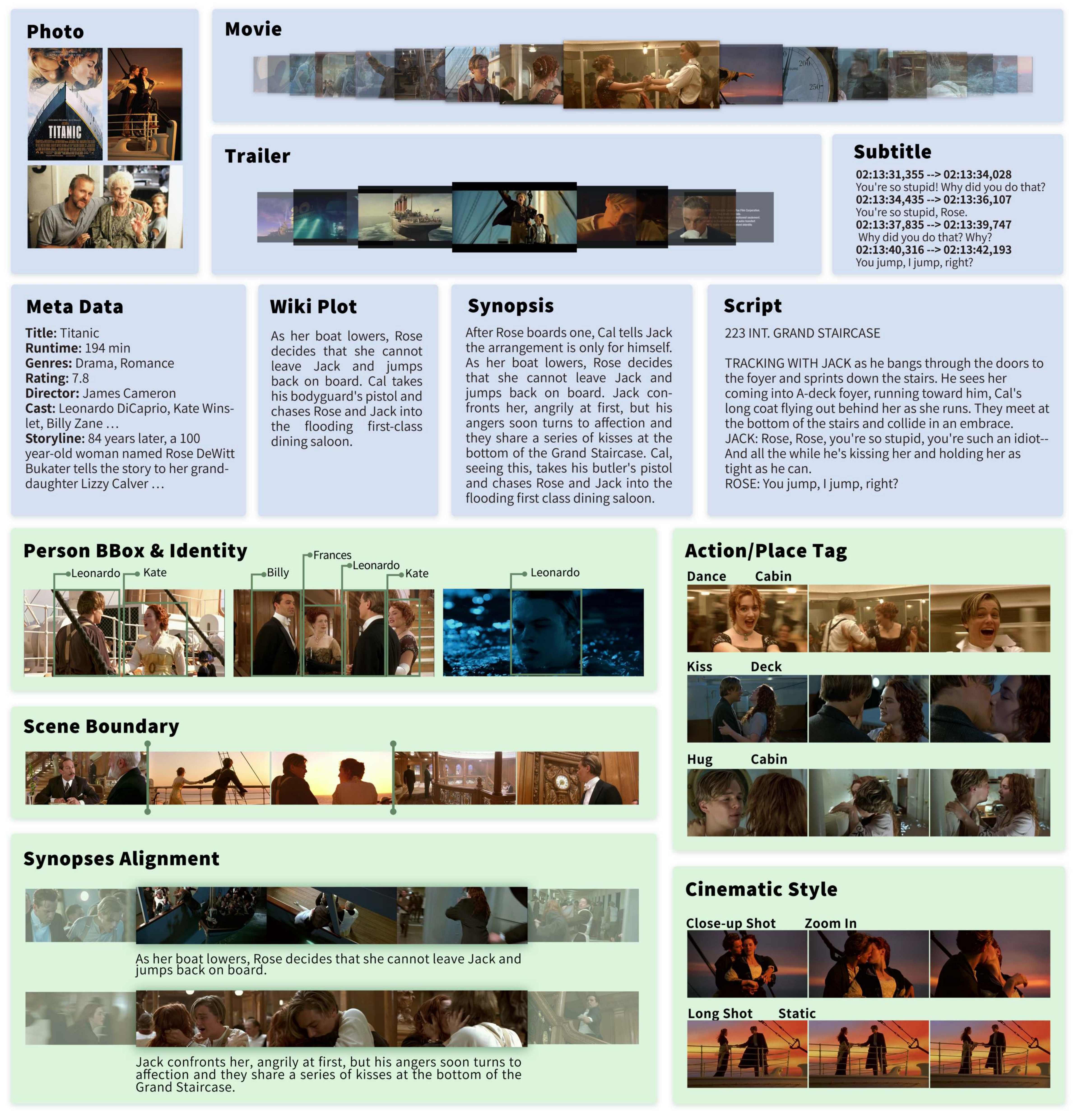}
	\caption{\small
		MovieNet is a holistic dataset for movie understanding, which contains massive data from different modalities and high-quality annotations in different aspects. Here we show some data (in blue) and annotations (in green) of \emph{Titanic} in MovieNet.
	}
	\label{fig:dataset}
\end{figure}


``You jump, I jump, right?'' When Rose gives up the lifeboat and exclaims to Jack, we are all deeply touched by the beautiful moving love story told by the movie \emph{Titanic}. As the saying goes, ``Movies dazzle us, entertain us, educate us, and delight us''. Movie, where characters would face various situations and perform various behaviors in various scenarios, is a reflection of our real world. It teaches us a lot such as the stories took place in the past, the culture and custom of a country or a place, the reaction and interaction of humans in different situations, \etc. Therefore, to understand movies is to understand our world.

It goes not only for human, but also for an artificial intelligence system.
We believe that movie understanding is a good arena for high-level machine intelligence,
considering its high complexity and close relation to the real world.
What's more, compared to web images~\cite{deng2009imagenet} and short videos~\cite{caba2015activitynet}, the hundreds of thousands of movies in history containing rich content and multi-modal information become better nutrition for the data-hungry deep models.

Motivated by the insight above, we build a holistic dataset for movie understanding named \emph{MovieNet} in this paper.
As shown in Fig.~\ref{fig:teaser}, MovieNet comprises three important aspects, namely \emph{data}, \emph{annotation}, and \emph{benchmark}.

First of all, MovieNet contains a large volume of data in multiple modalities,
including movies, trailers, photos, subtitles, scripts and meta information like genres, cast, director, rating \etc.
There are totally $3K$ hour-long videos, $3.9M$ photos, $10M$ sentences of text and $7M$ items of meta information in MovieNet.

From the annotation aspect, MovieNet contains massive labels to support different research topics of movie understanding.
Based on the belief that middle-level entities, \eg character, place, are important for high-level story understanding,
various kinds of annotations on semantic elements are provided in MovieNet, including character bounding box and identity, scene boundary, action/place tag and aligned description in natural language.
In addition, since movie is an art of filming, the cinematic styles, \eg, view scale, camera motion, lighting, \etc, are also beneficial  for comprehensive video analysis. Thus we also annotate the view scale and camera motion for more than $46K$ shots.
Specifically, the annotations in MovieNet include:
(1) $1.1M$ characters with bounding boxes and identities;
(2) $40K$ scene boundaries;
(3) $65K$ tags of action and place;
(4) $12K$ description sentences aligned to movie segments;
(5) $92K$ tags of cinematic styles.

Based on the data and annotations in MovieNet, we exploit some research topics that cover different aspects of movie understanding,
\ie genre analysis, cinematic style prediction, character analysis, scene understanding, and movie segment retrieval.
For each topic, we set up one or several challenging benchmarks.
Then extensive experiments are executed to present the performances of different methods.
By further analysis on the experimental results, we will also show the gap of current approaches towards 
comprehensive movie understanding, as well as the advantages of holistic annotations for throughout video analytics.

To the best of our knowledge,
MovieNet is the first holistic dataset for movie understanding
that contains a large amount of data from different modalities and high-quality annotations in different aspects.
We hope that it would promote the researches on video editing, human-centric situation understanding, story-based video analytics and beyond.


\section{Related Datasets}
\label{sec:related}

\noindent{\textbf{Existing Works.}}
Most of the datasets of movie understanding focus on a specific element of movies, \eg genre~\cite{zhou2010movie,simoes2016movie}, character~\cite{arandjelovic2005automatic,bauml2013semi,haurilet2016naming,nagrani2018benedict,tapaswi2012knock,everingham2006hello,Huang_2018_CVPR}, action~\cite{laptev2008learning,duchenne2009automatic,marszalek2009actions,bojanowski2013finding,bojanowski2014weakly}, scene~\cite{rasheed2005detection,chasanis2008scene,han2011video,park2010exploiting,del2013state,rao2020unified}
and description~\cite{shao2018find}. Also their scale is quite small and the annotation quantities are limited.
For example, \cite{everingham2006hello,tapaswi2012knock,bauml2013semi} take several episodes from TV series for character identification, \cite{laptev2008learning} uses clips from twelve movies for action recognition,
and \cite{park2010exploiting} exploits scene segmentation with only three movies.
Although these datasets focus on some important aspects of movie understanding, their scale is not enough for the data-hungry learning paradigm.
Furthermore, the deep comprehension should go from middle-level elements to high-level story while each existing dataset can only support a single task, causing trouble for comprehensive movie understanding.

\noindent{\textbf{MovieQA.}}
MovieQA~\cite{tapaswi2016movieqa} consists of $15K$ questions designed for $408$ movies. As for sources of information, it contains video clips, plots, subtitles, scripts, and DVS (Descriptive Video Service).
To evaluate story understanding by QA is a good idea, but there are two problems. (1) Middle-level annotations, \eg, character identities,
are missing. Therefore it is hard to develop an effective approach towards high-level understanding.
(2) The questions in MovieQA come from the wiki plot. Thus it is more like a textual QA problem rather than story-based video understanding. A strong evidence is that the approaches based on textual plot can get a much higher accuracy than those based on ``video+subtitle''.

\noindent{\textbf{LSMDC.}}
LSMDC~\cite{rohrbach2015dataset} consists of $200$ movies with audio description (AD) providing linguistic descriptions of movies for visually impaired people.
AD is quite different from the natural descriptions of most audiences, limiting the usage of the models trained on such datasets.
And it is also hard to get a large number of ADs.
Different from previous work~\cite{tapaswi2016movieqa,rohrbach2015dataset}, we provide multiple sources of textual information and different annotations of middle-level entities in MovieNet, leading to a better source for story-based video understanding.


\noindent{\textbf{AVA.}}
Recently, AVA dataset~\cite{gu2018ava}, an action recognition dataset with 430 15-min movie clips 
annotated with 80 spatial-temporal atomic visual actions,
is proposed.
AVA dataset aims at facilitating the task of recognizing atomic visual actions.
However, regarding the goal of story understanding, the AVA dataset is not applicable since 
(1) The dataset is dominated by labels like \emph{stand} and \emph{sit}, making it extremely unbalanced.
(2) Actions like \emph{stand, talk, watch} are less informative in the perspective of story analytics.
Hence, we propose to annotate semantic level actions for both action recognition and story understanding tasks.

\noindent{\textbf{MovieGraphs.}}
MovieGraphs~\cite{vicol2018moviegraphs} is the most related one that provides graph-based annotations of social situations depicted in clips of $51$ movies. The annotations consist of characters, interactions, attributes, \etc.
Although sharing the same idea of multi-level annotations, MovieNet is different from MovieGraphs in three aspects:
(1) MovieNet contains not only movie clips and annotations, but also photos, subtitles, scripts, trailers, \etc, which can provide richer data for various research topics.
(2) MovieNet can support and exploit different aspects of movie understanding while MovieGraphs focuses on situation recognition only.
(3) The scale of MovieNet is much larger than MovieGraphs.

\section{Visit MovieNet: Data and Annotation}
MovieNet contains various kinds of data from multiple modalities and high-quality annotations on different aspects for movie understanding.
Fig.~\ref{fig:dataset} shows the data and annotations of the movie \emph{Titanic} in MovieNet.
Comparisons between MovieNet and other datasets for movie understanding are shown in Tab.~\ref{tab:compare-data} and Tab.~\ref{tab:compare-annotation}.
All these demonstrate the tremendous advantage of MovieNet on both quality, scale and richness.

\begin{table}[t]
	\centering
	\caption{\small Comparison between MovieNet and related datasets in terms of data.}
	\resizebox{\columnwidth}{!}{
	\begin{tabular}{c|c|cccccccc}
		\hline
		\multicolumn{1}{l|}{}  & \# movie & ~trailer~ & ~photo~  & ~meta~  & ~script~ & ~synop.~ & ~subtitle~ & ~plot~ & ~AD  \\ \hline
		MovieQA\cite{tapaswi2016movieqa} & 140  &   &  &     &      &  & \cmark  & \cmark &  \\ \hline
		LSMDC\cite{rohrbach2015dataset} & 200  &  &  &   & \cmark  &  &  &  & \cmark  \\ \hline
		MovieGraphs\cite{vicol2018moviegraphs}  & 51  &  &   &   &   &  &   &    &   \\ \hline
		AVA\cite{gu2018ava} & 430   &  &  &  &  &  &  &  &  \\ \hline
		MovieNet    & 1,100    & \cmark  & \cmark &  \cmark & \cmark & \cmark  &  \cmark  & \cmark &  \\ \hline
	\end{tabular}
}
	\label{tab:compare-data}
\end{table}

\begin{table}[t]
	\centering
	\caption{\small Comparison between MovieNet and related datasets in terms of annotation.}
	\begin{tabular}{c|c|c|c|c|c}
		\hline
		\multicolumn{1}{l|}{}   & \# character  & \# scene & \# cine. tag & \# aligned sent. & \# action/place tag \\ \hline
		MovieQA\cite{tapaswi2016movieqa}                 & -          & -                 & -                  & 15K                 & -                   \\ \hline
		LSMDC\cite{rohrbach2015dataset}                   & -          & -                 & -                  & 128K                 & -                   \\ \hline
		MovieGraphs\cite{vicol2018moviegraphs}              & 22K        &       -            & -                  & 21K                 & 23K                 \\ \hline
		AVA\cite{gu2018ava}              & 116K        &       -            & -                  & -                 & 360K                 \\ \hline
		MovieNet                & 1.1M       & 42K               & 92K                & 25K                 & 65K                \\ \hline
	\end{tabular}
	\label{tab:compare-annotation}
\end{table}

\subsection{Data in MovieNet}
\label{sec:dataset}

\noindent{\textbf{Movie.}}
We carefully selected and purchased the copies of $1,100$ movies, the criteria of which are
(1) colored; (2) longer than $1$ hour; (3) cover a wide range of genres, years and countries.
%

\noindent{\textbf{Metadata.}}
We get the meta information of the movies from IMDb and TMDb\footnote{IMDb: https://www.imdb.com; ~TMDb: https://www.themoviedb.org},
including title, release date, country, genres, rating, runtime, director, cast, storyline, \etc.
Here we briefly introduce some of the key elements, please refer to supplementary material for detail:
(1) Genre is one of the most important attributes of a movie.
There are total $805K$ genre tags from $28$ unique genres in MovieNet.
(2) For cast, we get both their names, IMDb IDs and the character names in the movie.
%
(3) We also provide IMDb ID, TMDb ID and Douban ID of each movie,
with which the researchers can get additional meta information from these websites conveniently.
The total number of meta information in MovieNet is $375K$.
Please note that each kind of data itself, even without the movie, can support some research topics~\cite{huang2020caption}.
So we try to get each kind of data as much as we can.
Therefore the number here is larger than $1,100$.
So as other kinds of data we would introduce below.

\noindent{\textbf{Subtitle.}}
The subtitles are obtained in two ways.
Some of them are extracted from the embedded subtitle stream in the movies.
For movies without original English subtitle,
we crawl the subtitles from YIFY\footnote{https://www.yifysubtitles.com/}.
All the subtitles are manually checked to ensure that they are aligned to the movies.

\noindent{\textbf{Trailer.}}
We download the trailers from YouTube according to their links from IMDb and TMDb.
We found that this scheme is better than previous work~\cite{cascante2019moviescope}, 
which use the titles to search trailers from YouTube,
since the links of the trailers in IMDb and TMDb have been manually checked by the organizers and  audiences.
Totally, we collect $60K$ trailers belonging to $33K$ unique movies.

\noindent{\textbf{Script.}}
Script, where the movement, actions, expression and dialogs of the characters are narrated, is a valuable textual source for research topics of movie-language association.
We collect around $2K$ scripts from IMSDb and Daily Script\footnote{IMSDb: https://www.imsdb.com/; ~DailyScript: https://www.dailyscript.com/}.
The scripts are aligned to the movies by matching the dialog with subtitles.

\noindent{\textbf{Synopsis.}}
A synopsis is a description of the story in a movie written by audiences.
We collect $11K$ high-quality synopses from IMDb, all of which contain more than $50$ sentences.
Synopses are also manually aligned to the movie, which would be introduced in Sec.~\ref{subsec:annotation}.

\noindent{\textbf{Photo.}}
We collect $3.9M$ photos of the movies from IMDb and TMDb,
including poster, still frame, publicity, production art, product, behind the scene and event.

\subsection{Annotation in MovieNet}
\label{subsec:annotation}
To provide a high-quality dataset supporting different research topics on movie understanding,
we make great effort to clean the data and manually annotate various labels on different aspects, including character, scene, event and cinematic style.
Here we just demonstrate the \emph{content} and the \emph{amount} of annotations due to the space limit. Please refer to supplementary material for details.

\noindent{\textbf{Cinematic Styles.}}
Cinematic style, such as view scale, camera movement, lighting and color, is an important aspect of comprehensive movie understanding since it influences how the story is telling in a movie.
In MovieNet, we choose two kinds of cinematic tags for study, namely view scale and camera movement.
Specifically, the view scale include five categories, \ie \emph{long shot}, \emph{full shot}, \emph{medium shot}, \emph{close-up shot} and \emph{extreme close-up shot},
while the camera movement is divided into four classes, \ie \emph{static shot}, \emph{pans and tilts shot}, \emph{zoom in} and \emph{zoom out}.
The original definitions of these categories come from \cite{giannetti1999understanding} and we simplify them for research convenience.
We totally annotate $47K$ shots from movies and trailers, each with one tag of view scale and one tag of camera movement.

\noindent{\textbf{Character Bounding Box and Identity.}}
Person plays an important role in human-centric videos like movies.
Thus to detect and identify characters is a foundational work towards movie understanding.
The annotation process of character bounding box and identity contains $4$ steps:
(1) Some key frames, the number of which is $758K$, from different movies are selected for bounding box annotation.
(2) A detector is trained with the annotations in step-1.
(3) We use the trained detector to detect more characters in the movies and manually clean the detected bounding boxes.
(4) We then manually annotate the identities of all the characters.
To make the cost affordable, we only keep the top
$10$ cast in credits order according to IMDb, which can cover the main characters for most movies.
Characters not belong to credited cast were labeled as ``others''. 
In total, we got $1.1M$ instances of $3,087$ unique credited cast and $364K$ ``others''.


\noindent{\textbf{Scene Boundary.}}
In terms of temporal structure,
a movie contains two hierarchical levels -- shot, and scene.
Shot is the minimal visual unit of a movie while
scene is a sequence of continued shots that are semantically related.
To capture the hierarchical structure of a movie is important for movie understanding.
Shot boundary detection has been well solved by~\cite{sidiropoulos2011temporal},
while scene boundary detection, also named scene segmentation, remains an open question.
In MovieNet, we manually annotate the scene boundaries to support the researches on scene segmentation,
resulting in $42K$ scenes.

\noindent{\textbf{Action/Place Tags.}}
To understand the event(s) happened within a scene, action and place tags are required.
Hence, we first split each movie into clips according to the scene boundaries and then manually annotated place and action tags for each segment.
For place annotation, each clip is annotated with multiple place tags, \eg, \{deck, cabin\}. While for action annotation, we first detect sub-clips that contain characters and actions, then we assign multiple action tags to each sub-clip.
We have made the following efforts to keep tags diverse and informative:
(1) We encourage the annotators to create new tags.
(2) Tags that convey little information for story understanding, \eg, \emph{stand} and \emph{talk}, are excluded. 
Finally, we merge the tags and filtered out $80$ actions and $90$ places with a minimum frequency of $25$ as the final annotations.
In total, there are $42K$ segments with $19.6K$ place tags and $45K$ action tags. 

\noindent{\textbf{Description Alignment}}
Since the event is more complex than character and scene,
a proper way to represent an event is to describe it with natural language.
Previous works have already aligned script~\cite{marszalek2009actions}, Descriptive Video Service (DVS)~\cite{rohrbach2015dataset}, book~\cite{zhu2015aligning} or wiki plot~\cite{tapaswi2014story,tapaswi2015aligning,tapaswi2016movieqa} to movies.
However, books cannot be well aligned since most of the movies would be quite different from their books.
DVS transcripts are quite hard to obtain, limiting the scale of the datasets based on them~\cite{rohrbach2015dataset}.
Wiki plot is usually a short summary that cannot cover all the important events of the movie.
Considering the issues above,
we choose synopses as the story descriptions in MovieNet.
The associations between the movie segments and the synopsis paragraphs are manually annotated by three different annotators with a coarse-to-fine procedure.
Finally, we obtained $4,208$ highly consistent paragraph-segment pairs.

\section{Play with MovieNet: Benchmark and Analysis}
\label{sec:experiment}

With a large amount of data and holistic annotations, MovieNet can support various research topics. 
In this section, we try to analyze movies from five aspects, namely \emph{genre}, \emph{cinematic style}, \emph{character}, \emph{scene} and \emph{story}.
For each topic, we would set up one or several benchmarks based on MovieNet.
Baselines with currently popular techniques and analysis on experimental results are also provided to
show the potential impact of MovieNet in various tasks.
The topics of the tasks have covered different perspectives of comprehensive movie understanding.
But due to the space limit, here we can only touched the tip of the iceberg. More detailed analysis are provided in the supplementary material and more interesting topics to be exploited are introduced in Sec.~\ref{sec:discussion}.

\begin{table}[t]
	\caption{\small
		(a). Comparison between MovieNet and other benchmarks for genre analysis.
		(b). Results of some baselines for genre classification in MovieNet}
	\begin{minipage}[t]{0.5\linewidth}\centering
		\subfloat[\label{tab:data-genres}]{
			\begin{tabular}{l|c|ccc}
				\hline
				& genre & movie & trailer & photo \\ \hline
				MGCD\cite{zhou2010movie}       & 4       & - & 1.2K      & -   \\
				LMTD\cite{simoes2016movie}       & 4      & -  & 3.5K      & -  \\
				MScope\cite{cascante2019moviescope} & 13    & -   & 5.0K      & 5.0K \\ \hline
				MovieNet   & \textbf{21}    & \textbf{1.1K}   & \textbf{68K}     & \textbf{1.6M} \\ \hline
		\end{tabular}}
		\hfill
	\end{minipage}
	\begin{minipage}[t]{0.5\linewidth}\centering
		\subfloat[\label{tab:exp-genres}]{
			\begin{tabular}{c|c|ccc}
				\hline
				Data                     & Model           & r@0.5  & p@0.5 & mAP  \\ \hline
				\multirow{3}{*}{Photo}   & VGG16~\cite{simonyan2014very}          & 27.32 & 66.28 & 32.12 \\
				& ResNet50~\cite{he2016deep}       & \textbf{34.58} & \textbf{72.28} & \textbf{46.88} \\ \hline
				\multirow{4}{*}{Trailer}
				& TSN-r50\cite{wang2016temporal}         & 17.95 &	\textbf{78.31} & 43.70 \\ 
				& I3D-r50~\cite{carreira2017quo}         & 16.54 & 69.58 & 35.79 \\
				& TRN-r50~\cite{zhou2018temporal}         & \textbf{21.74} & 77.63 & \textbf{45.23} \\ \hline
		\end{tabular}}
	\end{minipage}
\end{table}

\begin{figure}[t]
	\subfloat[\label{fig:genre-framework}]{
		\includegraphics[width=0.56\linewidth]{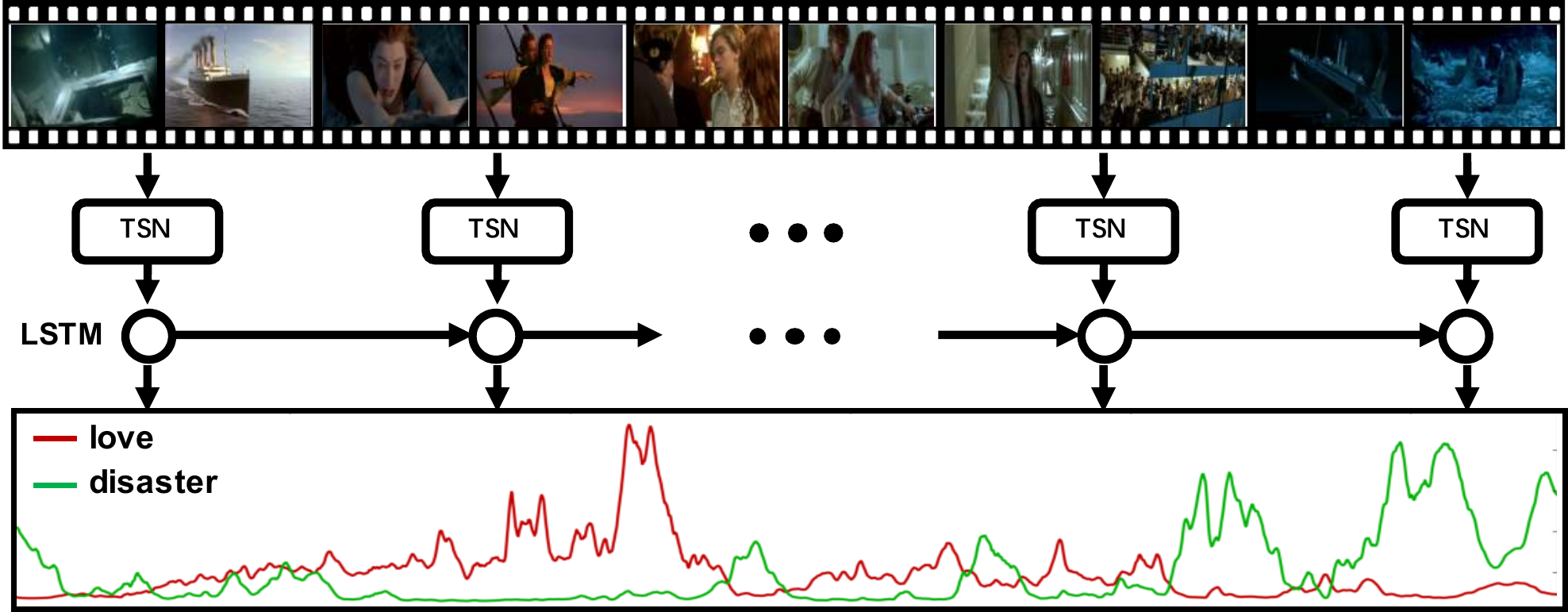}
	} \hfill
	\subfloat[\label{fig:genre-sample}]{
		\includegraphics[width=0.43\linewidth]{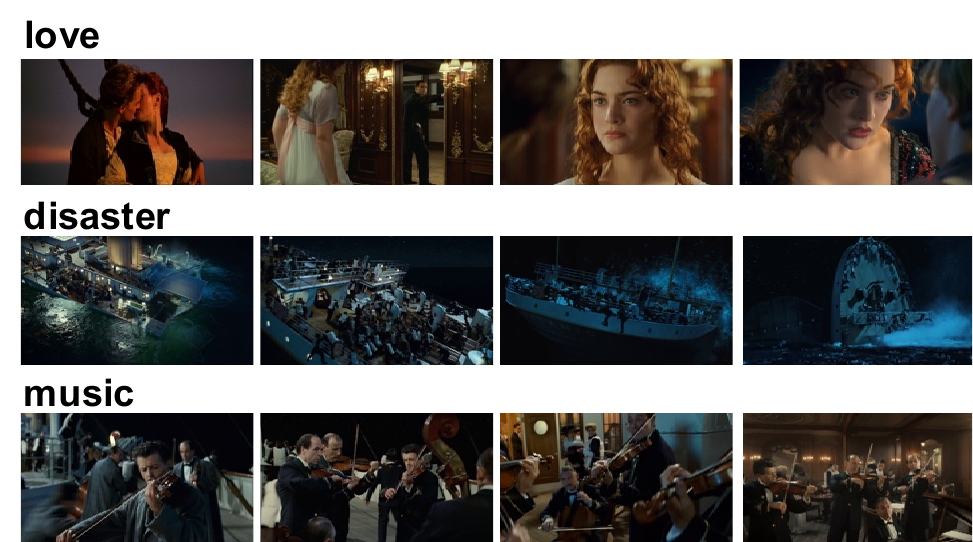}
	}
	\caption{\small
		(a). Framework of genre analysis in movies.
		(b). Some samples of genre-guided trailer generation for movie \emph{Titanic}.
	}
\end{figure}

\subsection{Genre Analysis}
Genre is a key attribute for any media with artistic elements.
To classify the genres of movies has been widely studied by previous works~\cite{zhou2010movie,simoes2016movie,cascante2019moviescope}.
But there are two drawbacks for these works.
(1) The scale of existing datasets is quite small.
(2) All these works focus on image or trailer classification while ignore a more important problem, \ie how to analyze the genres of a long video.

MovieNet provides a large-scale benchmark for genre analysis,
which contains $1.1K$ movies, $68K$ trailers and $1.6M$ photos.
The comparison between different datasets are shown in Tab.~\ref{tab:data-genres}, from which we can see that MovieNet is much larger than previous datasets. 

Based on MovieNet, we first provide baselines for both image-based and video-based genre classification, the results are shown Tab.~\ref{tab:exp-genres}. Comparing the result of genre classification in small datasets~\cite{simoes2016movie,cascante2019moviescope} to ours in MovieNet, we find that the performance drops a lot when the scale of the dataset become larger. 
The newly proposed MovieNet brings two challenges to previous methods.
(1) Genre classification in MovieNet becomes a long-tail recognition problem where the label distribution is extremely unbalanced. For example, the number of ``Drama'' is $40$ times larger than that of ``Sport'' in MovieNet.
(2) Genre is a high-level semantic tag depending on action, clothing and facial expression of the characters, and even BGM. Current methods are good at visual representation. When facing a problem that need to consider higher-level semantics, they would all fail.
We hope MovieNet would promote researches on these challenging topics.

Another new issue to address is how to analyze the genres of a movie. Since movie is extremely long and not all segments are related to its genres, this problem is much more challenging. 
Following the idea of learning from trailers and applying to movies~\cite{huang2018trailers}, we adopt the visual model trained with trailers as shot-level feature extractor. 
Then the features are fed to a temporal model to capture the temporal structure of the movie. The overall framework is shown in Fig.~\ref{fig:genre-framework}. With this approach, we can get the genre response curve of a movie. Specifically, we can predict which part of the movie is more relevant to a specific genre. What's more, the prediction can also be used for genre-guided trailer generation, as shown in Fig.~\ref{fig:genre-sample}. 
From the analysis above, we can see that MovieNet would promote the development of this challenging and valuable research topic.

\begin{table}[b]
	\caption{\small
		(a). Comparison between MovieNet and other benchmarks for cinematic style prediction.
		(b). Results of some baselines for cinematic style prediction in MovieNet}
	\begin{minipage}[t]{0.5\linewidth}\centering
		\subfloat[\label{tab:data-cinematic}]{
			\begin{tabular}{c|c|c|cc}
				\hline
				& shot & video & scale & move.   \\ \hline
				Lie 2014~\cite{lie} & 327   &  327     &          &   \cmark   \\
				Sports 2007~\cite{2007accv}  & 1,364   &  8     &   \cmark &     \\
				Context 2011~\cite{xu2011using}   & 3,206   & 4    & \cmark         &      \\
				Taxon 2009~\cite{wang2009taxonomy} & 5,054 & 7  &         & \cmark  \\
				MovieNet & 46,857  & 7,858  &  \cmark  & \cmark           \\
				\hline
		\end{tabular}}
		\hfill
	\end{minipage}
	\begin{minipage}[t]{0.5\linewidth}\centering
		\subfloat[\label{tab:exp-cinematic}]{
			\begin{tabular}{c|c|c}
				\hline
				Method         & scale acc. & move. acc. \\ \hline
				I3D~\cite{carreira2017quo}  & 76.79  & 78.45 \\
				TSN~\cite{wang2016temporal} & 84.08  & 70.46 \\
				TSN+R$^3$Net\cite{deng2018r3net}  & \textbf{87.50}  & \textbf{80.65} \\ \hline
		\end{tabular}}
	\end{minipage}
\end{table}

\subsection{Cinematic Style Analysis}

\begin{table}[t]
	\begin{minipage}{0.48\linewidth}\centering
		\caption{\small Datasets for person analysis.}
		\begin{tabular}{l|ccc}
			\hline
			& ~~~ID~~~ & instance & source \\ \hline
			COCO\cite{lin2014microsoft}   & -  & 262K   & web image      \\
			CalTech\cite{dollar2011pedestrian}  & -   & 350K & surveillance    \\
			Market\cite{zheng2015scalable}   & 1,501       & 32K  & surveillance       \\
			CUHK03\cite{li2014deepreid}   & 1,467       & 28K    & surveillance     \\
			AVA\cite{gu2018ava} & - & 426K & movie \\
			CSM\cite{huang2018person}   & 1,218       & 127K   & movie      \\ \hline
			MovieNet & \textbf{3,087}   & \textbf{1.1M}  & movie       \\ \hline
		\end{tabular}
		\label{tab:data-person}
	\end{minipage}
	\hfill
	\begin{minipage}{0.48\textwidth}
		\centering
		\includegraphics[width=\linewidth]{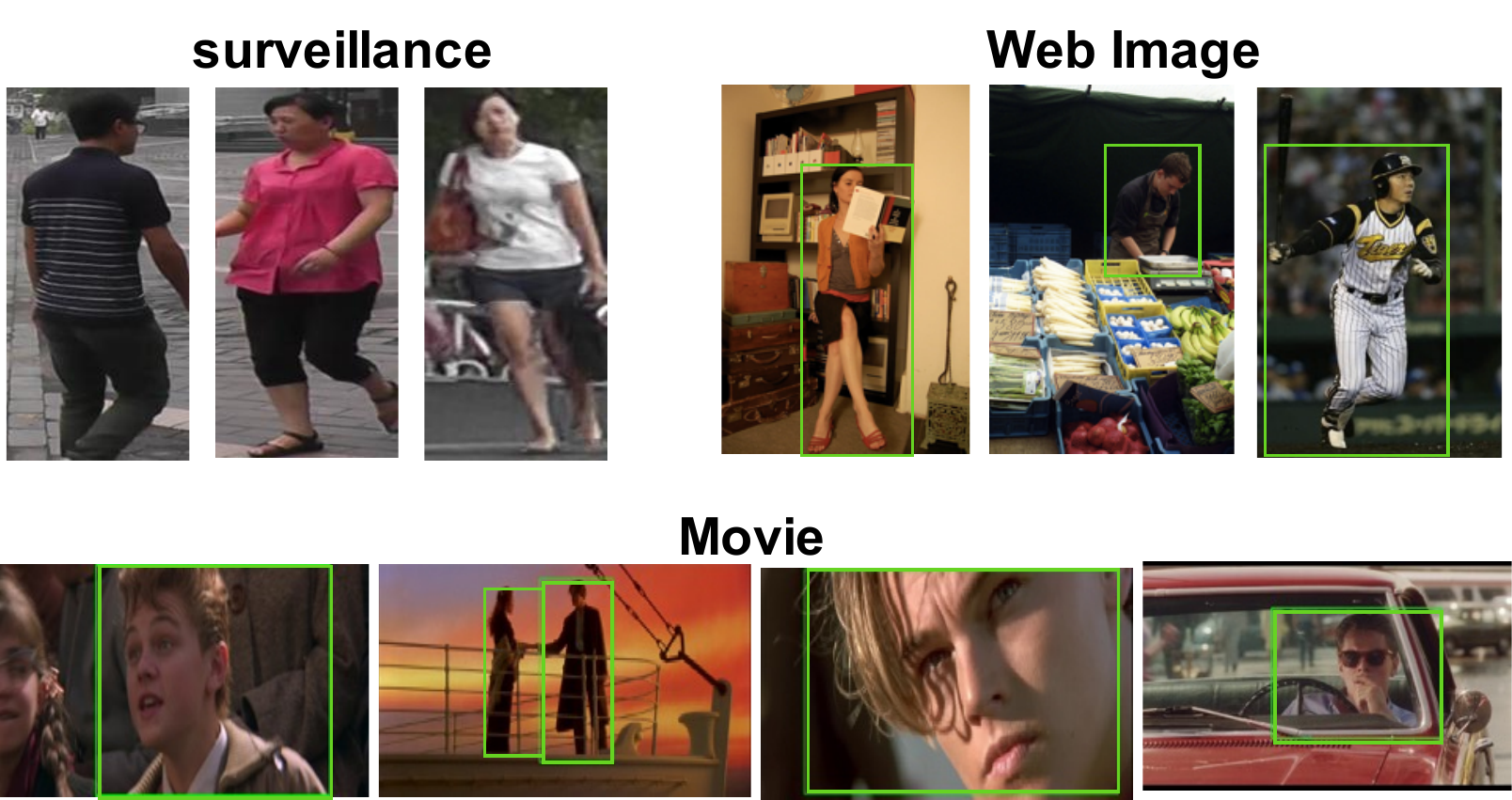}
		\captionof{figure}{ \small
			Persons in different data sources
		}
		\label{fig:sample-person}
	\end{minipage}
\end{table}

\begin{table}[t]
	\caption{\small
		Results of (a). Character Detection and (b).Character Identification}
	\begin{minipage}[]{0.45\linewidth}\centering
		\subfloat[\label{tab:exp-person-det}]{
			\begin{tabular}{c|c|c}
				\hline
				Train Data                  & Method            & ~~mAP~~ \\ \hline
				COCO\cite{lin2014microsoft}     & FasterRCNN   & 81.50 \\ \hline
				Caltech\cite{dollar2011pedestrian} & FasterRCNN   &  5.67   \\ \hline
				CSM\cite{huang2018person}   & FasterRCNN   &  89.91   \\ \hline
				\multirow{3}{*}{MovieNet} & FasterRCNN   & 92.13 \\ \cline{2-3}
				& RetinaNet      &  91.55   \\ \cline{2-3} 
				& CascadeRCNN  & \textbf{95.17 } \\ \hline
			\end{tabular}
		}
	\end{minipage}
	\hfill
	\begin{minipage}[]{0.55\linewidth}\centering
		\subfloat[\label{tab:exp-person-id}]{
			\begin{tabular}{c|c|c|c}
				\hline
				Train Data          & cues        & Method      & ~~mAP~~ \\ \hline
				Market\cite{zheng2015scalable}     & body   & r50-softmax & 4.62 \\ \hline
				CUHK03\cite{li2014deepreid}       & body   & r50-softmax & 5.33 \\ \hline
				CSM\cite{huang2018person}         & body  & r50-softmax & 26.21 \\ \hline
				\multirow{3}{*}{MovieNet} & body & r50-softmax & 32.81    \\ \cline{2-4} 
				& body+face  & two-step\cite{loy2019wider}  &  63.95  \\ \cline{2-4} 
				& body+face  & PPCC\cite{huang2018person} & \textbf{75.95} \\ \hline
			\end{tabular}
		}
	\end{minipage}
\end{table}

As we mentioned before, cinematic style is about how to present the story to audience in the perspective of filming art.
For example, a \emph{zoom in} shot is usually used to attract the attention of audience to a specific object.
In fact, cinematic style is crucial for both video understanding and editing.
But there are few works focusing on this topic and no large-scale datasets for this research topic too.

Based on the tags of cinematic style we annotated in MovieNet,
we set up a benchmark for cinematic style prediction.
Specifically, we would like to recognize the view scale and camera motion of each shot.
Comparing to existing datasets, MovieNet is the first dataset that covers both view scale and camera motion,
and it is also much larger, as shown in Tab.~\ref{tab:data-cinematic}.
Several models for video clip classification such as TSN~\cite{wang2016temporal} and I3D~\cite{carreira2017quo} are applied to tackle this problem, the results are shown in Tab.~\ref{tab:exp-cinematic}.
Since the view scale depends on the portion of the subject
in the shot frame, 
to detect the subject is important for cinematic style prediction.
Here we adopt the approach from saliency detection~\cite{deng2018r3net} to get the subject maps of each shot, with which better performances are achieved, as shown in Tab.~\ref{tab:exp-cinematic}.
Although utilizing subject points out a direction for this task, there is still a long way to go.
We hope that MovieNet can promote the development of this important but ignored topic for video understanding.

\subsection{Character Recognition}
It has been shown by existing works~\cite{vicol2018moviegraphs,xia2020online,loy2019wider}
that movie is a human-centric video where characters play an important role.
Therefore, to detect and identify characters is crucial for movie understanding.
Although person/character recognition is not a new task,
all previous works either focus on other data sources~\cite{zheng2015scalable,li2014deepreid,lin2014microsoft} or small-scale benchmarks~\cite{haurilet2016naming,bauml2013semi,tapaswi2012knock},
leading to the results lack of convincingness for character recognition in movies.

We proposed two benchmarks for character analysis in movies, namely, character detection and character identification.
We provide more than $1.1M$ instances from $3,087$ identities to support these benchmarks.
As shown in Tab.~\ref{tab:data-person}, MovieNet contains much more instances and identities comparing to some popular datasets about person analysis.
The following sections will show the analysis on character detection and identification respectively.

\noindent{\textbf{Character Detection.}}
Images from different data sources would have large domain gap, as shown in Fig.~\ref{fig:sample-person}.
Therefore, a character detector trained on general object detection dataset, \eg COCO~\cite{lin2014microsoft}, or pedestrian dataset, \eg CalTech~\cite{dollar2011pedestrian}, is not good enough for detecting characters in movies.
This can be supported by the results shown in Tab.~\ref{tab:exp-person-det}.
To get a better detector for character detection, we train different popular models~\cite{ren2015faster,lin2017focal,cai2018cascade} with MovieNet 
using toolboxes from~\cite{mmdetection,Chen_2019_CVPR}.
We can see that with the diverse character instances in MovieNet, a Cascade R-CNN trained with MovieNet can achieve extremely high performance, \ie $95.17$\% in mAP.
That is to say, character detection can be well solved by a large-scale movie dataset with current SOTA detection models.
This powerful detector would then benefit research on character analysis in movies.

\noindent{\textbf{Character Identification.}}
To identify the characters in movies is a more challenging problem,
which can be observed by the diverse samples shown in Fig.~\ref{fig:sample-person}.
We conduct different experiments based on MovieNet, the results are shown in Tab.~\ref{tab:exp-person-id}.
From these results, we can see that:
(1) models trained on ReID datasets are inefficient for character recognition due to domain gap;
(2) to aggregate different visual cues of an instance is important for character recognition in movies;
(3) the current state-of-the-art can achieve $75.95$\% mAP, which demonstrates that it is a challenging problem which need to be further exploited.

\subsection{Scene Analysis}

\begin{table}[tb]
	\begin{minipage}[t]{0.47\textwidth}\centering
	\caption{\small Dataset for scene analysis.}
	\label{tab:data-scene}
	\begin{tabular}{l|c|c|c}
		\hline
		&  scene & action & place     \\ 				\hline
		OVSD~\cite{rotman2017optimal}   & 300  & - & -      \\
		BBC~\cite{baraldi2015deep}    & 670  & - & -     \\
		Hollywood2~\cite{marszalek2009actions}    & -         & 1.7K          & 1.2K         \\
		MovieGraph\cite{vicol2018moviegraphs}    & -     & 23.4K         & 7.6K         \\
		AVA~\cite{gu2018ava}              & -    & 360K        & -            \\ \hline
		MovieNet           & \textbf{42K}   & \textbf{45.0K}     & \textbf{19.6K}       \\
		\hline
	\end{tabular}
\end{minipage}
\hfill
	\begin{minipage}[t]{0.49\linewidth}\centering
		\caption{\small Datasets for story understanding in movies in terms of 
			(1) number of sentences per movie; (2) duration (second) per segment.}
		\label{tab:msa_compare}
		\begin{tabular}{c|c|c}
			\hline
			Dataset & sent./mov. & dur./seg. \\ \hline
			MovieQA~\cite{tapaswi2016movieqa}   & 35.2   & 202.7       \\
			MovieGraphs~\cite{vicol2018moviegraphs}   & 408.8   & 44.3       \\ \hline
			MovieNet     & 83.4   & 428.0         \\ \hline
		\end{tabular}
	\end{minipage}
\end{table}

\begin{table}[t]
	\centering
	\begin{minipage}[]{0.54\linewidth}\centering
		\caption{\small Results of Scene Segmentation}
		\label{tab:exp-scene-seg}	
		\begin{tabular}{c|c|ccc}
			\hline
			Dataset & Method            
			& AP($\uparrow$) & $M_{iou}$($\uparrow$)  \\
			\hline 
			OVSD~\cite{rotman2017optimal} & MS-LSTM  & 0.313 & 0.387 \\ \hline
			BBC~\cite{baraldi2015deep} & MS-LSTM &  0.334 & 0.379 \\
			\hline
			\multirow{3}{*}{MovieNet} 
			& Grouping~\cite{rotman2017optimal} & 0.336  & 0.372 \\
			& Siamese~\cite{baraldi2015deep}  & 0.358 & 0.396 \\
			& MS-LSTM & \textbf{0.465} & \textbf{0.462} \\			
			\hline
		\end{tabular}
	\end{minipage}
	\hfill
	\begin{minipage}[]{0.45\linewidth}\centering
		\caption{\small Results of Scene Tagging}
		\addtolength{\tabcolsep}{4pt}
		\label{tab:exp-scene-tag}
		\begin{tabular}{c|c|c}
			\hline
			Tags & Method     & mAP   \\ \hline
			\multirow{3}{*}{action} & TSN~\cite{wang2016temporal}     & 14.17 \\
			& I3D~\cite{carreira2017quo}  & 20.69 \\ 
			& SlowFast~\cite{feichtenhofer2019slowfast} & \textbf{23.52}\\ \hline
						\multirow{2}{*}{place} &  I3D~\cite{carreira2017quo}  &  7.66\\
			& TSN~\cite{wang2016temporal}   & \textbf{8.33} \\ \hline
		\end{tabular}
	\end{minipage}
\end{table}

As mentioned before, scene is the basic semantic unit of a movie.
Therefore, it is important to analyze the scenes in movies.
The key problems in scene understanding is probably \emph{where is the scene boundary} and \emph{what is the content in a scene}.
As shown in Tab.~\ref{tab:data-scene}, MovieNet, which contains more than $43$K scene boundaries and $65$K action/place tags,
is the only one that can support both \emph{scene segmentation} and \emph{scene tagging}.
What's more, the scale of MovieNet is also larger than all previous works.

\noindent{\textbf{Scene Segmentation}}
We first test some baselines~\cite{rotman2017optimal,baraldi2015deep} for scene segmentation.
In addition, we also propose a sequential model, named Multi-Semtantic LSTM (MS-LSTM) based on Bi-LSTMs~\cite{graves2005framewise,rao2020local} to study the gain brought by using multi-modality and multiple semantic elements, including audio, character, action and scene.
From the results shown in Tab.~\ref{tab:exp-scene-seg},
we can see that
(1) Benefited from large scale and high diversity, models trained on MovieNet can achieve better performance.
(2) Multi-modality and multiple semantic elements are important for scene segmentation, which highly raise the performance.

\noindent{\textbf{Action/Place Tagging}}
To further understand the stories within a movie,
it is essential to perform analytics on the key elements of storytelling, \ie, place and action. 
We would introduce two benchmarks in this section.
Firstly, for action analysis, the task is multi-label action recognition that aims to 
recognize all the human actions or interactions in a given video clip. 
We implement three standard action recognition models, \ie, TSN~\cite{wang2016temporal}, I3D~\cite{carreira2017quo} and SlowFast Network~\cite{feichtenhofer2019slowfast} modified from~\cite{mmaction2019} in experiments. 
Results are shown in Tab.~\ref{tab:exp-scene-tag}.
For place analysis, we propose another benchmark for multi-label place classification.
We adopt I3D~\cite{carreira2017quo} and TSN~\cite{wang2016temporal} as our baseline models and the results are shown in Tab.~\ref{tab:exp-scene-tag}.
From the results, we can see that action and place tagging is an extremely challenging problem due to the high diversity of different instances.

\subsection{Story Understanding}

\begin{figure}[t]
	\centering
	\includegraphics[width=\linewidth]{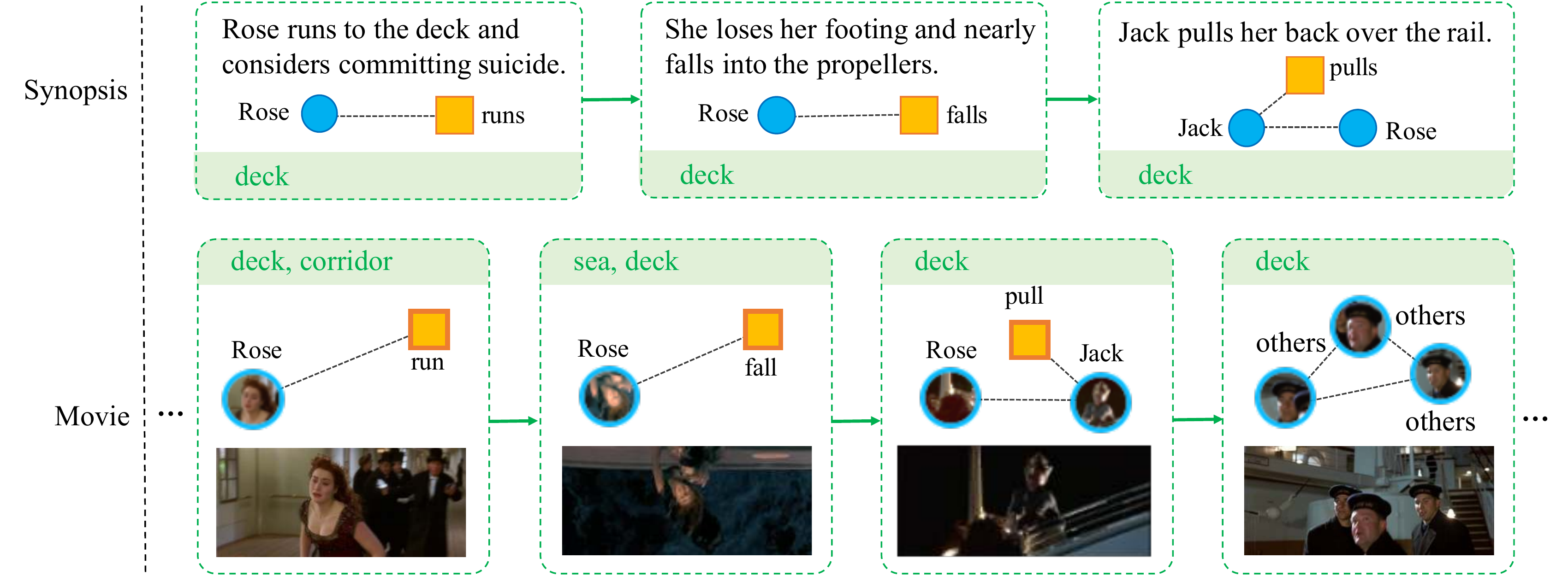}
	\caption{\small
		Example of synopses paragraph and movie segment in MovieNet-MSR. It demonstrate the spatial-temporal structures
		of stories in movies and synopses. We can also see that character, action and place are the key element for story understanding.
	}
	\label{fig:story_graph}
\end{figure}

Web videos are broadly adopted in previous works~\cite{caba2015activitynet,xu2016msr} as the source of video understanding.
Compared to web videos, the most distinguishing feature of movies is the story.
Movies are created to tell stories and the most explicit way to demonstrate a story is to describe it using natural language, \eg synopsis.
Inspired by the above observations,
we choose the task of movie segment retrieval with natural language to analyze the stories in movies.
Based on the aligned synopses in MovieNet, we set up a benchmark for movie segment retrieval.
Specifically, given a synopsis paragraph, we aim to find the most relevant movie segment that covers the story in the paragraph.
It is a very challenging task due to the rich content in movie and high-level semantic descriptions in synopses.
Tab.~\ref{tab:msa_compare} shows the comparison of our benchmark dataset with other related datasets.
We can see that our dataset is more complex in terms of descriptions compared with MovieQA~\cite{tapaswi2016movieqa}
while the segments are longer and contain more information than those of MovieGraphs~\cite{vicol2018moviegraphs}.

Generally speaking, a story can be summarized as ``\emph{somebody do something in some time at some place}''.
As shown in Fig.~\ref{fig:story_graph}, both stories represented by language and video can be composed 
as sequences of \{character, action, place\} graphs.
That being said, to understand a story is to (1) recognize the key elements of story-telling, namely, character, action, place \etc;
(2) analyze the spatial-temporal structures of both movie and synopsis.
Hence, our method first leverage middle-level entities (\eg character, scene), as well as multi-modality (\eg subtitle) to assist retrieval. Then we explore the spatial-temporal structure from both movies and synopses by formulating
middle-level entities into graph structures. 
Please refer to supplementary material for details.

\noindent{\textbf{Using middle-level entities and multi-modality.}}
We adopt VSE~\cite{frome2013devise} as our baseline model where the vision and language features are embedded into a joint space.
Specifically, the feature of the paragraph is obtained by taking the average of Word2Vec~\cite{mikolov2013efficient} feature of each sentence while the visual feature is obtained by taking the average of the appearance feature extracted from ResNet~\cite{he2016deep} on each shot.
We add subtitle feature to enhance visual feature.
Then different semantic elements including character, action and cinematic style are aggregated in our framework. 
We are able to obtain action features and character features thanks to the models trained on other benchmarks on MovieNet, \eg, action recognition and character detection.
Furthermore, we observe that the focused elements vary under different cinematic styles. For example, we should focus more on actions in a full shot while more on character and dialog in a close-up shot.
Motivated by this observation, we propose a cinematic-style-guided attention module that predicts the weights over each element (\eg, action, character) within a shot, which would be used to enhance the visual features. 
The experimental results are shown in Tab.~\ref{tab:msa_exp}. 
Experiments show that by considering different elements of the movies, the performance improves a lot. 
We can see that a holistic dataset which contains holistic annotations to support middle-level entity analyses is important for movie understanding.

\noindent{\textbf{Explore spatial-temporal graph structure in movies and synopses.}}
Simply adding different middle-level entities improves the result.
Moreover, as shown in Fig.~\ref{fig:story_graph}, we observe that stories in movies and synopses persist two important structure:
(1) the temporal structure in movies and synopses is that the story can be composed as a sequence of events 
following a certain temporal order.
(2) the spatial relation of different middle-level elements, \eg, character co-existence and their interactions, can be formulated
as graphs.
We implement the method in~\cite{xiong2019graph} to formulate the above structures as two graph matching problems.
The result are shown in Tab.~\ref{tab:msa_exp}.
Leveraging the graph formulation for the internal structures of stories in movies and synopses,
the retrieval performance can be further boosted, which in turn, show that
the challenging MovieNet would provide a better source to story-based movie understanding. 

\begin{table}[t]
	\centering
	\caption{\small Results of movie segment retrieval. Here, G stands for global appearance feature, S for subtitle feature, A for action, P  for character and C for cinematic style.}
	\label{tab:msa_exp}
	\begin{tabular}{l|cccc}
		\hline
		Method             & Recall@1   & Recall@5   & Recall@10  & MedR \\ \hline
		Random             & 0.11  & 0.54  & 1.09  & 460  \\
		G             & 3.16 & 11.43 & 18.72 & 66   \\
		G+S             & 3.37  & 13.17 & 22.74 & 56   \\
		G+S+A      & 5.22  & 13.28 & 20.35 & 52   \\
		G+S+A+P & 18.50 & 43.96 & 55.50 & 7    \\
		G+S+A+P+C & 18.72 & 44.94 & 56.37 & 7    \\ 
		MovieSynAssociation~\cite{xiong2019graph} & \textbf{21.98} & \textbf{51.03} & \textbf{63.00} & \textbf{5}    \\ \hline
	\end{tabular}
\end{table}




\section{Discussion and Future Work}
\label{sec:discussion}

In this paper, we introduce MovieNet, a holistic dataset containing different aspects of annotations to support comprehensive movie understanding.

We introduce several challenging benchmarks on different aspects of movie understanding, \ie discovering filming art, recognizing middle-level entities and understanding high-level semantics like stories.
Furthermore, the results of movie segment retrieval demonstrate that integrating filming art and middle-level entities
according to the internal structure of movies would be helpful for story understanding.
These in turn, show the effectiveness of holistic annotations.


In the future, our work would go on in two aspects.
(1) \textbf{Extending the Annotation.}
Currently our dataset covers $1,100$ movies.
In the future, we would further extend the dataset to include more movies and annotations. 
(2) \textbf{Exploring more Approaches and Topics.}
To tackle the challenging tasks proposed in the paper, we would explore more effective approaches.
Besides, there are more meaningful and practical topics that can be addressed with MovieNet from the perspective of video editing, such as movie deoldify, trailer generation, \etc


\clearpage
\appendix
\renewcommand\thefigure{\thesection\arabic{figure}} 
\setcounter{figure}{0} 
\renewcommand\thetable{\thesection\arabic{table}} 
\setcounter{table}{0} 

\section*{Supplementary Material}
\label{sec:overview}
In the following sections, we provide overall details
about MovieNet, including data, annotation, experiments and the toolbox. The content is organized as follows:

(1) We provide details about the content of particular data 
and how to collect and clean them in Sec.~\ref{sec:data}:
\begin{itemize}
	\item \textbf{Meta Data.} The list of meta data is given followed
	by the content of these meta data. See Sec.~\ref{subsec:metadata}.
	\item \textbf{Movie.} The statistics of the movies are provided. 
	See Sec.~\ref{subsec:movie}
	\item \textbf{Subtitle.} The collection and post-processing procedure
	of obtaining and aligning subtitles are given.
	See Sec.~\ref{subsec:subtitle}.
	\item \textbf{Trailer.} We provide the process of selecting and processing
	the trailers.
	See Sec.~\ref{subsec:trailer}.
	\item \textbf{Script.} We automatically align the scripts to movies. The details
	of the method will be presented.
	See Sec.~\ref{subsec:script}.
	\item \textbf{Synopsis.} The statistics of synopsis will
	be introduced.
	See Sec.~\ref{subsec:synopsis}.
	\item \textbf{Photo.} The statistics and some examples of photo will be shown.
	See Sec.~\ref{subsec:photo}.
\end{itemize}

(2) We demonstrate annotation in MovieNet with the 
description about the design of annotation interface and workflow,
see Sec.~\ref{sec:annotation}.
 
 \begin{itemize}
 	\item \textbf{Character Bounding Box and Identity.} 
 	We provide step by step procedure of collecting images and 
 	annotating the images with a semi-automatic algorithm.
 	See Sec.~\ref{subsec:person_anno}.
 	\item \textbf{Cinematic Styles.} 
 	We present the analytics on cinematic styles and introduce
 	the workflow and interface of annotating cinematic styles.
 	See Sec.~\ref{subsec:cs_anno}.
 	\item \textbf{Scene Boundaries.} 
 	We demonstrate how to effectively annotate scene boundaries
 	with the help of an optimized annotating workflow.
 	See Sec.~\ref{subsec:sseg_anno}.
 	\item \textbf{Action and Place Tags.} 
 	We describe the procedure of jointly labeling the action and place
 	tags over movie segments. The workflow and interface are presented.
 	See Sec.~\ref{subsec:stag_anno}.
 	\item \textbf{Synopsis Alignment.} 
 	We provide the introduction of an efficient coarse-to-fine
 	annotating workflow to align a synopsis paragraph to a movie segment.
 	See Sec.~\ref{subsec:syn_anno}.
 	\item \textbf{Trailer Movie Alignment.}
 	We introduce a automatic approach that align shots in trailers to 
 	the original movies. This annotation facilitate tasks like trailer generation.
 	See Sec.~\ref{subsec:trailer_align}.
 \end{itemize}
 
(3) We set up several benchmarks on our MovieNet and conduct
experiments on each benchmark. The implementation details of
experiments on each benchmark will be introduced in Sec.~\ref{sec:supp_experiment}:

 \begin{itemize}
	\item \textbf{Genre Classification.} 
	Genre Classification is a multi-label classification task build
	on MovieNet genre classification benchmark. See details at Sec.~\ref{subsec:exp_genre}.
	\item \textbf{Cinematic Styles Analysis.} 
	On MovieNet cinematic style prediction benchmark, there are two classification
	tasks, namely \emph{scale classification} and \emph{movement classification}.
	See Sec.~\ref{subsec:exp_cs} for implementation details.
	\item \textbf{Character Detection.} 
	We introduce the detection task as well as model, implementation details
	on MovieNet character detection benchmarks. See Sec.~\ref{subsec:exp_pdet}.
	\item \textbf{Character Identification.}
	We further introduce the challenging benchmark setting for  MovieNet character identification.
	See details in Sec.~\ref{subsec:exp_pi}.
	\item \textbf{Scene Segmentation.}
	The scene segmentation task is a boundary detection task for cutting
	the movie by scene. The details about feature extraction, baseline models
	and evaluation protocols will be introduced in Sec.~\ref{subsec:exp_sseg}.
	\item \textbf{Action Recognition.} 
	We present the task of multi-label action classification task on MovieNet
	with the details of baseline models and experimental results. See Sec.~\ref{subsec:exp_action}.
	\item \textbf{Place Recognition.} 
    Similarly, we present the task of multi-label place classification task on MovieNet. See Sec.~\ref{subsec:exp_place}.
	\item \textbf{Story Understanding.} 
	For story understanding, we leverage the benchmark MovieNet segment retrieval to
	explore the potential of overall analytics using different aspects of MovieNet.
	The experimental settings and results will be found in Sec.~\ref{subsec:exp_msr}.
\end{itemize}

(4) To manage all the data and provide support for all the benchmarks, we build up a codebase for
managing MovieNet with handy processing tools. 
Besides the codes for the benchmarks, we would also release this toolbox, the features of
this tool box are introduced in Sec.~\ref{sec:tool}

\setcounter{table}{0} 
\setcounter{figure}{0} 

\section{Data in MovieNet}
\label{sec:data}

MovieNet contains various kinds of data from multiple modalities and high-quality annotations on different aspects for movie understanding. They are introduced in detail below.
And for comparison, the overall comparison of the data in MovieNet with other related dataset are shown in Tab.~\ref{tab:overall_compare}.

\subsection{Meta Data}
\label{subsec:metadata}

\begin{table}[t]
	\centering
	\caption{Comparison between MovieNet and related datasets in terms of data.}
	\begin{tabular}{c|c|c|c|c|c|c|c|c|c|c}
		\hline
		\multicolumn{1}{l|}{}  & movie & trailer & photo  & meta & genre & script & synop. & subtitle & plot & AD  \\ \hline
		MovieScope~\cite{cascante2019moviescope}             & -        & 5,027      & 5,027     & 5,027   & 13K       & -         & -           & -           & 5,027        & -      \\ \hline
		MovieQA~\cite{tapaswi2016movieqa}                & 140      & -          & -         & -       & -         & -         & -           & 408         & 408          & -      \\ \hline
		LSMDC~\cite{rohrbach2015dataset}                  & 200       & -          & -         & -       & -         & 50        & -           & -           & -            & 186     \\ \hline
		MovieGraphs~\cite{vicol2018moviegraphs}             & 51       & -          & -         & -       & -         & -         & -           & -           & -            & -      \\ \hline
		AVA~\cite{gu2018ava}             & 430       & -          & -         & -       & -         & -         & -           & -           & -            & -      \\ \hline
		MovieNet               & 1,100    & 60K     & 3.9M & 375K &  805K     & 986      & 31K      &  5,388   & 46K   & -      \\ \hline
	\end{tabular}
	\label{tab:overall_compare}
\end{table}

\begin{figure}[t]
	\centering
	\includegraphics[width=\linewidth]{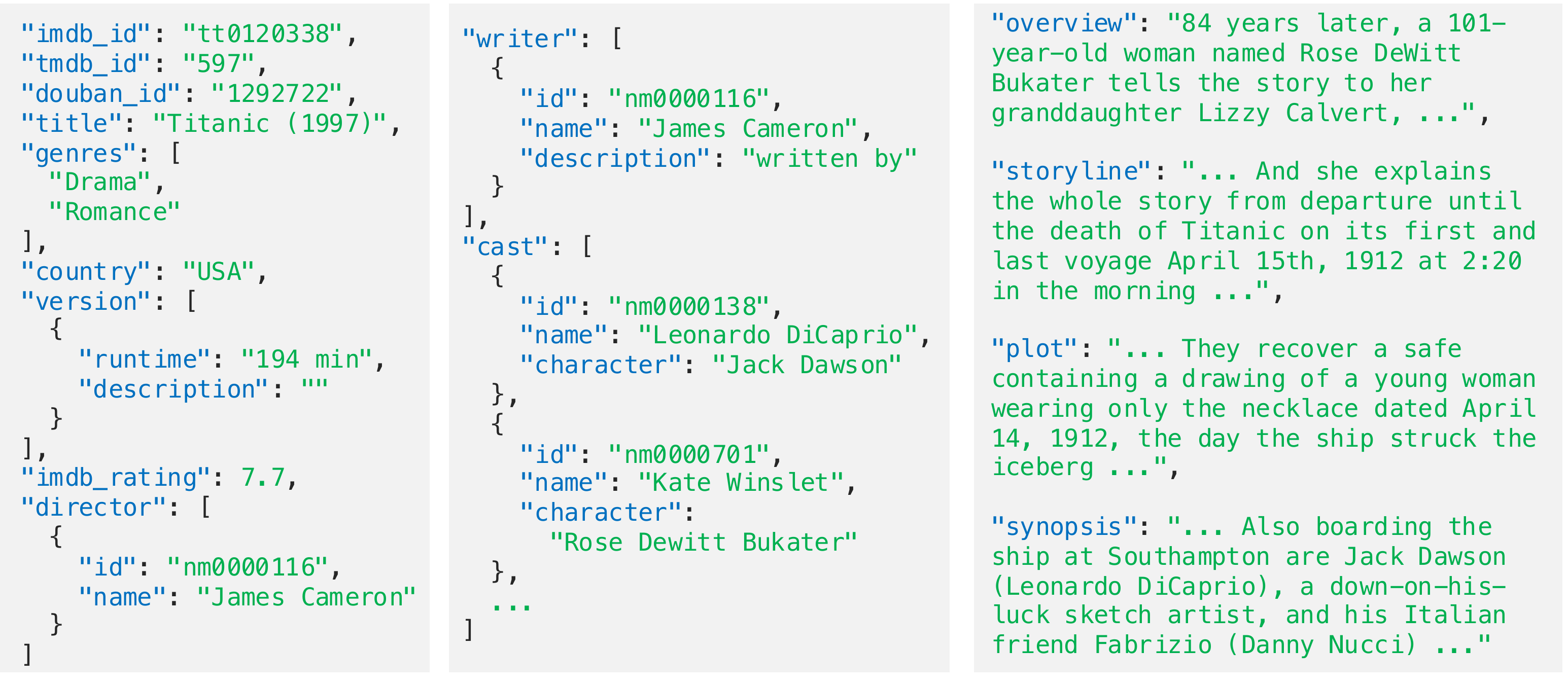}
	\caption{\small
		A sample of metadata from the movie \emph{Titanic}.
	}
	\label{fig:meta_example}
\end{figure}

MovieNet contains meta data of $375K$ movies.
Note that the number of metadata is significantly large than the movies provided with video sources (\ie $1,100$)
because we belief that metadata itself can support various of tasks.
It is also worth noting that the metadata of all the $1,100$ selected movies are included in this metadata set.
 Fig.~\ref{fig:meta_example} shows a sample of the meta data, which is from \emph{Titanic}. 
 More details of each item in the meta data would be introduced below.

\begin{figure}[ht]
	\centering
	\includegraphics[width=\linewidth]{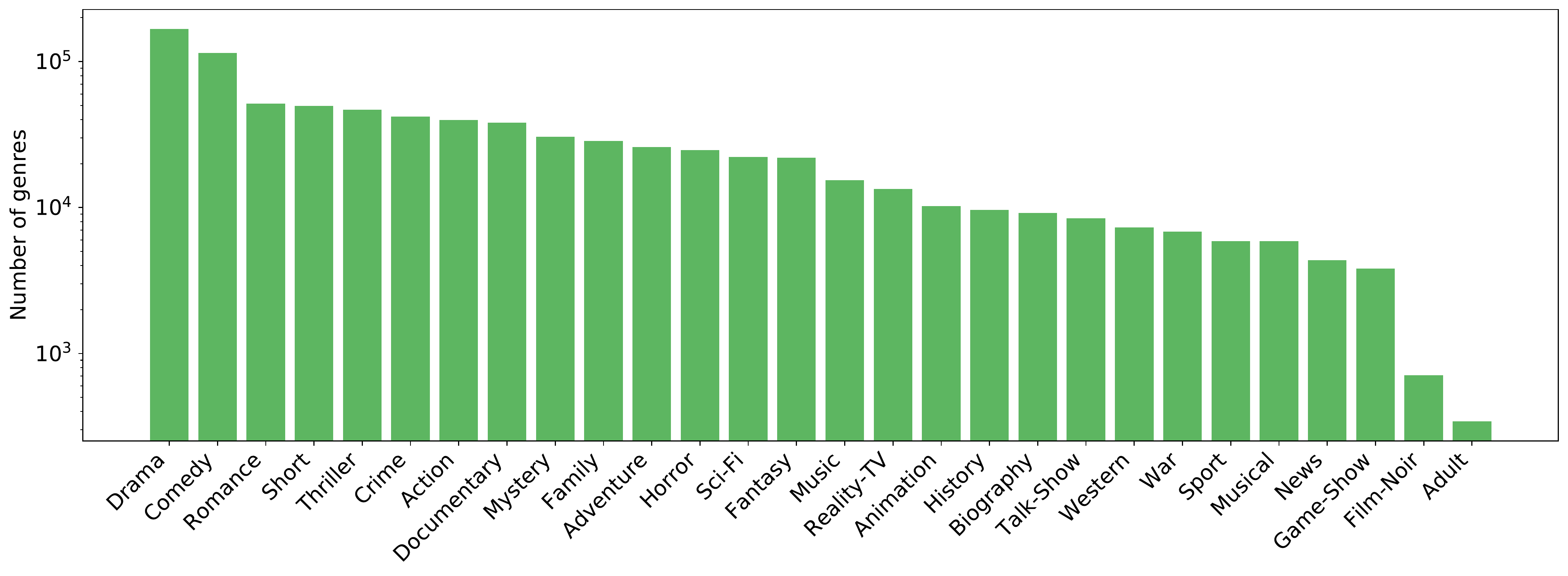}
	\caption{\small
		Statistics of genres in metadata. It shows the number of genres for each genre category (y-axis in log-scale).
	}
	\label{fig:meta_genres}
\end{figure}

\begin{figure}[ht]
	\centering
	\includegraphics[width=\linewidth]{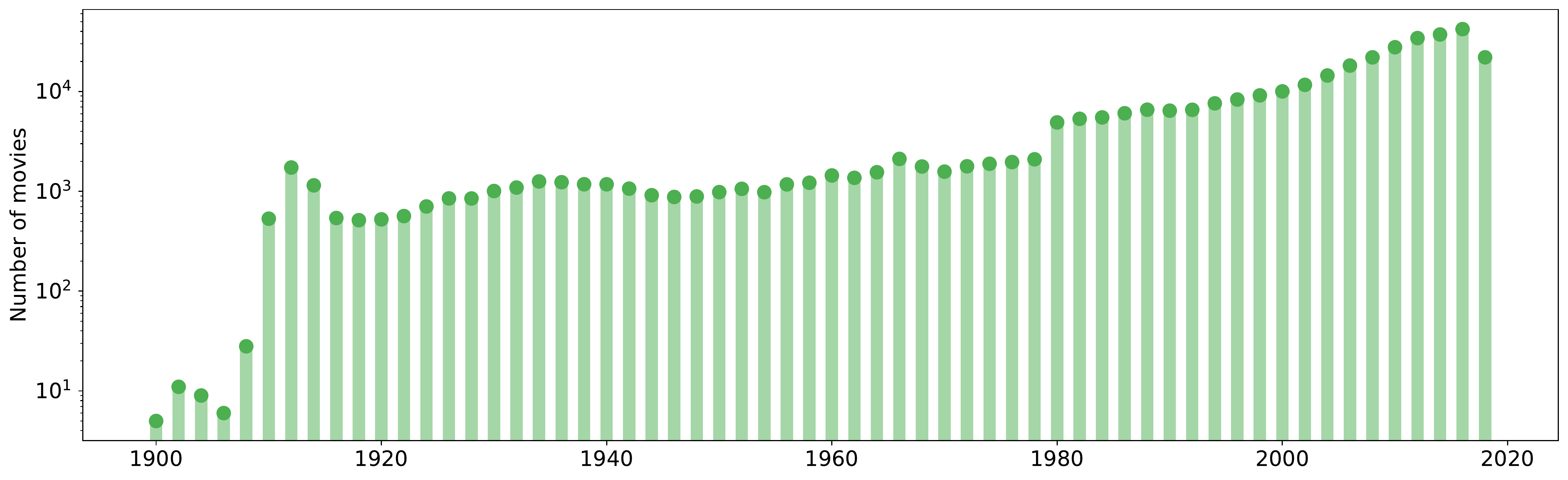}
	\caption{\small
		Distribution of release date of the movies in metadata. It shows the number of movies in each year (y-axis in log-scale).
		Note that the number of movies generally increases as time goes by.
	}
	\label{fig:meta_year}
\end{figure}

\begin{figure}[ht]
	\centering
	\includegraphics[width=\linewidth]{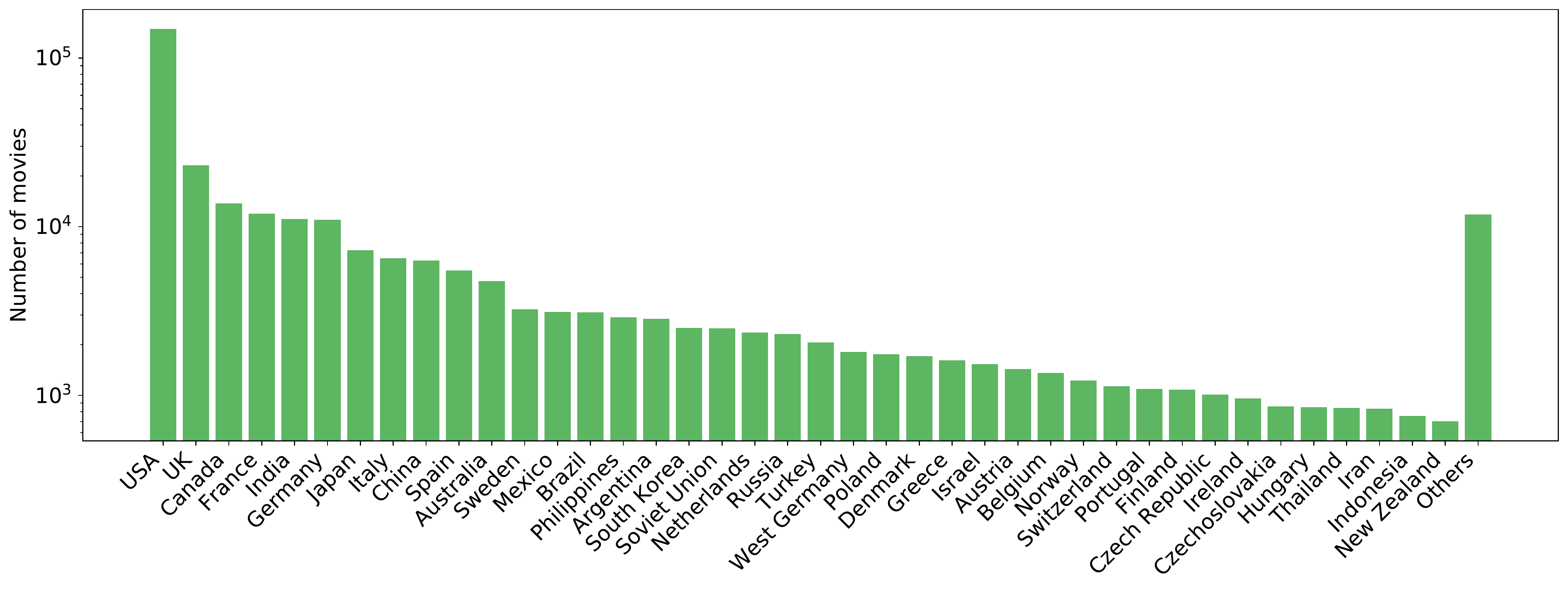}
	\caption{\small
		The countries that the movies belong to in metadata. Here we show top 40 countries with the left as ``Others''.
		The number of movies (y-axis) is in log-scale. 
	}
	\label{fig:meta_country}
\end{figure}

\begin{figure}[ht]
	\centering
	\includegraphics[width=\linewidth]{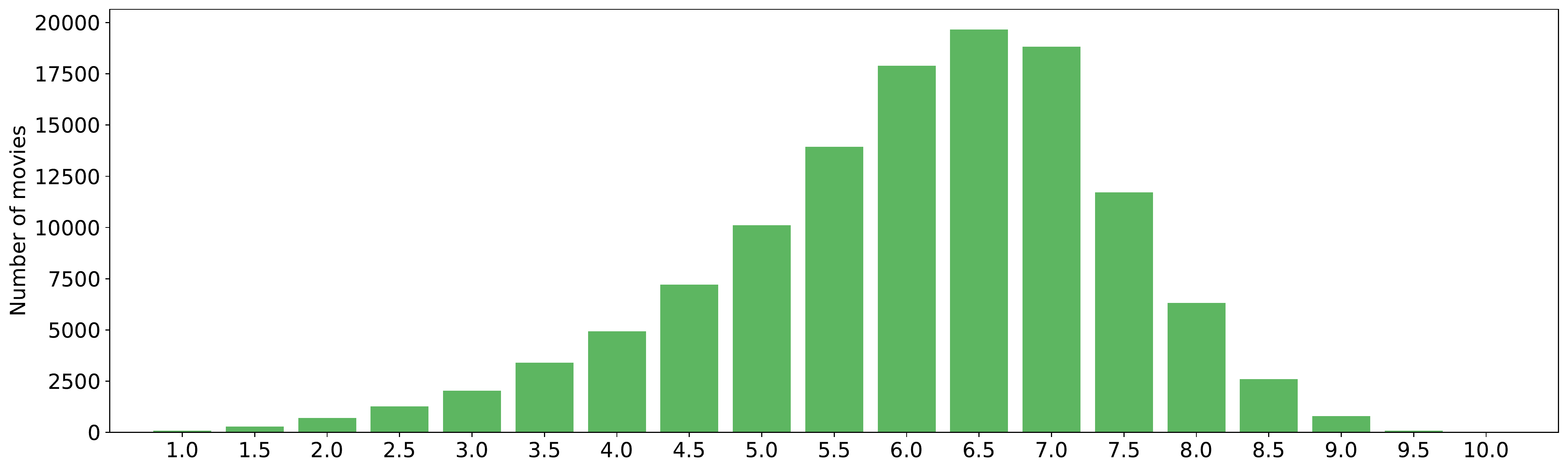}
	\caption{\small
		Distribution of score ratings in MovieNet metadata.
	}
	\label{fig:meta_rating}
\end{figure}

\begin{itemize}
	\item \textbf{IMDb ID.}
	IMDb ID is the ID of a movie in the IMDb website\footnote{https://www.imdb.com/}. IMDb ID is usually a string begins with ``tt'' and follows with $7$ or $8$ digital numbers, \eg ``tt0120338'' for the movie \emph{Titanic}. One can easily get some information of a movie from IDMb with its ID. For example, the homepage of \emph{Titanic} is ``https://www.imdb.com/title/tt0120338/''. The IMDb ID is also taken as the ID of a movie in MovieNet.
	\item \textbf{TMDb ID.}
	TMDb ID is the ID of a movie in the TMDb website\footnote{https://www.themoviedb.org/}. We find that some of the content in TMDb is of higher-quality than IMDb. For example, TMDb provides different versions of trailers and higher resolution posters. Therefore, we take it as a supplement of IMDb. TMDb provides APIs for users to search for information. With the TMDb ID provided in MovieNet, one can easily get more information if needed.
	\item \textbf{Douban ID.}
	Douban ID is the ID of a movie in Douban Movie\footnote{https://movie.douban.com/}. We find that for some Asian movies, such as those from China and Japan, IMDb and TMDb contains few information. Therefore, we turn to a Chinese movie website, namely Douban Movie, for more information of Asian movies. We also provide Douban ID for some of the movies in MovieNet for convenience.
	\item \textbf{Version.}
	For movie with over one versions,
	\eg normal version, director's cut,
	we provide runtime and description of each version to help researchers align the annotations
	with their own resources.
	\item \textbf{Title.}
	The title of a movie following the format of IMDb, \eg, \emph{Titanic (1997)}.
	\item \textbf{Genres.}
	Genre is a category basedon similarities either in the narrative elements or in the emotional response to the movie, \eg, comedy, drama. There are totally $28$ unique genres from the movies in MovieNet. Fig.~\ref{fig:meta_genres} shows the distribution of the genres.
	\item \textbf{Release Date.}
	Release Date is the date when the movie published. Fig~\ref{fig:meta_year} shows the number of the movies released every year, from which we can see that the number of movies continuously grows every year.
	\item \textbf{Country.}
	Country refers to the country where the movie produced. The top-40 countries of the movies in MovieNet are shown in Fig.~\ref{fig:meta_country}.
	\item \textbf{Version.}
	A movie may have multiple versions, \eg, director's cut, special edition. And different versions would have different runtimes. Here we provide the runtimes and descriptions of the movies in MovieNet.
	\item \textbf{IMDb Rating.}
	IMDb rating is the rating of the movie uploaded by the users.
	The distribution of different ratings are shown in Fig.~\ref{fig:meta_rating}.
	\item \textbf{Director.}
	Director contains the director's name and ID.
	\item \textbf{Writer.}
	Writer contains the writer's name and ID.
	\item \textbf{Cast.}
	A list of the cast in the movie, each of which contains the actor/actress's name, ID and character's name.
	\item \textbf{Overview.}
	Overview is a brief introduction of the movie, which usually covers the background and main characters of the movie.
	\item \textbf{Storyline.}
	Storyline is a plot summary of the movie. It is longer and contains more details than the overview.
	\item \textbf{Wiki Plot.}
	Wiki Plot is the summary of the movie from Wikipedia and
	is usually longer than overview and storyline.
\end{itemize}


%
%

\begin{figure}[t]
	\centering
	\begin{minipage}[t]{\linewidth}\centering
		\subfloat[Duration (in seconds).]{
		\includegraphics[width=\linewidth]{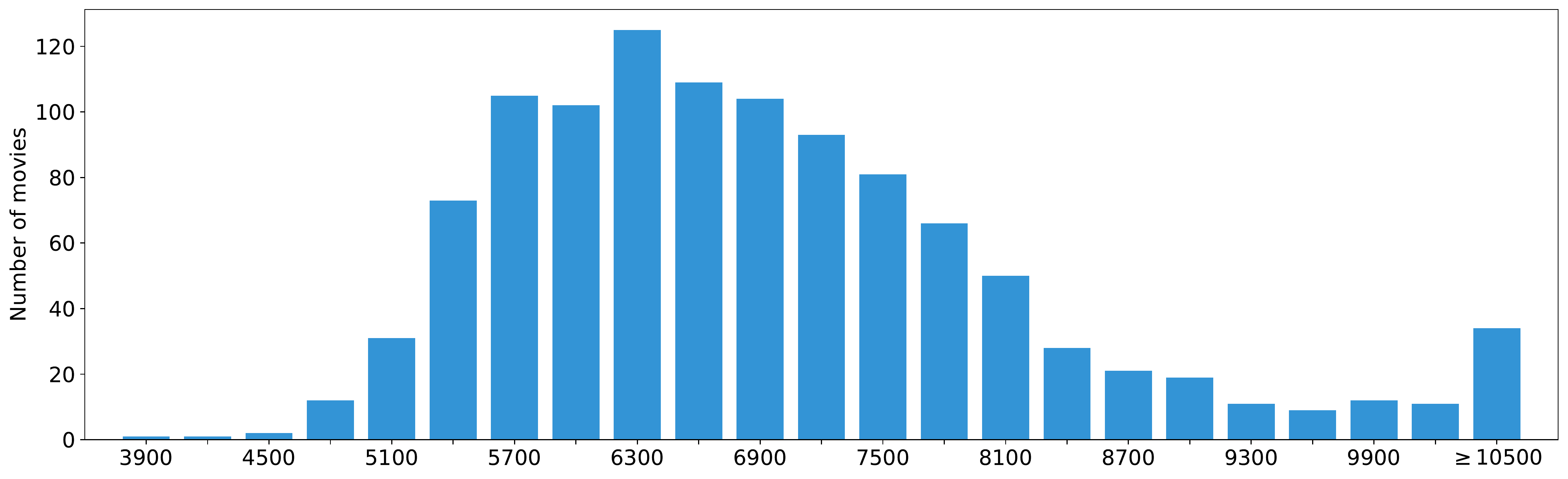}
	}
	\end{minipage}
	\quad
	\begin{minipage}[t]{\linewidth}
		\centering
		\subfloat[Number of shots.]{
		\includegraphics[width=\linewidth]{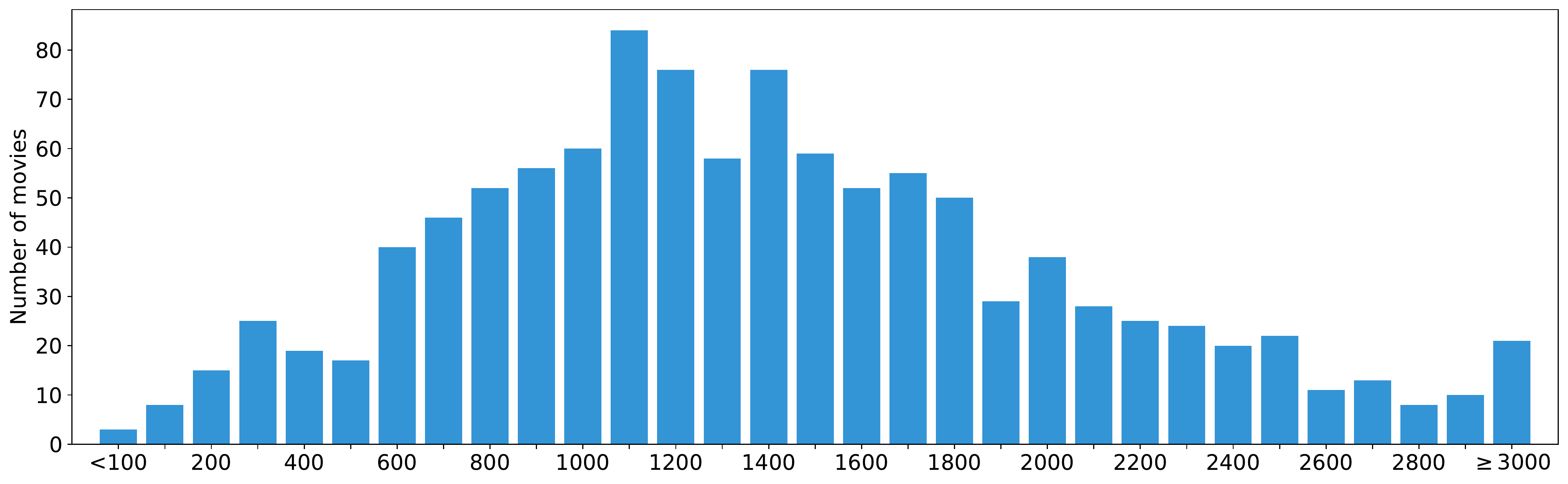}
	}
	\end{minipage}
	\caption{Distribution of duration and number of shots for the $1,100$ movies in MovieNet.}
	\label{fig:meta_movie_stat}
\end{figure}

\begin{figure}[t]
	\centering
	\begin{minipage}[t]{\linewidth}
		\centering
		\subfloat[Year.]{
		\includegraphics[width=\linewidth]{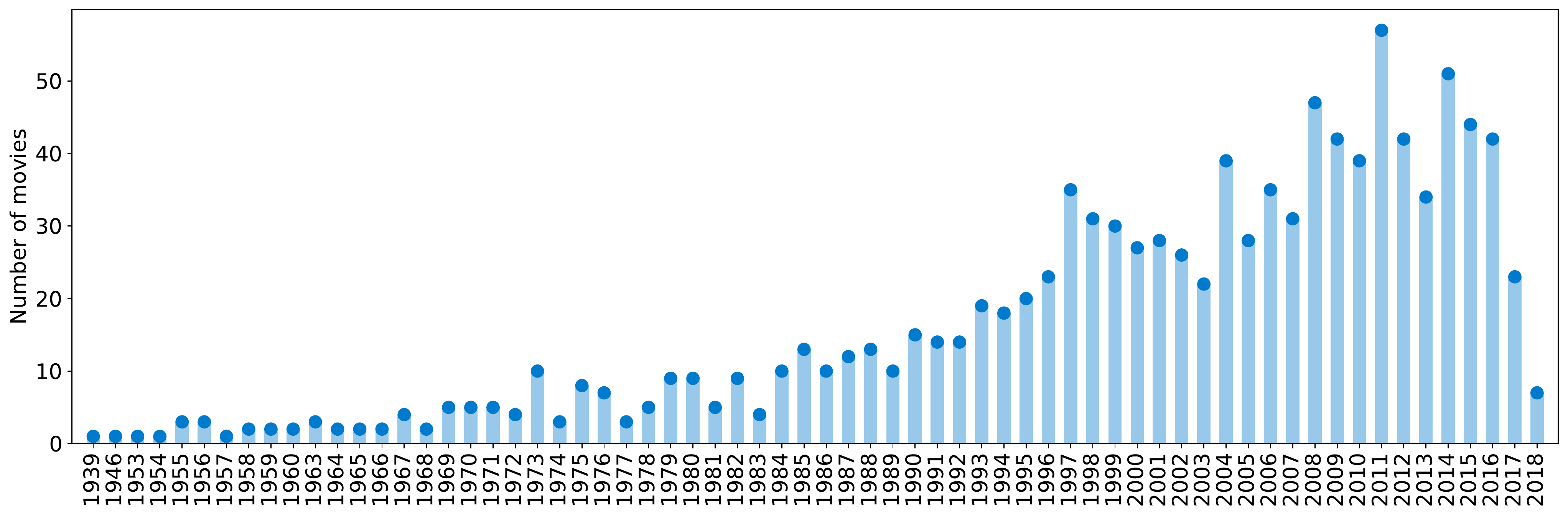}}
	\end{minipage}
	
	\begin{minipage}[t]{0.64\linewidth}
		\centering
		\subfloat[Country.]{
		\includegraphics[width=\linewidth]{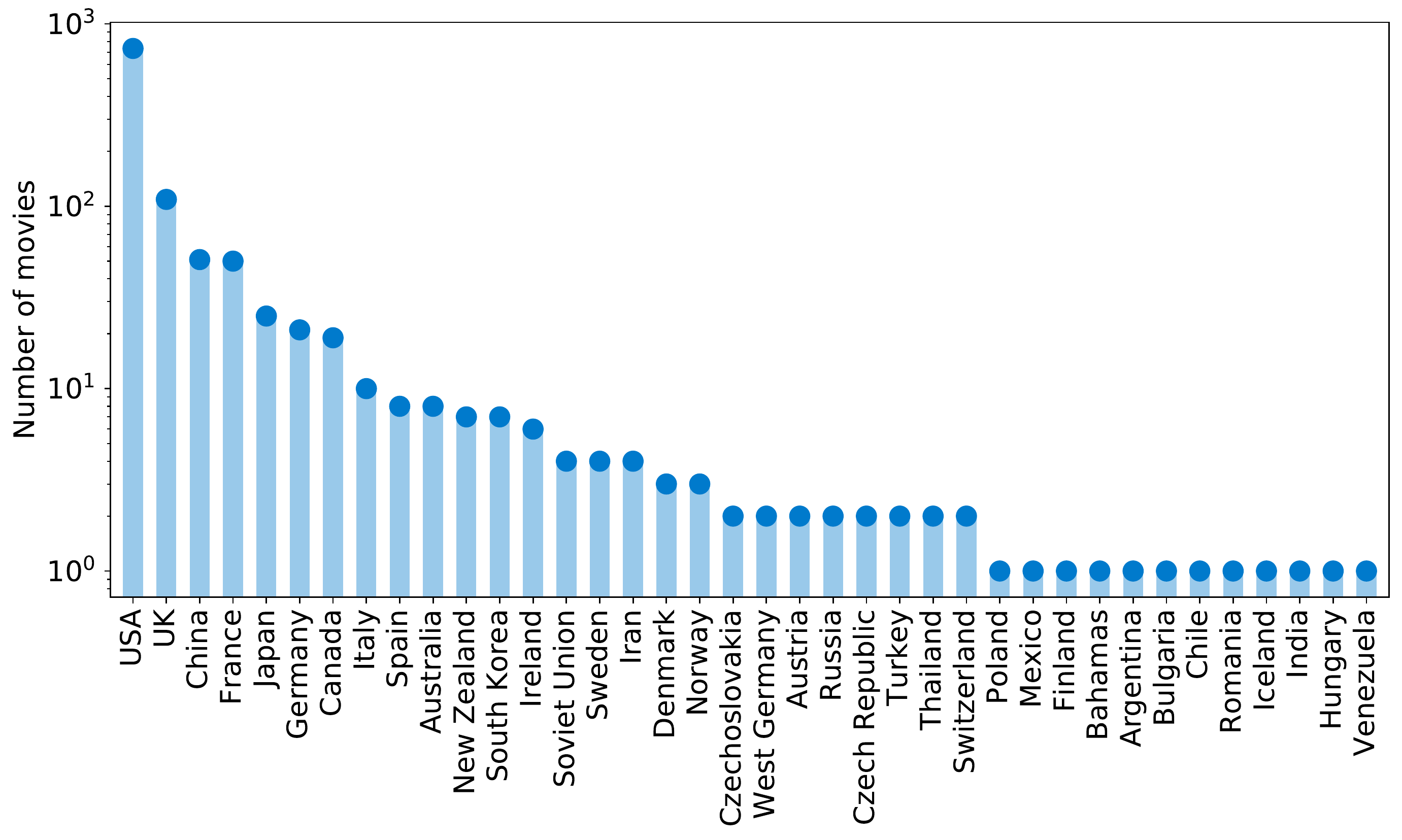}}
	\end{minipage}
	\begin{minipage}[t]{0.35\linewidth}
		\centering
		\subfloat[Genre.]{
		\includegraphics[width=\linewidth]{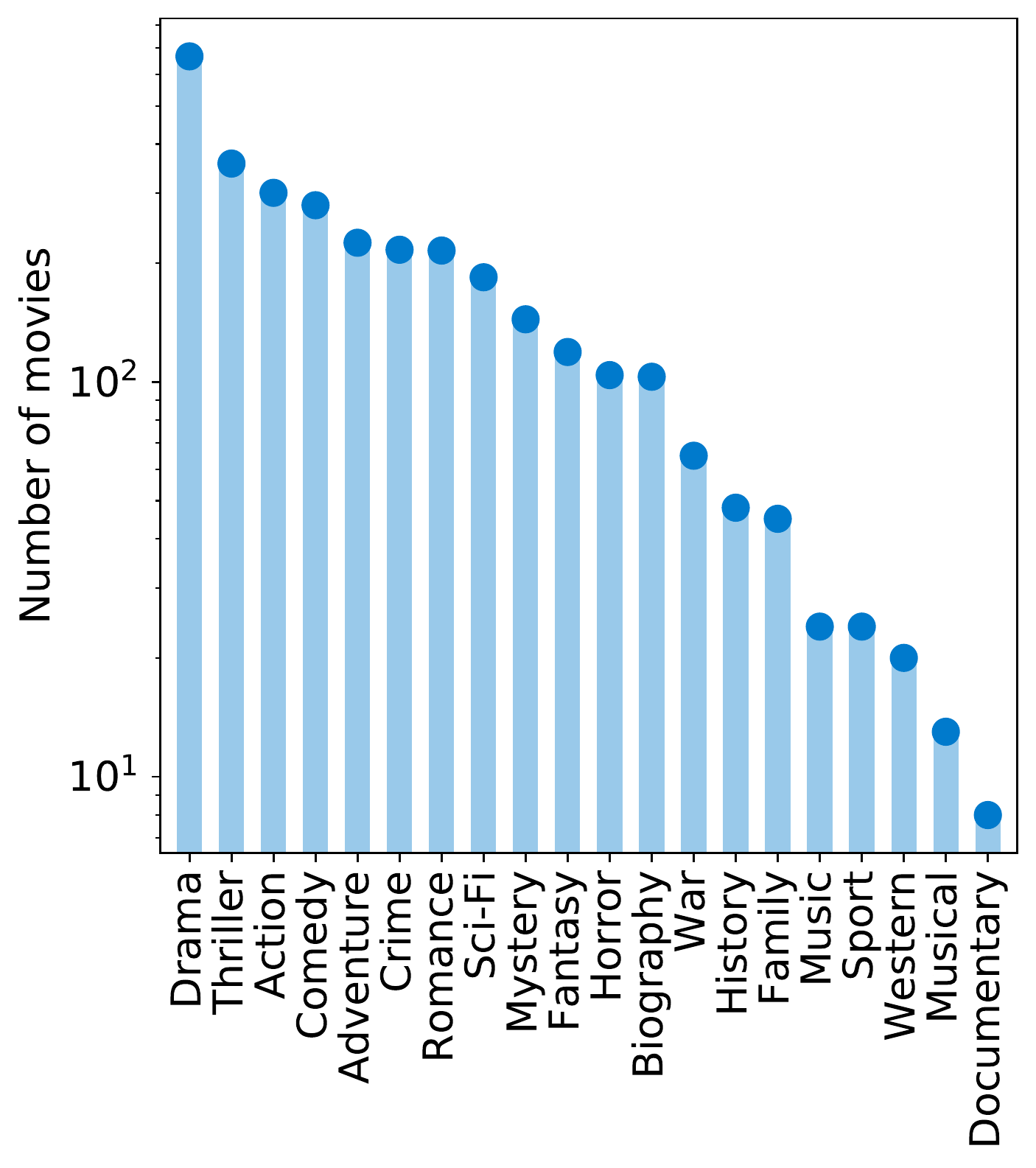}}
	\end{minipage}
	\caption{Distribution of release year, countries and genres for the $1,100$ movies in MovieNet (y-axis of country and genre in log scale).}
	\label{fig:meta_movie_diversity}
\end{figure}

\subsection{Movie}
\label{subsec:movie}
As we introduced in our paper, there are $1,100$ movies in MovieNet. Here we show some statistics of the $1,100$ movies in Fig.~\ref{fig:meta_movie_stat}, including the distributions of runtime and the shot number.
As mentioned in Sec.~\ref{subsec:metadata}, in addition to these $1,100$ movies, we also provided
metadata for other movies as much as we can. This also apply for other data like trailer and photo, 
and we would not clarify it in the next sections.  

It is mentioned in the paper that we select the movie that covers a wide range of years, countries and genres.
The distribution of these data are shown in Fig.~\ref{fig:meta_movie_diversity}.
We can see that the movies are diversity in terms of year, country and genre.

\noindent\textbf{Feature Representation.}
To play with a long video is nontrivial for the current
deep learning framework and computational power. For
the convenience of research, we propose multiply ways of feature representations for a movie.
\begin{itemize}
	\item \textbf{Shot-based visual feature.}
	For most of the task, \eg genre classification, shot-based representation is an efficient representation.
	A shot is a series of frames that runs for an uninterrupted period of time, which can be taken as the smallest visual unit of a movie. So we use shot-based representation for movies in our MovieNet. Specifically, we first separate each movie into shots with a shot detection tool~\cite{sidiropoulos2011temporal}. Then, we sample three key frames and extract visual features using
    models pre-trained on ImageNet.
	\item \textbf{Audio feature.}
	For each shot, we also cut the audio wave within this shot and then extract audio feature~\cite{chung2019naver} as the supplementary of visual feature.
	\item \textbf{Frame-based feature.}
	For those tasks like action recognition that need consider motion information, we also provide frame-based
	feature for these tasks. 
\end{itemize}

\subsection{Subtitle}
\label{subsec:subtitle}
For each movie from MovieNet, we provide an English subtitle that aligned with the movie.
It is often the case that the downloaded subtitle is not aligned with the video
because usually a movie has multiple versions, \eg \emph{director's cut} and \emph{extended}.
To make sure that each subtitle is aligned with the video source, before
manually checking, we make the following efforts:
(1) For the subtitle extracted from original video or downloaded from
the Internet, we first make sure the subtitles are complete and are English version (by applying regular expression).
(2) Then we clean the subtitle by removing noise such as HTML tags.
(3) We leverage the off-the-shelf tool\footnote{https://github.com/smacke/subsync} that transfers audio to text and matches text with the subtitle to produce a shift time.
(4) We filtered out those subtitles with a shift time surpasses a particular threshold
and download another subtitle.  
After that, we manually download the subtitles that are still problematic,
and then repeat the above steps.

Particularly, the threshold in step(4) is set to 60s by the following observations:
(1) Most of the aligned subtitles have a shift within 1 seconds.
(2) Some special cases, for example, a long scene without any dialog, would cause the tool to generate 
a shift of a few seconds. But the shift is usually less than 60s.
(3) The subtitles that do not align with the original movies are usually either another version or crawled from another movie.
In such cases, the shift will be larger than 60s.

After that, we ask annotators to manually check if the auto aligned subtitles are still misaligned.
It turns out that the auto alignment are quite effective that few of the subtitles are still problematic.

\begin{figure}[t]
	\centering
	\begin{minipage}[t]{\linewidth}
		\centering
		\subfloat[Duration (in seconds).]{
		\includegraphics[width=\linewidth]{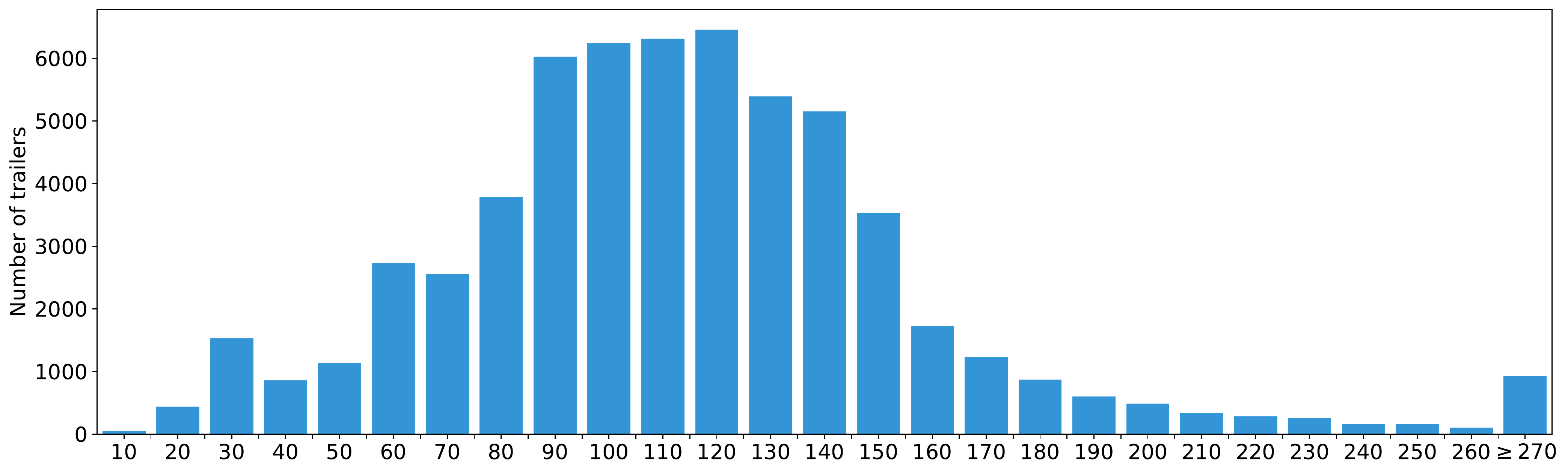}}
	\end{minipage}
	\quad
	\begin{minipage}[t]{\linewidth}
		\centering
		\subfloat[Number of shots.]{
		\includegraphics[width=\linewidth]{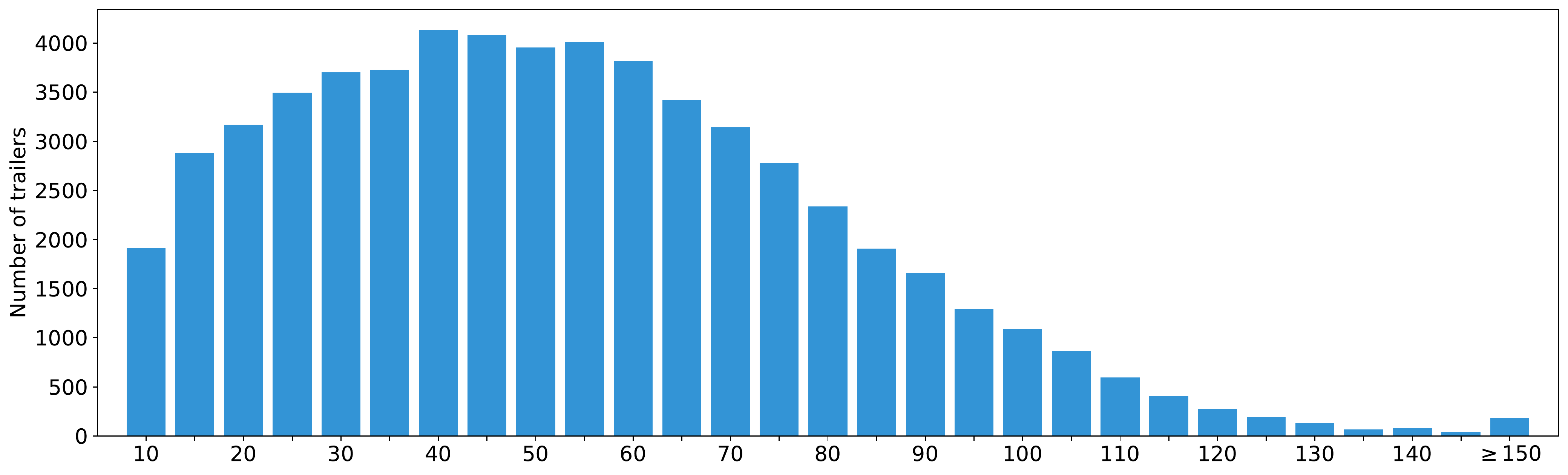}}
	\end{minipage}
	\caption{Distribution of duration and number of shots for the trailers in MovieNet.}
	\label{fig:meta_trailer_stat}
\end{figure}

\subsection{Trailer}
\label{subsec:trailer}
There are $60K$ trailers from $33K$ unique movies in MovieNet. The statistics of the trailers are shown in Fig.~\ref{fig:meta_trailer_stat}, including the distributions of runtime and shot number.

Besides some attractive clips from the movie, which we name as content-shots, trailers usually contains exract shots to show some important information, \eg, the name of the director, release date, \etc. We name these shots as \emph{info-shots}. Info-shots are quite different from other shots since they contain less visual content. For most of the tasks with trailers, we usually focus on content-shots only. Therefore, it is necessary for us to distinguish info-shots and content-shots. We develop a simple approach to tackling this problem.

Given a shot, we first use a scene text detector~\cite{tian2016detecting} to detect the text on each frame. Then we generate a binary map of each frame, where the areas covered by the text bounding boxes are set to $1$ and others are set to $0$.
Then we average all the binary maps of a shot and get a heat map.
By average the heat map we get an overall score $s$ to indicate how much text detected in a shot. The shot whose score is higher than a threshold $\alpha$ and a the average contrast is lower than $\beta$ is taken as a info-shot in MovieNet. Here we take the contrast into consideration by the observation that info-shots usually have simple backgrounds.

\begin{figure}[t]
	\centering
	\includegraphics[width=\linewidth]{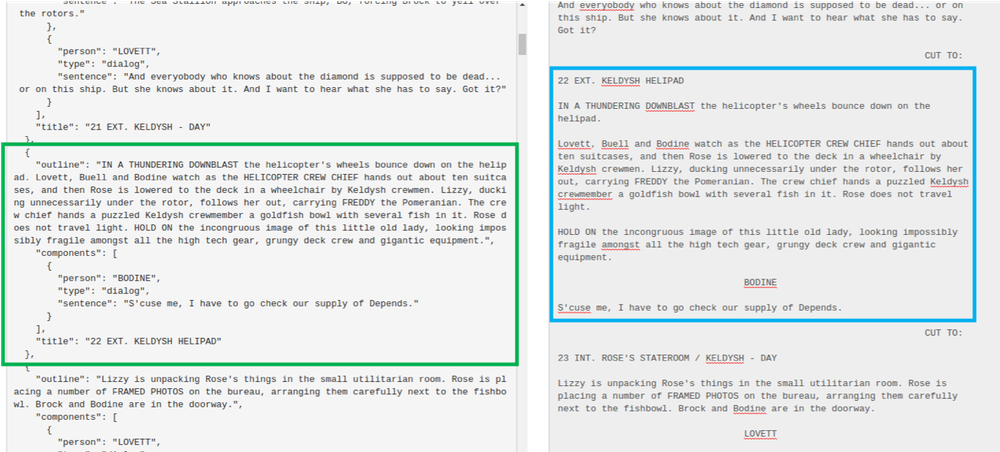}
	\caption{
		Here we show an example of the parsed script. The left block shows the formatted script snippet and the right block shows the 
		corresponding raw ones of \emph{``Titanic''}.
	}
	\label{fig:script_align_tool}
\end{figure}

\subsection{Script}
\label{subsec:script}
We provide aligned scripts in MovieNet. Here we introduce the details of script alignment.
As mentioned in the paper, we align the movie script to movies 
by automatically matching the dialog with subtitles. This process is introduced below.

Particularly, a movie script is a written work by filmmakers narrating the storyline and dialogs. It is useful for tasks like movie summarization. To obtain the data for these tasks, we need to align scripts to the movie timelines. 

In the preprocessing stage, we develop a script parsing algorithm
using regular expression matching to format a script as a list of scene cells, where scene cell denotes for the combination of a storyline snippet and a dialog snippet for a specific event. 
An example is shown in Fig.~\ref{fig:script_align_tool}.
To align each storyline snippet to the movie timeline, we choose to connect dialog snippet to subtitle first. To be specific, we formulate script-timeline alignment problem as an optimization problem for dialog-subtitle alignment. The idea comes from the observation that dialog is designed as the outline of subtitle.

\begin{algorithm}[t]
	\caption{Script Alignment}
	\label{alg:script_align}
	\begin{algorithmic}
		\INPUT $  \mS \in \mathbb{R}^{M \times N} $
		\State $ R \gets Array(N) $
		\State $ val \gets Matrix(M, N) $
		\State $ inds \gets Matrix(M, N) $
		\For{$ col \gets 0, N-1 $}
		\For{$ row \gets 0, M-1 $}
		\State $ a \gets val[row, col-1] + \mS[row, col] $
		\State $ b \gets val[row-1, col] $
		\If{$ a > b $}
		\State $inds[row, col] \gets row$
		\State $val[row,col] \gets a$
		\Else
		\State $inds[row, col] \gets inds[row-1, col]$
		\State $val[row,col] \gets b$
		\EndIf
		\EndFor
		\EndFor
		\State $index \gets M-1$
		\For{$col \gets N-1, 0$}
		\State $index \gets inds[index, col] $
		\State $R \gets index$
		\EndFor
		\OUTPUT $ R $
	\end{algorithmic}
\end{algorithm}

Let $dig_i$ denote the dialog snippet in $i^{th}$ scene cell, $sub_j$ denote the $j^{th}$ subtitle sentence. We use TF-IDF~\cite{ramos2003using} to extract text feature for dialog snippet and subtitle sentence. Let $f_i=\text{TF-IDF}(dig_i)$ denote the TF-IDF feature vector of $i^{th}$ dialog snippet and $g_j=\text{TF-IDF}(sub_j)$ denote that of $j^{th}$ subtitle sentence. For all the $M$ dialog snippets and $N$ subtitle sentences, the similarity matrix $\mS$ is given by

$$s_{i,j}=\mS(i,j)=\frac{f_i^Tg_j}{|f_i||g_j|}$$

For $j^{th}$ subtitle sentence, we assume the index of matched dialog snippet $i_j$ should be smaller than $i_{j+1}$, which is the index of matched dialog for ${(j+1)}^{th}$ subtitle sentence. By taking this assumption into account, we formulate the dialog-subtitle alignment problem as the following optimization problem,

\begin{equation}\label{eq:opt}
\begin{aligned}
& \underset{i_j}{\text{max}}
& & \sum_{j=0}^{N-1}s_{i_j,j} \\
& \text{s.t.}
& & 0 \leq i_{j-1}\leq i_j \leq i_{j+1} \leq M-1.
\end{aligned}
\end{equation}

This can be effectively solved by dynamic programming algorithm. Let $L(p,q)$ denote the optimal value for the above optimization problem with $\mS$ replaced by its submatrix $\mS[0, \ldots, p, 0, \ldots, q]$. The following equation holds, 

$$L(i,j)=\max\{L(i,j-1), L(i-1,j)+s_{i,j}\}$$

\begin{figure}[t]
	\centering
	\includegraphics[width=\linewidth]{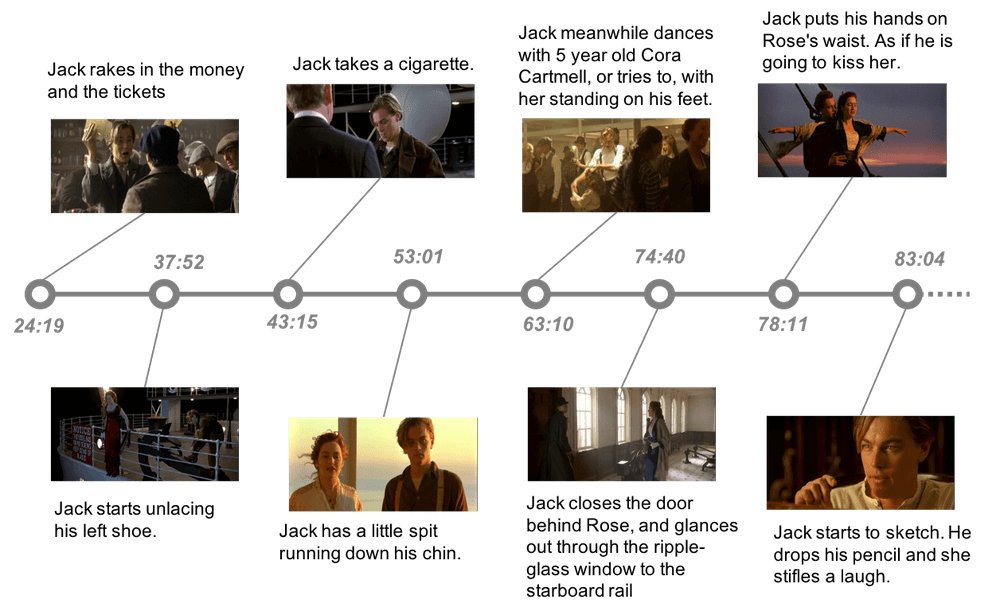}
	\caption{\small
		Qualitative result of script alignment. The example comes from the movie \emph{Titanic}. Each node marked by a timestamp is associated with a matched storyline snippet and a snapshot image.
	}
	\label{fig:script_align_show}
\end{figure}

It can be seen that the optimal value of the original problem is given by $L(M-1, N-1)$. To get the optimal solution, we apply the dynamic programing algorithm shown in Alg.~\ref{alg:script_align}. 
Once we obtain the connection between a dialog snippet and a subtitle sentence, we can directly assign the timestamp of the subtitle sentence to the script snippet who comes from the same scene cell as the dialog snippet.
Fig.~\ref{fig:script_align_show} shows the qualitative result of script alignment. It illustrates that our algorithm is able to draw the connection between storyline and timeline even without human assistance.

\subsection{Synopsis}
\label{subsec:synopsis}
\begin{figure}[t]
	\centering
	\includegraphics[width=\linewidth]{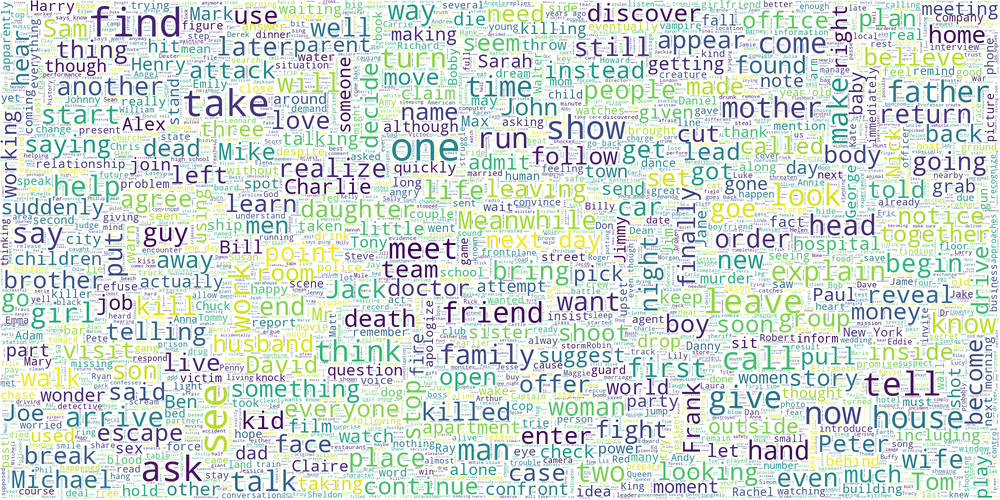}
	\caption{
		Word cloud of synopsis corpus in MovieNet.
	}
	\label{fig:wordcount}
\end{figure}

Here we show some statistics of synopsis in Tab.~\ref{tab:plot_vs_syn}
and wordcloud visualization in Fig.~\ref{fig:wordcount}. Compared to the wiki plot, we can see that synopsis is a higher-quality textual source which contains richer content and longer descriptions.

\begin{table}[t]
	\centering
	\caption{Comparison on the statistics of wiki plot with that of synopsis.}
	\label{tab:plot_vs_syn}
	\begin{tabular}{c|ccc}
		\hline
		& ~~\# sentence/movie~~ & ~~\# word/sentence~~ & ~~\# word/movie~~ \\ \hline
		wiki plot & 26.2             & 23.6            & 618.6           \\
		synopsis  & 98.4             & 20.4            & 2004.7          \\ \hline
	\end{tabular}
\end{table}

\begin{figure}[!t]
	\centering
	\includegraphics[width=0.5\linewidth]{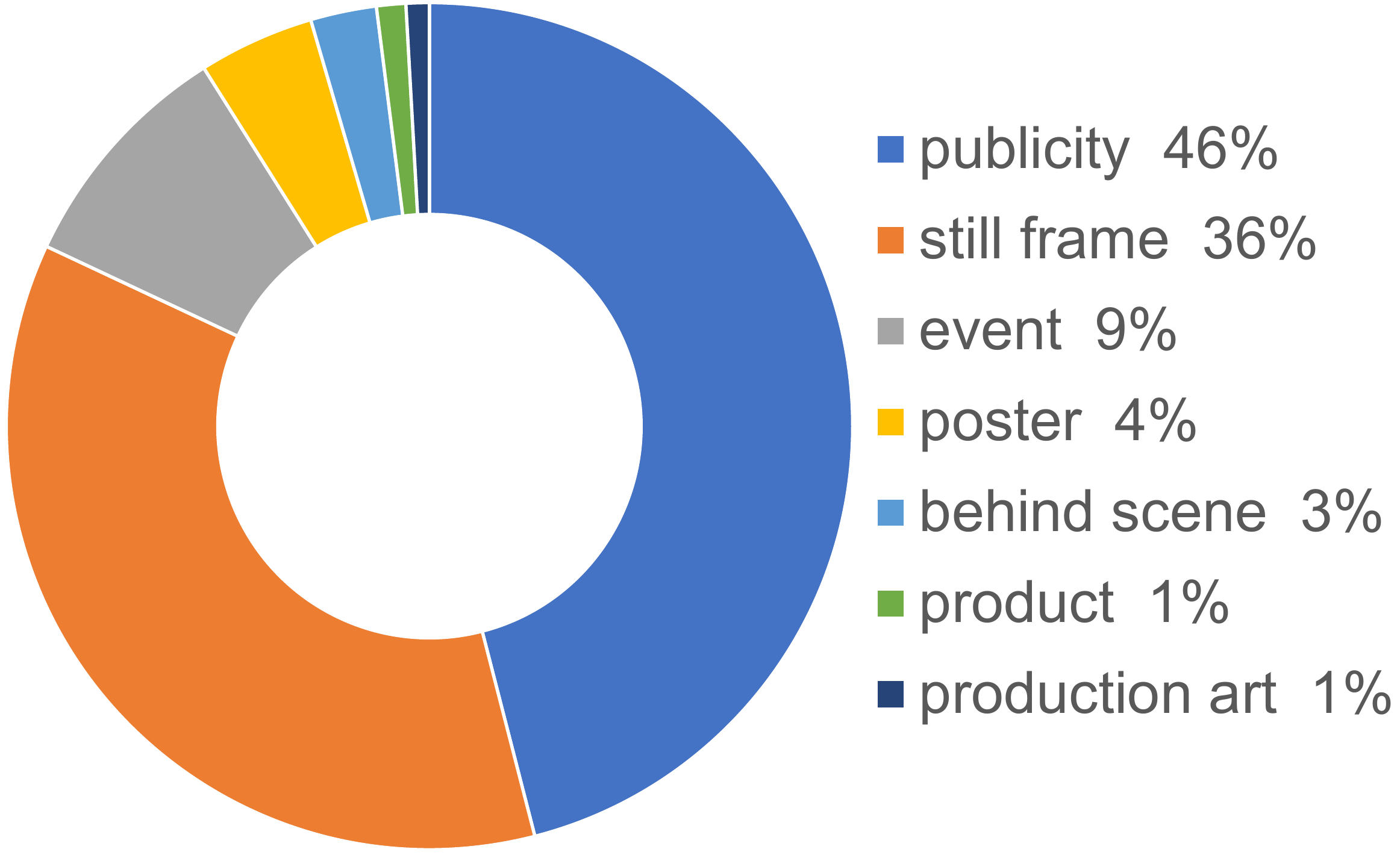}
	\caption{\small
		Percentage of different photo types in MovieNet.
	}
	\label{fig:phototype}
\end{figure}

\begin{figure}[!ht]
	\centering
	\includegraphics[width=\linewidth]{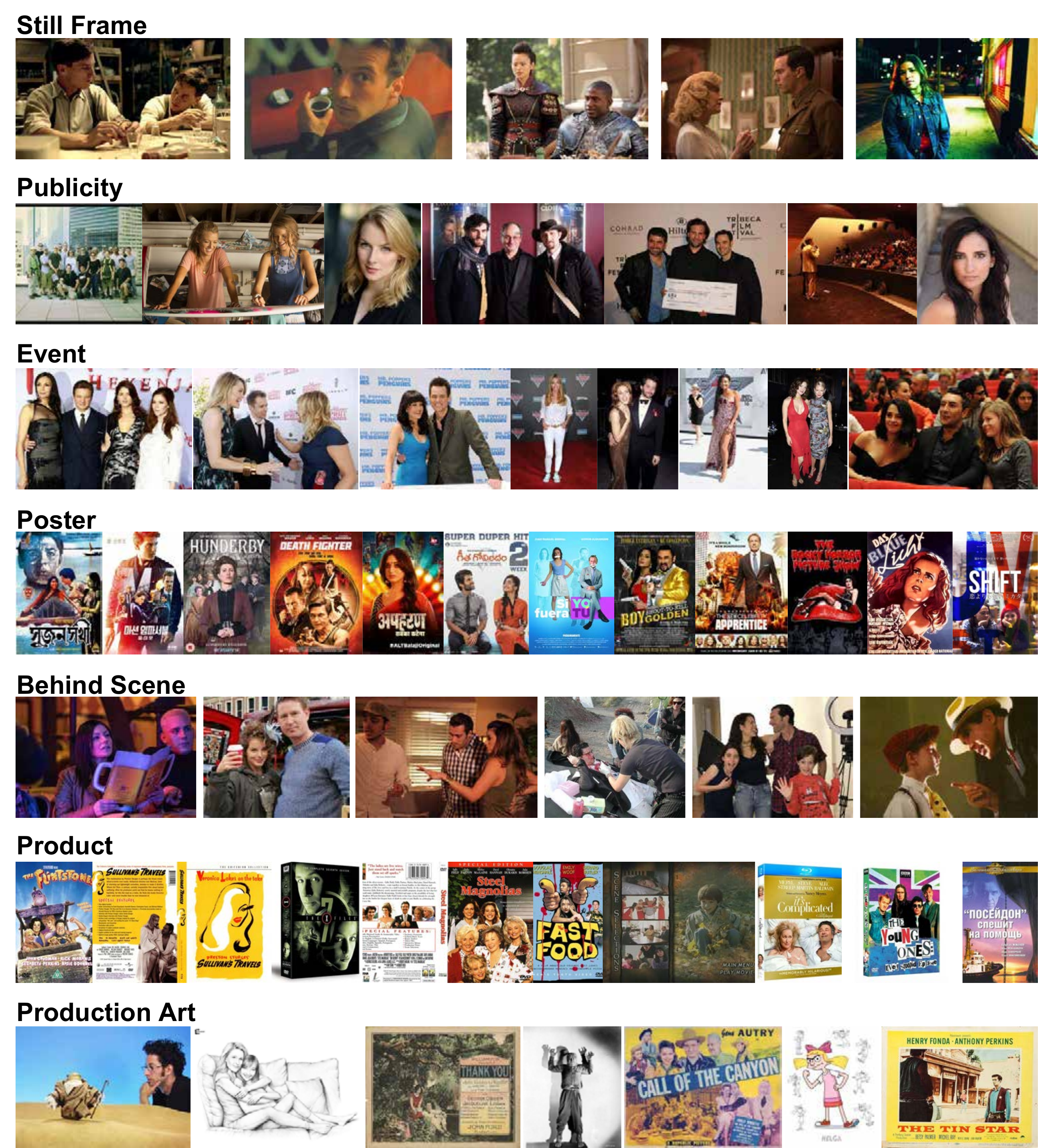}
	\caption{\small
		Samples for different types of photos in MovieNet.
	}
	\label{fig:sample_photo}
\end{figure}

\subsection{Photo}
\label{subsec:photo}
As we introduced in the paper, there are $3.9M$ photos from $7$ types in MovieNet. Here we show the percentage of each type in Fig.~\ref{fig:phototype}. Also some samples are shown in Fig.~\ref{fig:sample_photo}.

\setcounter{table}{0} 
\setcounter{figure}{0} 
\section{Annotation in MovieNet}
\label{sec:annotation}

\begin{figure}[!t]
	\vspace{3pt}
	\centering
	\includegraphics[width=0.84\linewidth]{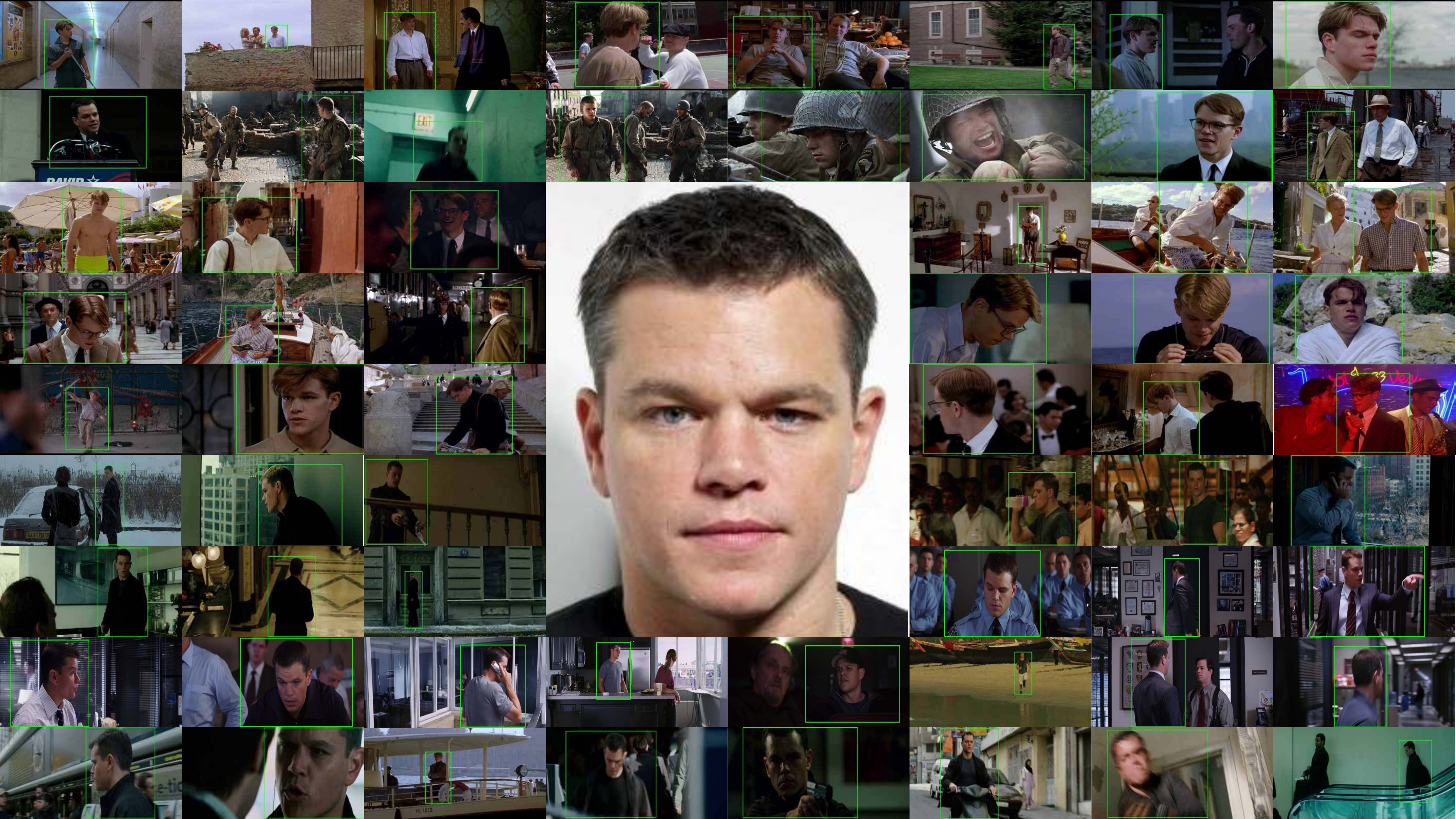} \\
	\vspace{3pt}
	\includegraphics[width=0.84\linewidth]{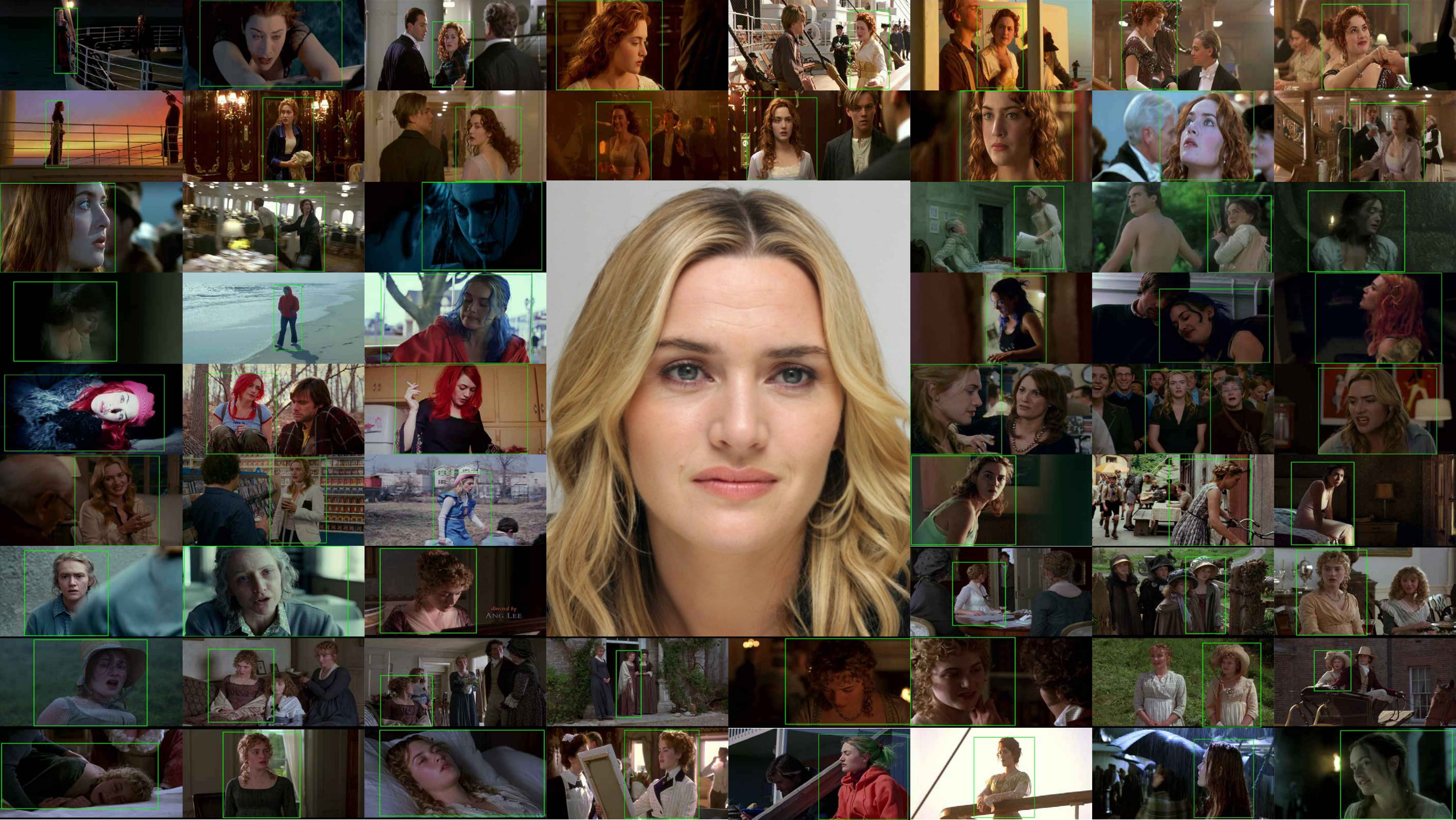} \\
	\vspace{3pt}
	\includegraphics[width=0.84\linewidth]{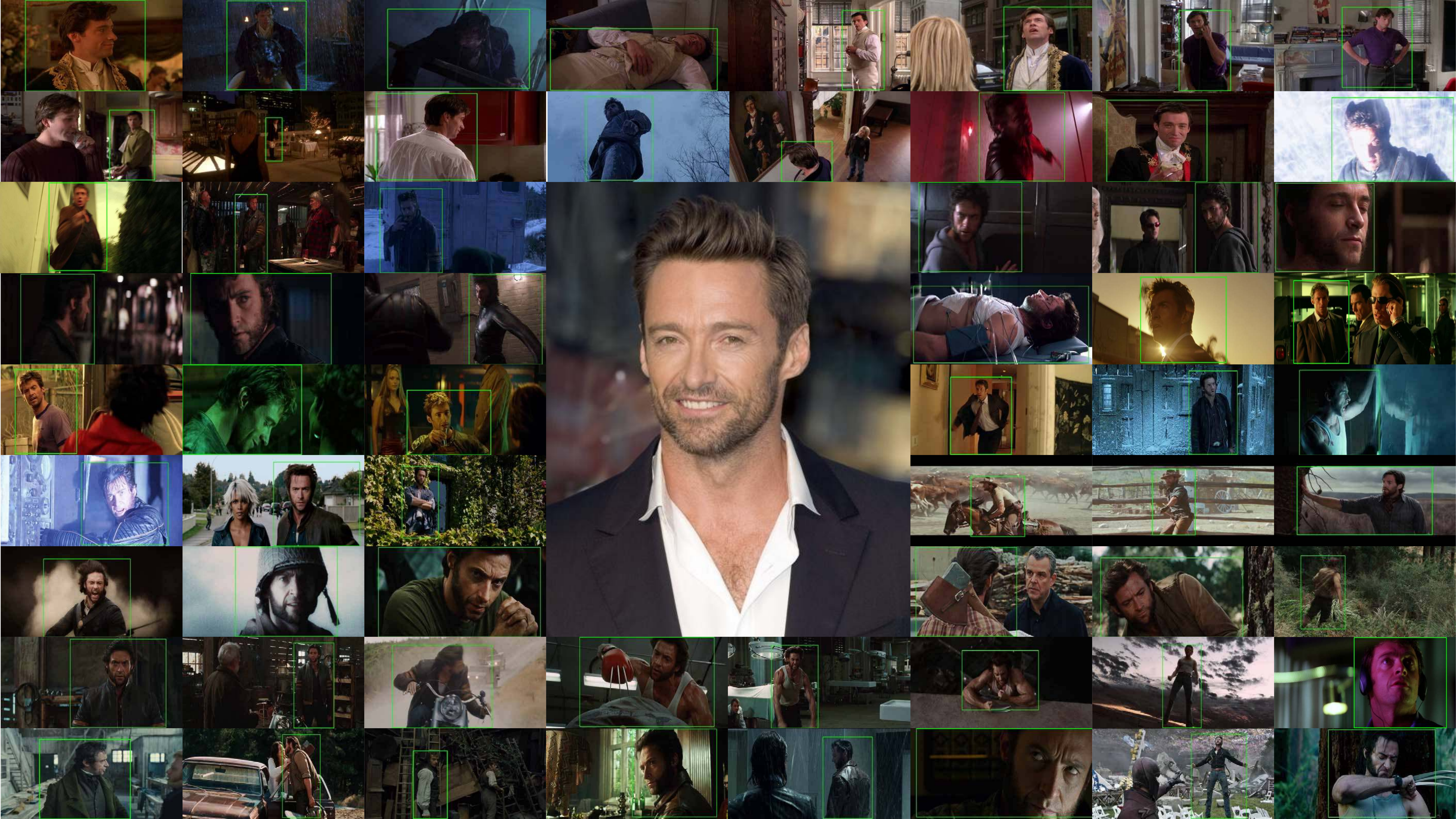}
	\caption{\small
		Samples of the character annotations in MovieNet with portrait in the center.
	}
	\label{fig:sample_person}
	\vspace{-0pt}
\end{figure}

To achieve high-quality annotations, we have made great effort on designing the workflow and labeling interface, the details of which would be introduced below.

\subsection{Character Bounding Box and Identity}
\label{subsec:person_anno}

\begin{figure}[!t]
	\vspace{-0pt}
	\begin{center}
		\includegraphics[width=0.9\linewidth]{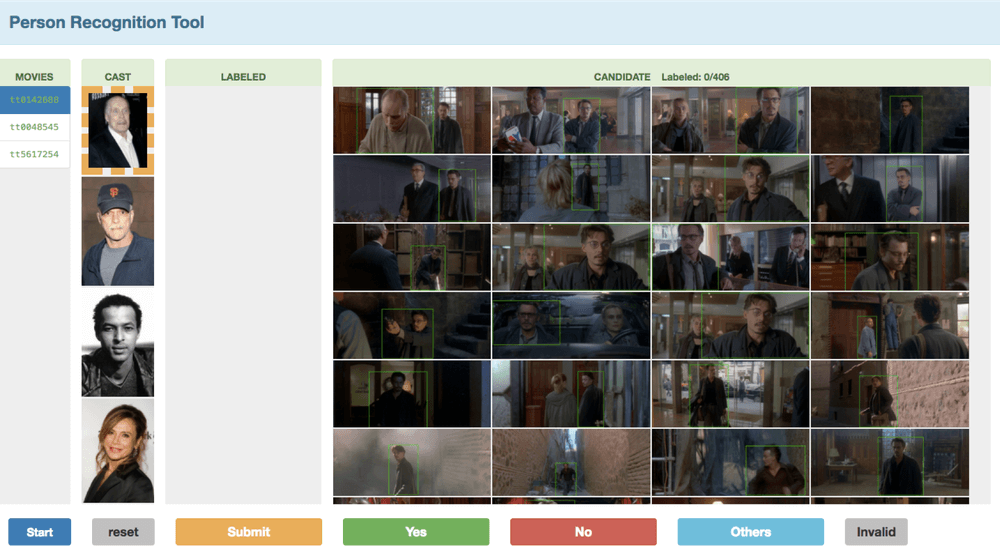}
	\end{center}
	\vspace{-10pt}
	\caption{\small
		Annotation interface of Character Identity (Stage 1). From left to right, they are (1) the movie list, (2) the cast list of the selected movie shown by their portraits, (3) the labeled samples annotated as the selected cast, which would be helpful for annotating more hard samples, and (4) the candidates of the selected cast, which is generated by our algorithm considering both face feature and body feature. The annotator can label positive samples (by clicking ``Yes'') or negative samples (by clicking ``No''). After several iterations, when they are familiar to all the cast in the movie, they can label the characters belong the credit cast list by clicking ``Others''.
	}
	\label{fig:inter_person}
	\vspace{-0pt}
\end{figure}


\begin{figure}[t]
	\centering
	\begin{minipage}[t]{\linewidth}
		\centering
		\includegraphics[width=\linewidth]{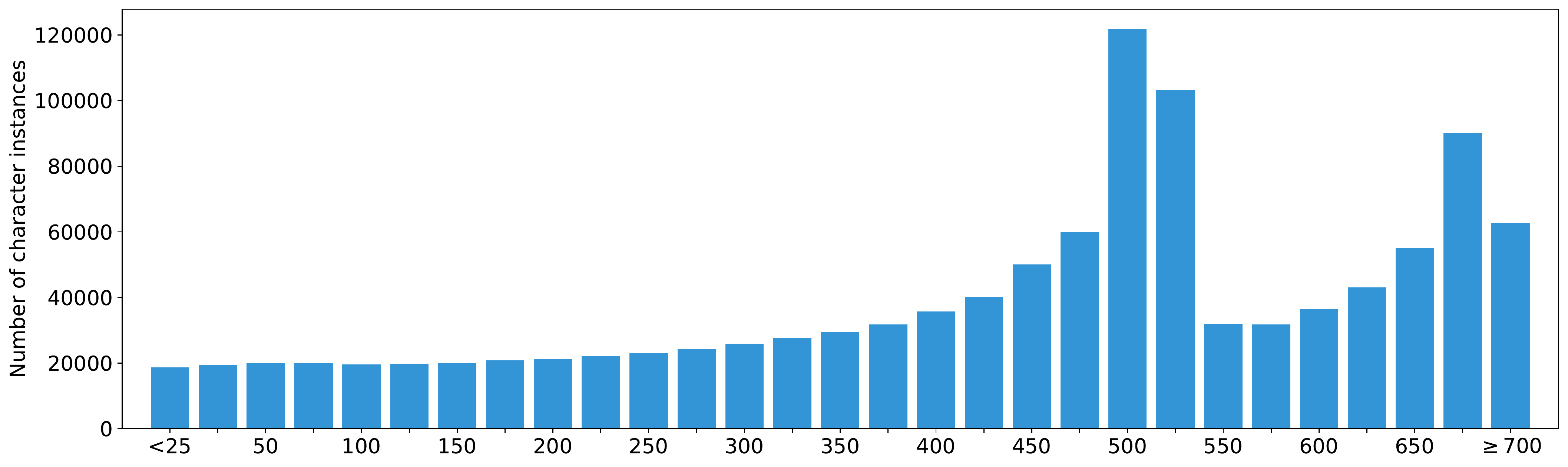}
		\caption{Height of character instance (pixel)}
	\end{minipage}
	\quad
	\begin{minipage}[t]{\linewidth}
		\centering
		\includegraphics[width=\linewidth]{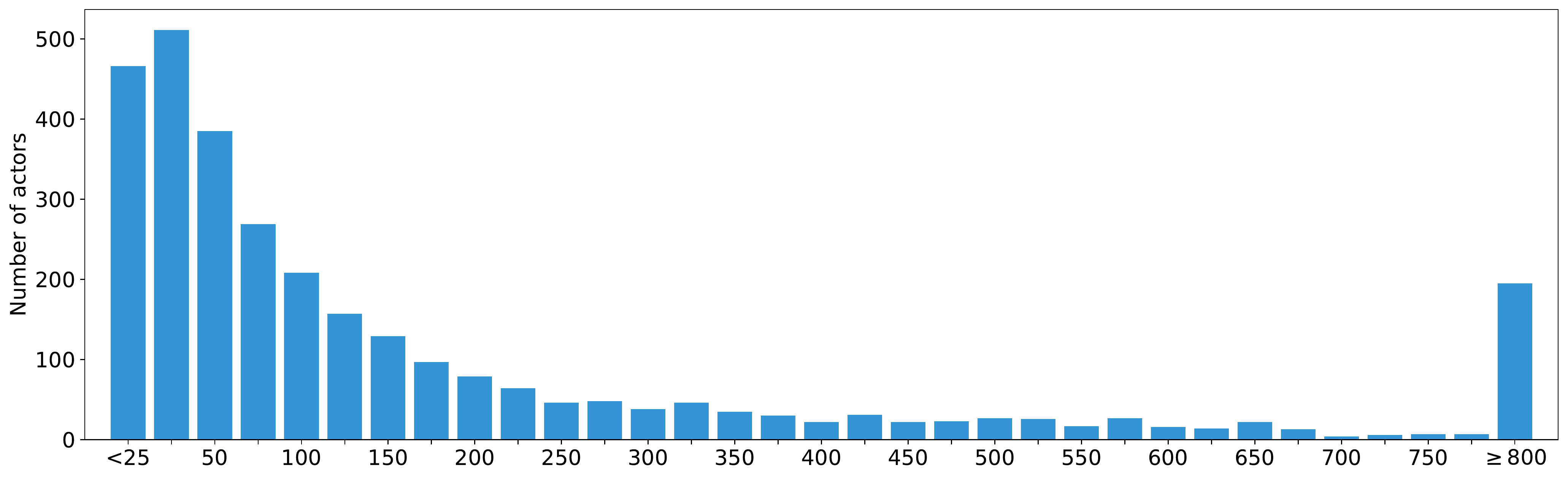}
		\caption{Number of character instance}
	\end{minipage}
	\caption{Statistics of character bounding box and identity annotation,
		including the height of character instance and the number of character instance.}
	\label{fig:meta_person_stat}
\end{figure}

\noindent\textbf{Workflow and Interface.}
Annotation of the character bounding box and identity follows six steps.
(1) We first randomly choose $758K$ key frames from the provided movies. Here the key frames are extracted by average sampling three frames each shot. Then the annotators are asked to annotate the bounding box of the characters in the frames, after which we get $1.3M$ character bounding boxes.
(2) With the $1.3M$ character bounding boxes, we train a character detector. Specifically, the detector is a Cascade R-CNN~\cite{cai2018cascade} with feature pyramid~\cite{lin2017feature}, using a ResNet-101~\cite{he2016deep} as backbone. We find that the detector can achieve a $95\%$ mAP.
(3) Since the identities in different frames within one shot are usually duplicated, we choose only one frame from each shot to annotate the identities of the characters. We apply the detector to the key frames  for identity annotation. Since the detetor performs good enough, we only manually clean the false positive boxes in this step. Resulting in $1.1M$ instances.
(4) To annotate the identities in a movie is a challenging task due to the large variance in visual appearances. We develop a semi-automatic system for the first step of identity annotation to reduce cost. We first get the portrait of each cast from IMDb or TMDb, some of which are shown in Fig.~\ref{fig:sample_person}.  
(5)We then extract the face feature with a face model trained on MS1M~\cite{guo2016ms} and extract the body feature with a model trained on PIPA~\cite{zhang2015beyond}. By calculating the feature similarity of the portrait and the instances in the movie, we sort the candidate list for each cast. And the annotator is then asked to determine whether each candidate is the cast or not. Also, the candidate list would update after each annotation, which is similar to active learning. The interface is shown in Fig.~\ref{fig:inter_person}. We find that this semi-automatic system can highly reduce the annotation cost.
(6) Since the semi-automatic system may introduce some bias and noise, we further design a step for cleaning. At this step, the frames are demonstrated in time order and the annotating results at the first step are shown. The annotator can clean the results with temporal context.

\noindent\textbf{Statistics and Samples.}
Here we show some statistics of the character annotation in Fig.~\ref{fig:meta_person_stat}, including the size distribution of bounding boxes and the distribution of instance number.
From the statistics we can see that the number of character instance is a long-tail distribution. 
However, for those famous actors like \emph{Leonardo Dicaprio}, they have much more character instances than others.
 Some samples are also shown in Fig.~\ref{fig:sample_person}. We can see that MovieNet contains large-scale and diverse characters, which would be helpful for the researches on character analysis.

\begin{figure}[!t]
	\centering
	\includegraphics[width=\columnwidth]{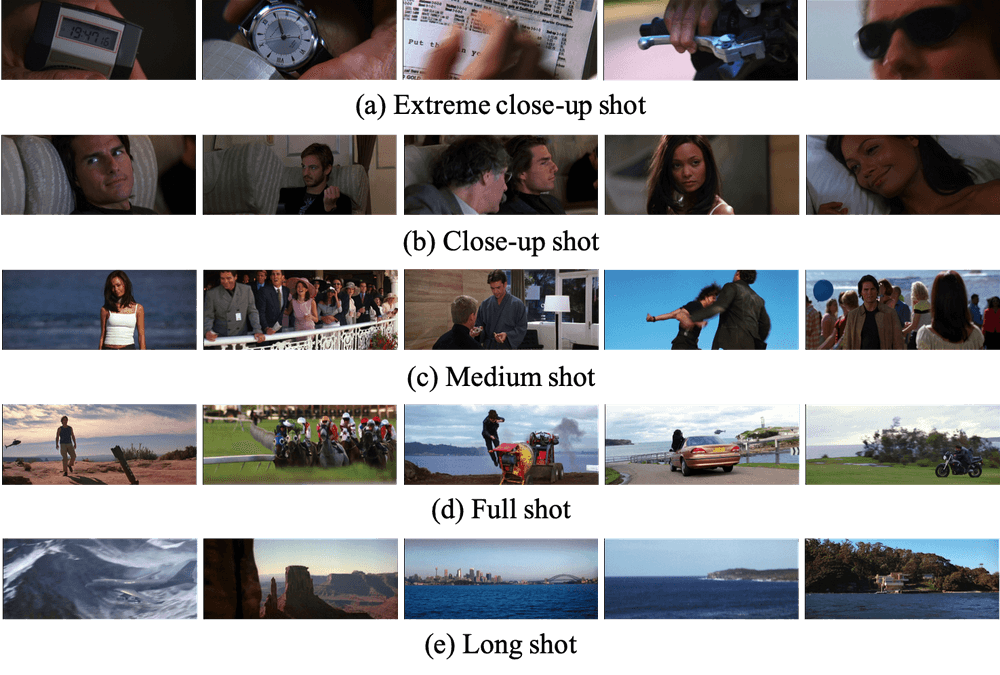}
	\vspace{-20pt}
	\caption{
		Examples of the cinematic style (scale type) of shots in \emph{Mission Impossible}. From (a) to (e), they are extreme close-up shot, close-up shot, medium shot, full shot and long shot.
	}
	\label{fig:data_csp_scale}
	\includegraphics[width=\columnwidth]{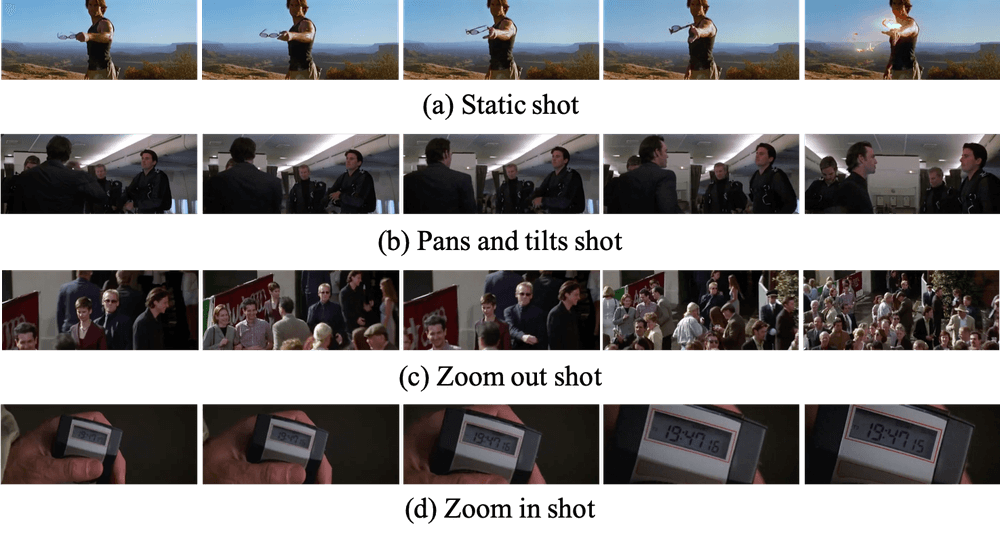}
	\vspace{-20pt}
	\caption{
		Examples of the cinematic style (movement type) of shots in \emph{Mission Impossible}. From (a) to (d), they are static shot, pans and tilts shot, zoom out shot, and zoom in shot.
	}
	\label{fig:data_csp_mov}
\end{figure}

\subsection{Cinematic Styles}
\label{subsec:cs_anno}

We annotated the commonly used two kinds of cinematic tags of a shot~\cite{giannetti1999understanding}.
Shot scale depict the portion of subject within the frames in a shot, while shot movement describe the camera movement or the lens change of a shot.

\begin{figure}[!t]
	\begin{center}
		\includegraphics[width=0.98\linewidth]{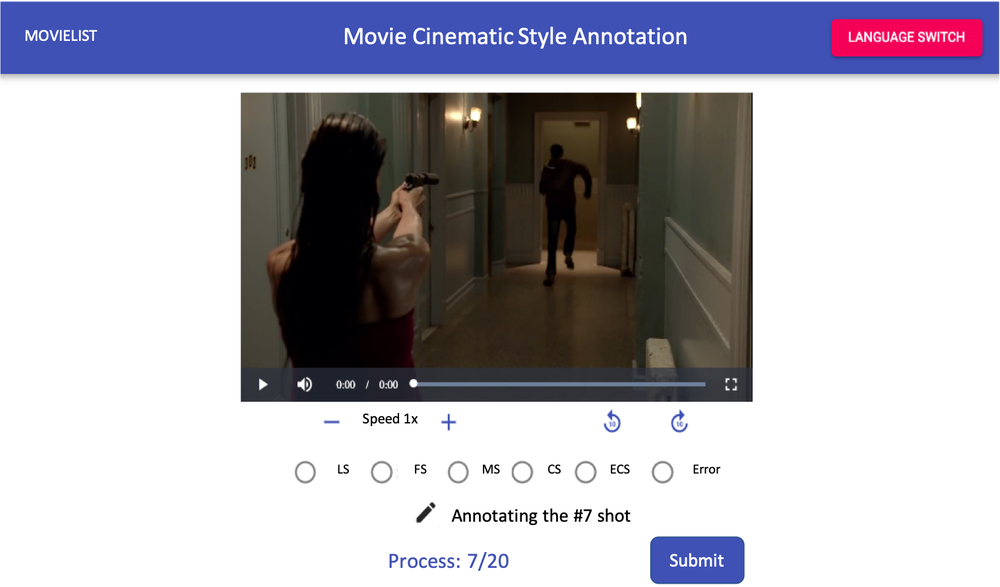}
	\end{center}
	\vspace{-20pt}
	\caption{\small
		Annotation interface of cinematic styles.
	}
	\label{fig:inter_csp}
\end{figure}

\noindent\textbf{Shot Scale.}
Shot scale has $5$ categories (as shown in Fig.~\ref{fig:data_csp_scale}): 
(1) \emph{extreme close-up shot}: it only shows a very small part of a subject, \eg,  an eye or a mouth of a person;
(2) \emph{close-up shot}: it concentrates on a relatively small part of a subject, \eg, the face of the hand of a person;
(3) \emph{medium shot}: it contains part of a subject \eg, a figure from the knees or waist up; 
(4) \emph{full shot}: it includes the full subject; 
(5) \emph{long shot}: it is taken from
a long distance and the subject is very small within the frames.

\noindent\textbf{Shot Moviement.}
Shot movement has $4$ categories (as shown in Fig.~\ref{fig:data_csp_mov}):
(1) \emph{static shot}: the camera is fixed but the subject is flexible to move; 
(2) \emph{pans and tilts shot}: the camera moves or rotates; 
(3) \emph{zoom out shot}: the camera zooms out for pull shot; 
(4) \emph{zoom in shot}: the camera zooms in for push shot.

\noindent\textbf{Annotation Categories.}
Compared with the definition in ~\cite{giannetti1999understanding}, we simplify
the cinematic styles to make the annotation affordable. But
we make sure the simplified categories are enough for most
applications.
For example, other cinematic styles like lighting are also important aspects. But the standard
of lighting is hard to develop and we are now working
on that with movie experts.

\noindent\textbf{Annotation Workflow.}
To ensure the quality of the annotation, we make efforts as follows,
(1) Instead of asking annotators to cut shot from movie or trailers and annotate their labels simultaneously,
we cut shots from movies and trailers with the off-the-shelf method~\cite{sidiropoulos2011temporal}.
It is going to mitigate annotators' burdens considering the shot detection is well solved.
(2) All the annotators went through a training phase first from cinematic professionals. And they are not allowed to annotate until they pass the professional test.
They use our web-based annotation tool, as shown in Fig.~\ref{fig:inter_csp}. Each task is annotated three times to ensure a high annotation consistency.

\noindent\textbf{Sample.}
We show five samples of each category of shot scale in one movie \emph{Mission Impossible}. 
As we can see from Fig.~\ref{fig:data_csp_scale}. 
Extreme close-up shot reveals the detailed information about characters or objects.
Close-up shot reveals information about characters' faces and their emotions.
Medium shot shows character involved activities.
Long shot shows  the scenes setting of a character or an object.
Full shot shows the landscapes that set up the movie.
Additionally we show four examples of different movement type shots, as shown in Fig.~\ref{fig:data_csp_mov}.

\subsection{Scene Boundary}
\label{subsec:sseg_anno}
\begin{figure}[!t]
	\begin{center}
		\includegraphics[width=\linewidth]{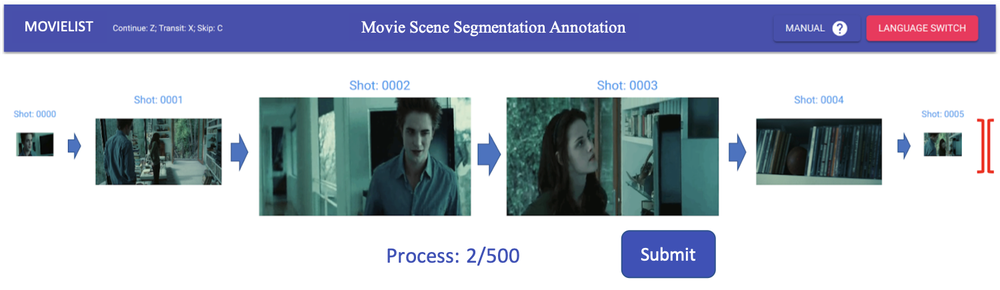}
	\end{center}
	\caption{\small
		Annotation interface of scene boundary.
	}
	\label{fig:inter_sseg}
	\vspace{-0pt}
\end{figure}

\noindent\textbf{Annotation Workflow.}
To increase the efficiency of annotating procedure, as well as to improve the label quality, we propose the
following annotating strategy.
It would be a prohibitive amount of work if the annotators are asked to go through the movies frame by frame. 
In this work, we adopt
a shot-based approach, based on the observation that a shot
is a part of exactly one scene. Hence, we consider a
scene as a continuous subsequence of shots, and consequently the scene boundaries can be selected from shot
boundaries. 
Also note that shot detection, as a task, is already well solved. Specifically, for each movie, we first divide it into shots using off-the-shelf methods~\cite{sidiropoulos2011temporal}. This
shot-based approach greatly simplifies and speeds up the
annotation process.

\noindent\textbf{Annotation Interface.}
We also developed a web-based annotation tool, as shown in Fig.~\ref{fig:inter_sseg} to facilitate human annotators to determine whether a scene transit or not between each pair of shots. On the web UI, annotators can watch two shots placed in the center, together with
the frames preceding and succeeding these shots, respectively on the left and right sides. Annotators are required to
make a decision as to whether the two shots belong to different scenes after watching both shots. The preceding
and succeeding frames also provide a useful context.

\noindent\textbf{Annotation Quality.}
A movie contains about $1K$ to $2K$ shots. It takes about
one hour for an annotator to work through a whole movie,
which can cause the difficulty to focus. To mitigate such
issues, we cut each movie into 3 to 5 chunks (with overlaps),
each containing about $500$ to $700$ shots. Then it only takes
about $15$ minutes to annotate one chunk. We found that it
is much easier for annotators to focus when they work on
chunks instead of the entire movie.
All the annotators went through
a training phase to learn how to use the annotation tool and
how to handle various ambiguous cases before they work on
the annotation tasks. 
We asked annotators to make careful
decisions between each pair of shots. If an annotator is not
sure about certain cases, they can choose “unsure” as an
answer and skip.
The collection constitutes two rounds. In the first round,
we dispatch each chunk of movies to three independent annotators so that we can check consistency. In the second
round, inconsistent annotations, \ie the results collected
from three annotators do not agree, will be re-assigned to
two additional annotators. For such cases, we will get five
results for each chunk.

\begin{figure}[t]
	\centering
	\includegraphics[width=0.95\linewidth]{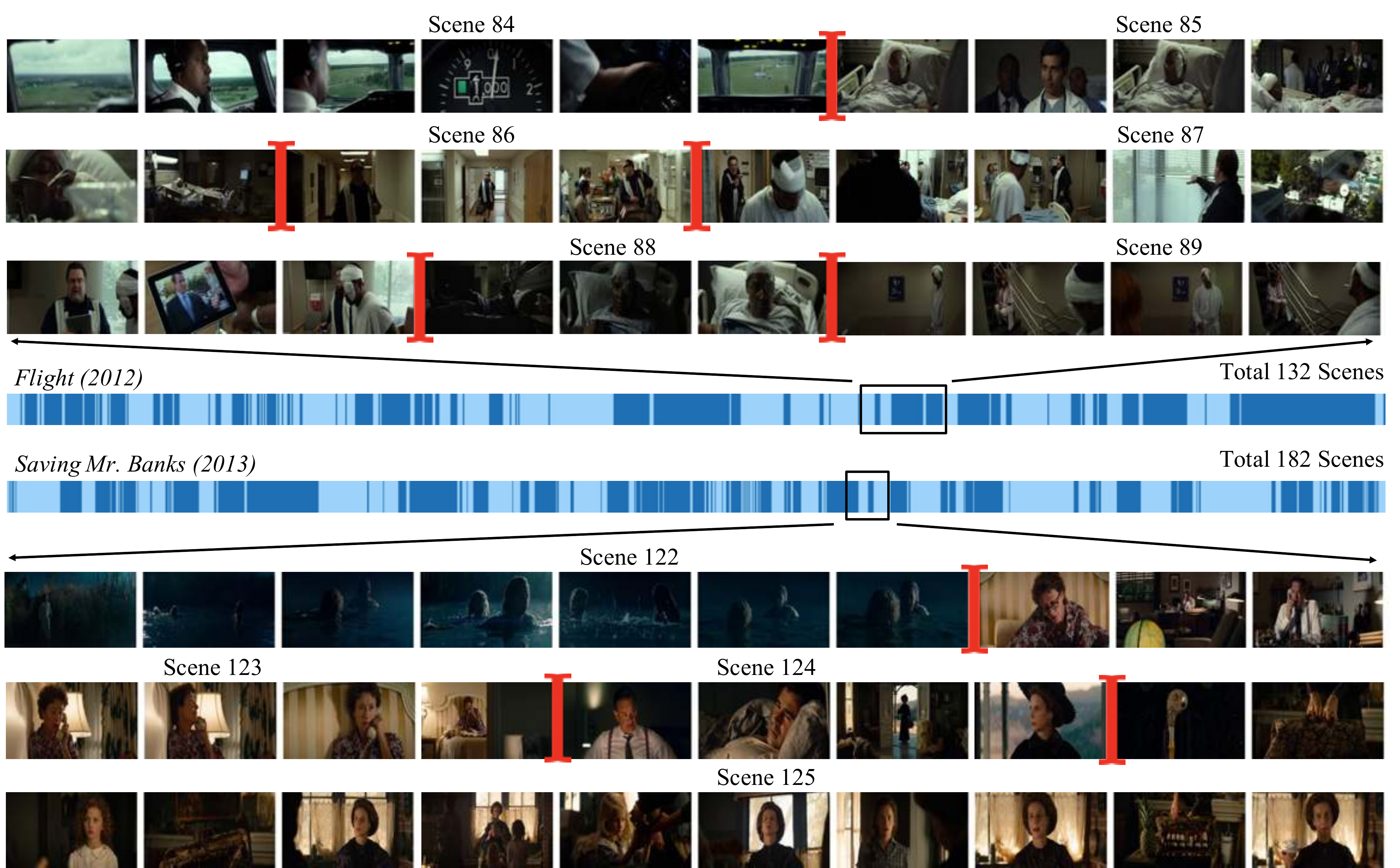}
	\caption{
		Examples of the annotated scenes from two movies. The two lines in the middle correspond to the whole movie time line
		where the dark blue and light blue regions represent different annotated scenes, while the representative frames sampled from some
		scenes are also shown. 
	}
	\label{fig:data_sseg}
\end{figure}

\noindent\textbf{Samples.}
We show two samples of scene boundaries in Fig.~\ref{fig:data_sseg}.
Segmenting scenes is challenging since visual cues are not enough to recognize the scene boundaries. 
For example, in the first movie \emph{Flight} scene 84 to scene 89 and the second movie \emph{Saving Mr. Banks} scene 123 to scene 124, most of the frames look very similar to each other. Additional semantic information such as character, action, audio are 
needed to make the right prediction.
The difficulty of segmenting vary among different scenes. Some easy cases are also listed, \eg the scene boundary between scene 122 and scene 123 is easy to recognize since visual changes are obvious.

\begin{figure}[!ht]
	\centering
	\begin{minipage}[t]{\linewidth}
		\centering
		\subfloat[Eat.]{
			\begin{tabular}[t]{c}
				\includegraphics[width=0.25\linewidth]{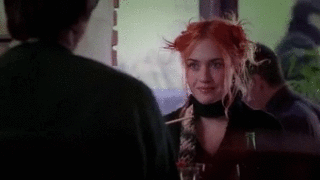}
				\includegraphics[width=0.25\linewidth]{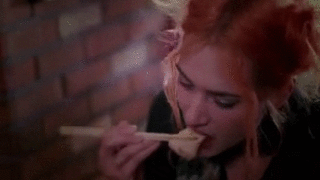}
				\includegraphics[width=0.25\linewidth]{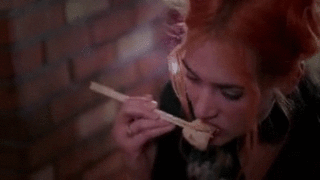}
				\includegraphics[width=0.25\linewidth]{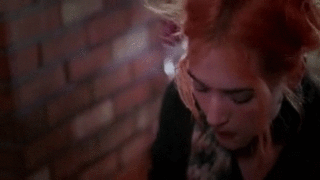}\\
				\includegraphics[width=0.25\linewidth]{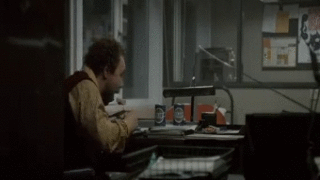}
				\includegraphics[width=0.25\linewidth]{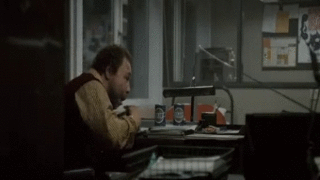}
				\includegraphics[width=0.25\linewidth]{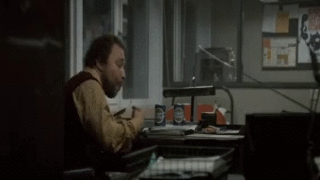}
				\includegraphics[width=0.25\linewidth]{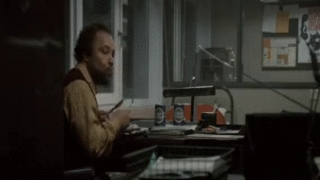}	
			\end{tabular}
		}
		
	\end{minipage}
	\quad
	\begin{minipage}[t]{\linewidth}
	\centering
	\subfloat[Kiss.]{
		\begin{tabular}[t]{c}
			\includegraphics[width=0.25\linewidth]{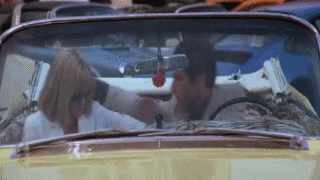}
			\includegraphics[width=0.25\linewidth]{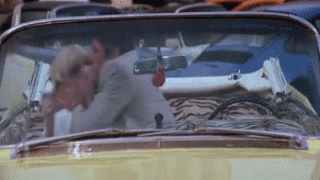}
			\includegraphics[width=0.25\linewidth]{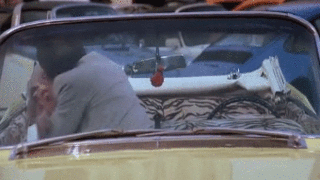}
			\includegraphics[width=0.25\linewidth]{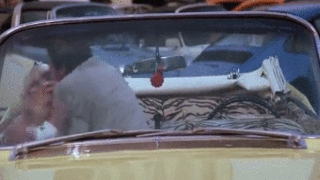}\\
			\includegraphics[width=0.25\linewidth]{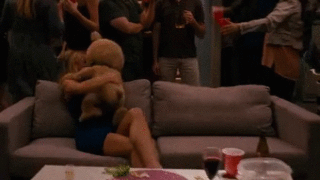}
			\includegraphics[width=0.25\linewidth]{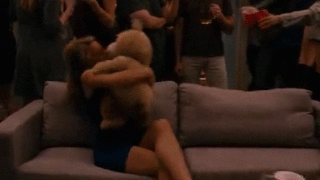}
			\includegraphics[width=0.25\linewidth]{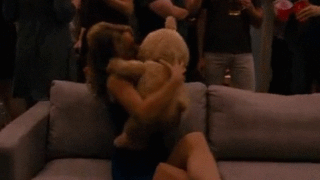}
			\includegraphics[width=0.25\linewidth]{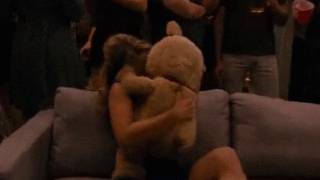}		
		\end{tabular}
	}
	
\end{minipage}
	\quad
\begin{minipage}[t]{\linewidth}
	\centering
	\subfloat[Shoot Gun.]{
		\begin{tabular}[t]{c}
			\includegraphics[width=0.25\linewidth]{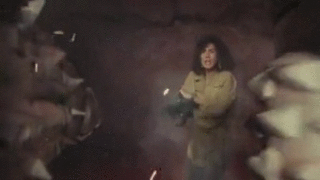}
			\includegraphics[width=0.25\linewidth]{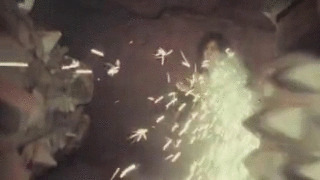}
			\includegraphics[width=0.25\linewidth]{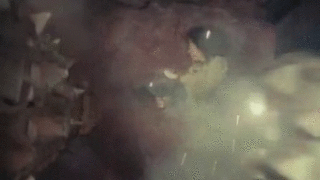}
			\includegraphics[width=0.25\linewidth]{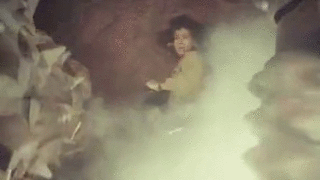}\\	
			\includegraphics[width=0.25\linewidth]{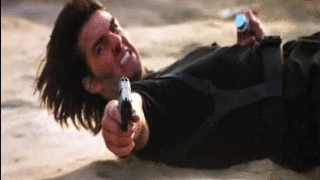}
			\includegraphics[width=0.25\linewidth]{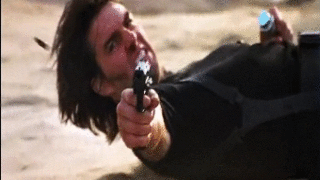}
			\includegraphics[width=0.25\linewidth]{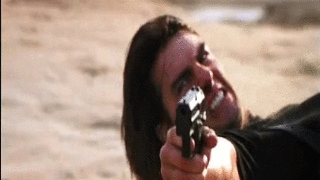}
			\includegraphics[width=0.25\linewidth]{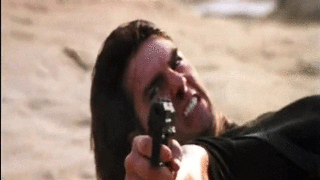}		
		\end{tabular}
	}
	
\end{minipage}

	\quad
\begin{minipage}[t]{\linewidth}
	\centering
	\subfloat[Play Guitar.]{
		\begin{tabular}[t]{c}
			\includegraphics[width=0.25\linewidth]{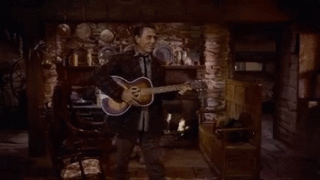}
			\includegraphics[width=0.25\linewidth]{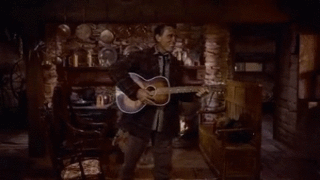}
			\includegraphics[width=0.25\linewidth]{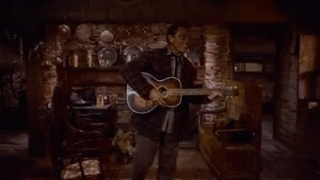}
			\includegraphics[width=0.25\linewidth]{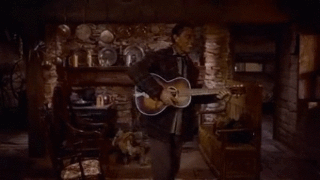}\\
			\includegraphics[width=0.25\linewidth]{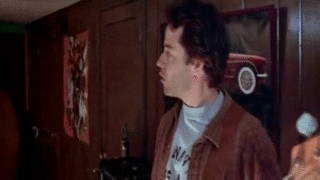}
			\includegraphics[width=0.25\linewidth]{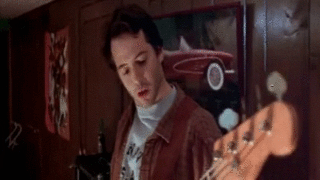}
			\includegraphics[width=0.25\linewidth]{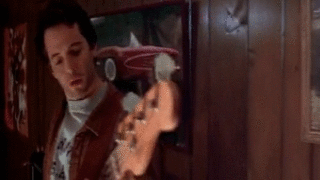}
			\includegraphics[width=0.25\linewidth]{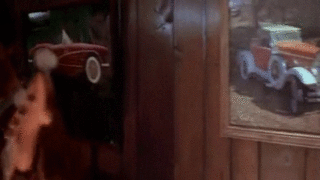}		
		\end{tabular}
	}
	
\end{minipage}
	\caption{Example actions sample frames from category ``eat'', ``kiss'', ``shoot gun'' and ``play guitar''.}
	\label{fig:example_action}
\end{figure}

\begin{figure}[!ht]
	\centering
	\begin{minipage}[t]{\linewidth}
		\centering
		\subfloat[Airport.]{
			\begin{tabular}[t]{c}
				\includegraphics[width=0.25\linewidth]{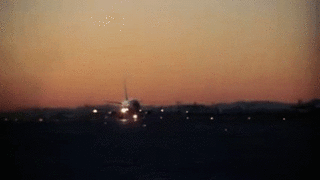}
				\includegraphics[width=0.25\linewidth]{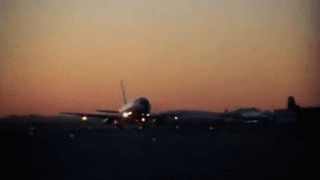}
				\includegraphics[width=0.25\linewidth]{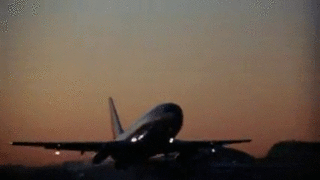}
				\includegraphics[width=0.25\linewidth]{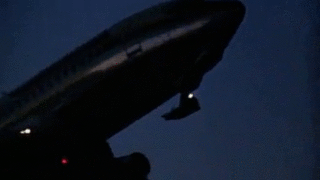}\\
				\includegraphics[width=0.25\linewidth]{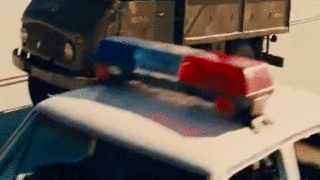}
				\includegraphics[width=0.25\linewidth]{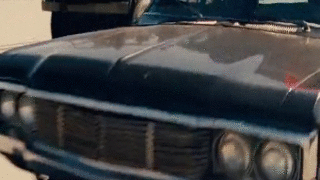}
				\includegraphics[width=0.25\linewidth]{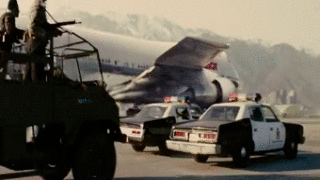}
				\includegraphics[width=0.25\linewidth]{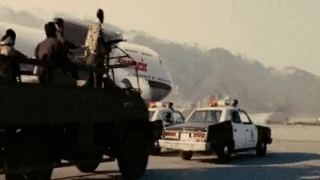}	
			\end{tabular}
		}
		
	\end{minipage}
	\quad
	\begin{minipage}[t]{\linewidth}
		\centering
		\subfloat[Pier.]{
			\begin{tabular}[t]{c}
				\includegraphics[width=0.25\linewidth]{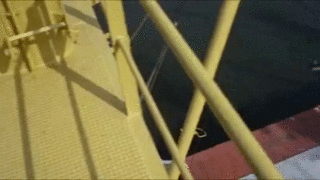}
				\includegraphics[width=0.25\linewidth]{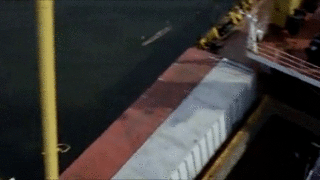}
				\includegraphics[width=0.25\linewidth]{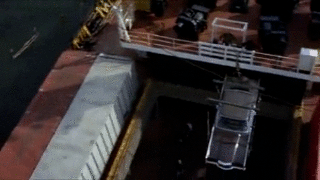}
				\includegraphics[width=0.25\linewidth]{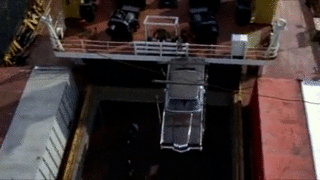}\\
				\includegraphics[width=0.25\linewidth]{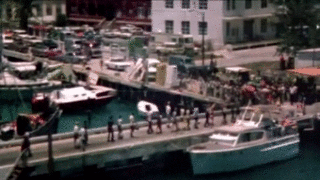}
				\includegraphics[width=0.25\linewidth]{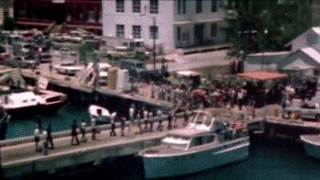}
				\includegraphics[width=0.25\linewidth]{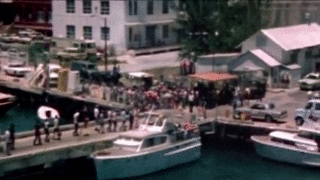}
				\includegraphics[width=0.25\linewidth]{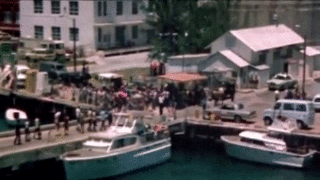}		
			\end{tabular}
		}
		
	\end{minipage}
	\quad
	\begin{minipage}[t]{\linewidth}
		\centering
		\subfloat[Inside Car.]{
			\begin{tabular}[t]{c}
				\includegraphics[width=0.25\linewidth]{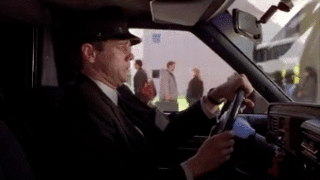}
				\includegraphics[width=0.25\linewidth]{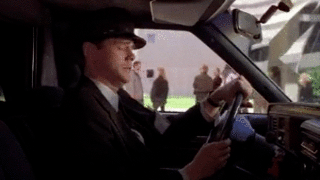}
				\includegraphics[width=0.25\linewidth]{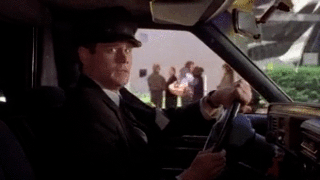}
				\includegraphics[width=0.25\linewidth]{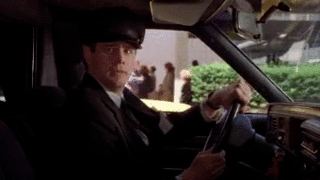}\\	
				\includegraphics[width=0.25\linewidth]{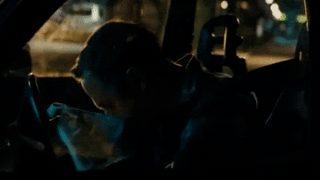}
				\includegraphics[width=0.25\linewidth]{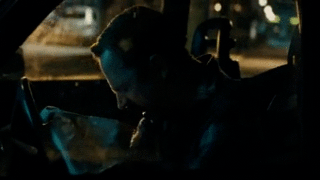}
				\includegraphics[width=0.25\linewidth]{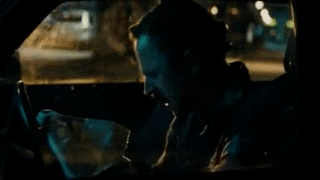}
				\includegraphics[width=0.25\linewidth]{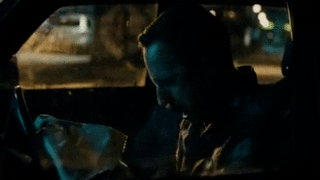}		
			\end{tabular}
		}
		
	\end{minipage}
	\quad
	\begin{minipage}[t]{\linewidth}
		\centering
		\subfloat[Spaceship.]{
			\begin{tabular}[t]{c}
				\includegraphics[width=0.25\linewidth]{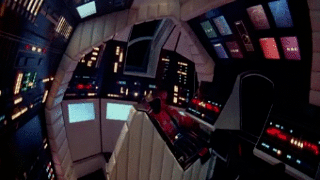}
				\includegraphics[width=0.25\linewidth]{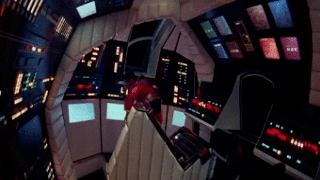}
				\includegraphics[width=0.25\linewidth]{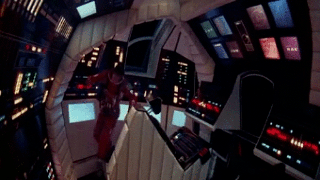}
				\includegraphics[width=0.25\linewidth]{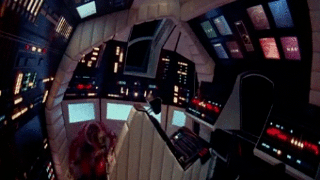}\\
				\includegraphics[width=0.25\linewidth]{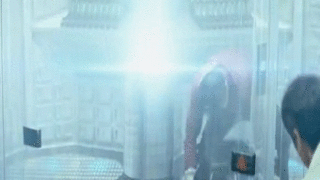}
				\includegraphics[width=0.25\linewidth]{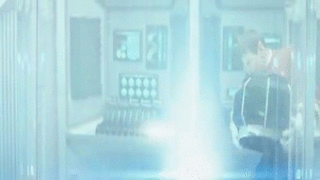}
				\includegraphics[width=0.25\linewidth]{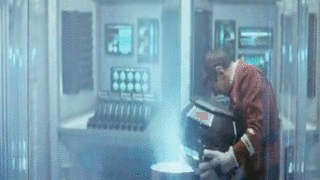}
				\includegraphics[width=0.25\linewidth]{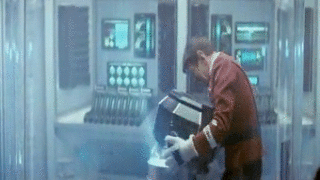}		
			\end{tabular}
		}
	\end{minipage}
	\caption{Example place sample frames from category ``airport'', ``pier'', ``inside car'' and ``spaceship''.}
	\label{fig:example_place}
\end{figure}

\subsection{Action and Place Tags}
\label{subsec:stag_anno}

%
In this section, we introduce the annotating procedure of
action and place tags in MovieNet. We develop an interface to jointly label
action and place tags. The detailed workflow and introduction
of interface will be expanded as follows.

\begin{figure}[!b]
	\centering
	\includegraphics[width=\linewidth]{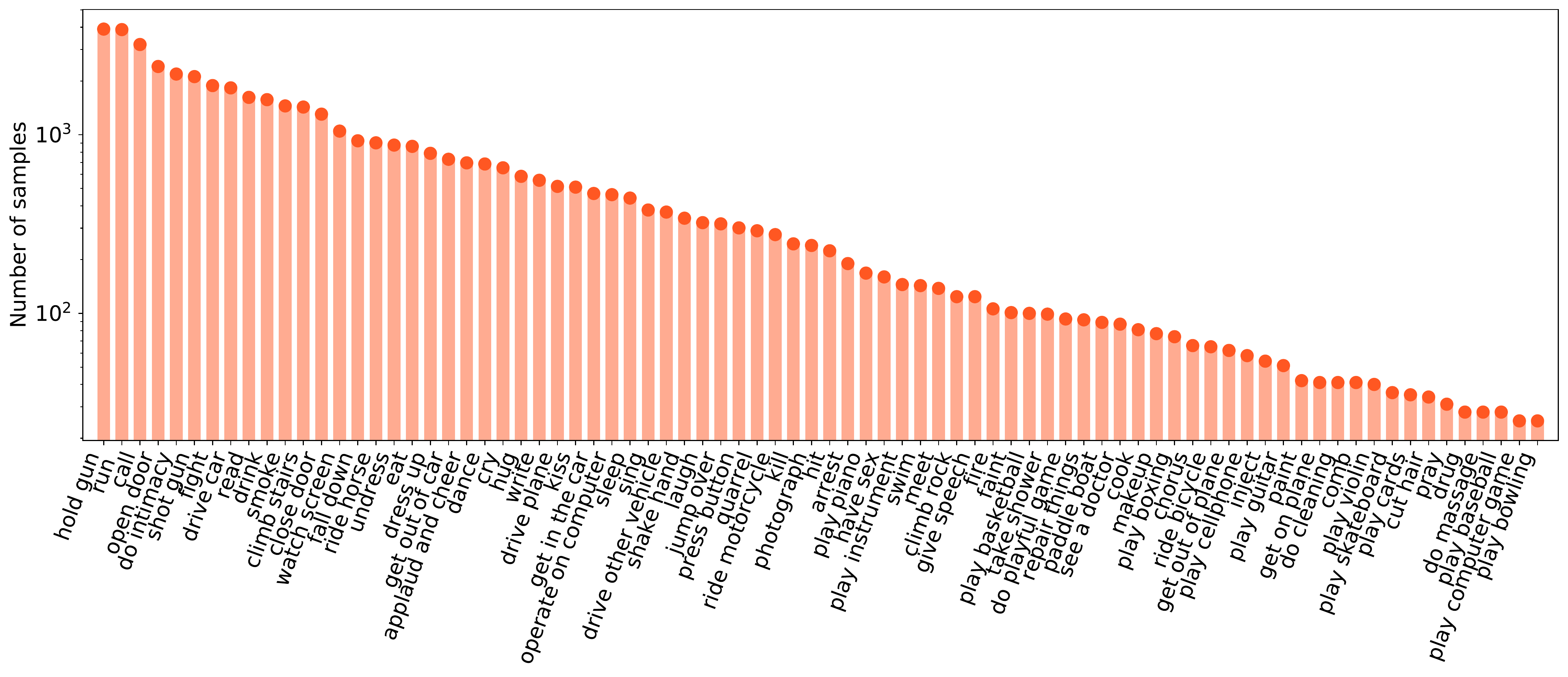}
	\caption{Distribution of action annotations in MovieNet (y-axis in log scale).
	}
	\label{fig:action_dist}
\end{figure}

\begin{figure}[!bh]
	\centering
	\includegraphics[width=\linewidth]{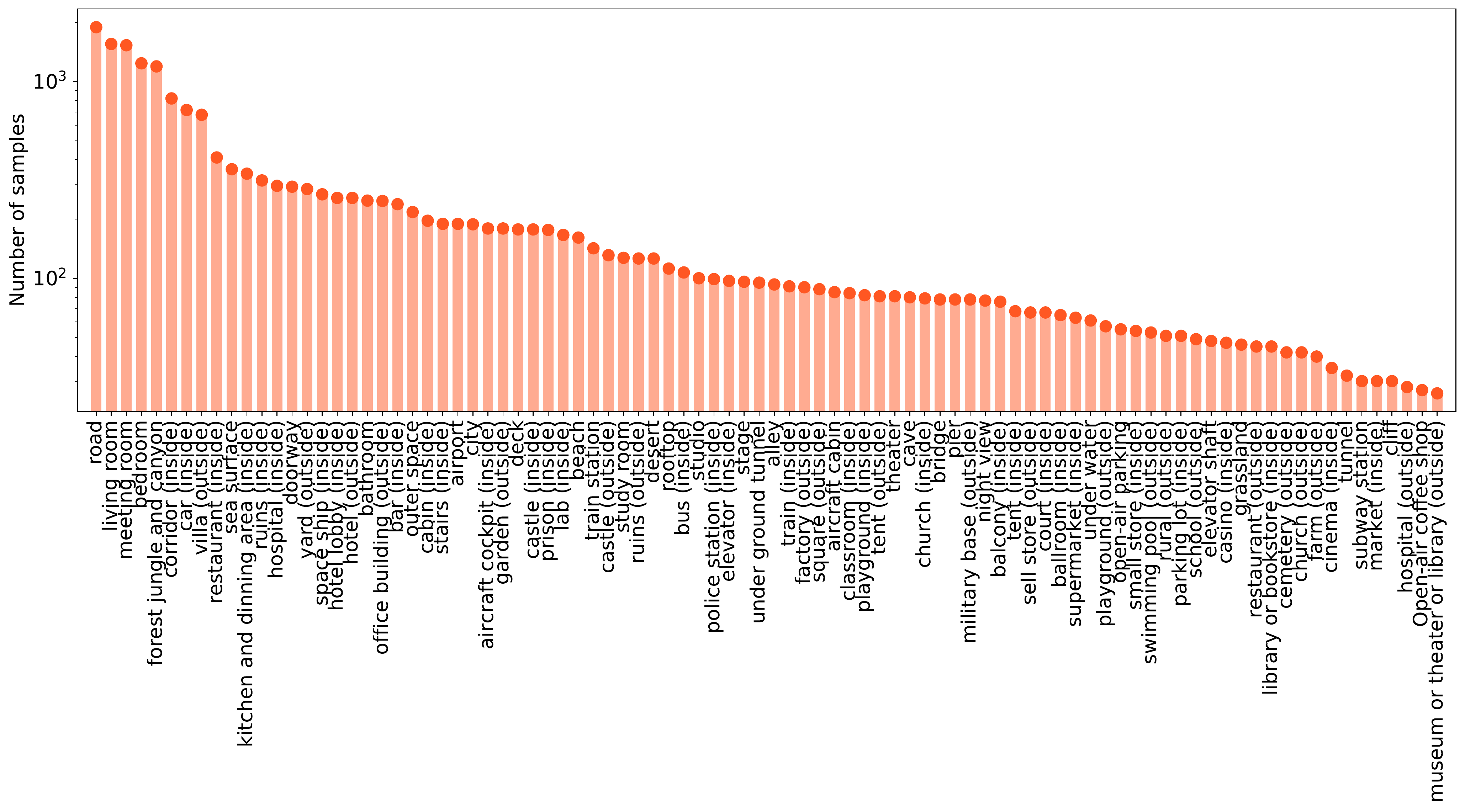}
	\caption{Distribution of place annotations in MovieNet (y-axis in log scale).
	}
	\label{fig:place_dist}
\end{figure}

\noindent\textbf{Annotation Workflow.}
We split each movie into segments according to scene boundaries.
Each scene video lasts around 2 min.
We manually annotated tags
of place and action for over each segment. 
For place annotation, each segment is annotated with multiple place tags, \eg, {\emph{deck}, \emph{cabin}}, that cover all the places appear
in this video. 
While for action annotation, we ask the annotators
to first
detect sub-clips that contain people and actions.
Here they are asked to annotate the boundary that cover an 
uninterrupted human actions.
 Then they are asked to assign multiple action tags to each sub-clip
 to describe the actions within it. 
 We have made
the following efforts to keep tags diverse and informative:
(1) We encourage the annotators to create new tags and (2)
Tags that convey little information for story understanding,
\eg, stand, sit, walk, watch, listen and talk, are excluded.  Note that in AVA~\cite{gu2018ava},
there are a large amount of this kind of actions, but here
we choose to ignore these tags. 
This makes our dataset focus on the actions that more related
to story telling.
Finally, we merge the
tags and filtered out 80 action classes and 90 place classes
with a minimum frequency of 25 as the final annotations. In
total, there are 13.7K segments with 19.6K scene tags and
41.3K action clips with 45K action tags. 
The detailed statistics are shown in Tab.~\ref{tab:mnet_stag_stat}.
Fig.~\ref{fig:example_action} and Fig.~\ref{fig:example_place}
show some of the samples from MovieNet action and place tags respectively.
The distribution of action and place tags are shown in 
Fig.~\ref{fig:action_dist} and Fig.~\ref{fig:place_dist} respectively.

\begin{table}[t]
	\centering
	\caption{Detailed statistics of MovieNet action/place tags.}
	\label{tab:mnet_stag_stat}
	\begin{tabular}{c|ccc|c}
		\hline
		& Train & Val  & Test  & Total \\ \hline
		\# Action Clip & 23747 & 7543 & 9969  & 41259 \\
		\# Action Tag  & 26040 & 8071 & 10922 & 45033 \\
		\# Place Clip  & 8101  & 2624 & 2975  & 13700 \\
		\# Place Tag   & 11410 & 3845 & 4387  & 19642 \\ \hline
	\end{tabular}
\end{table}

\begin{figure}[t]
	\centering
	\includegraphics[width=0.95\linewidth]{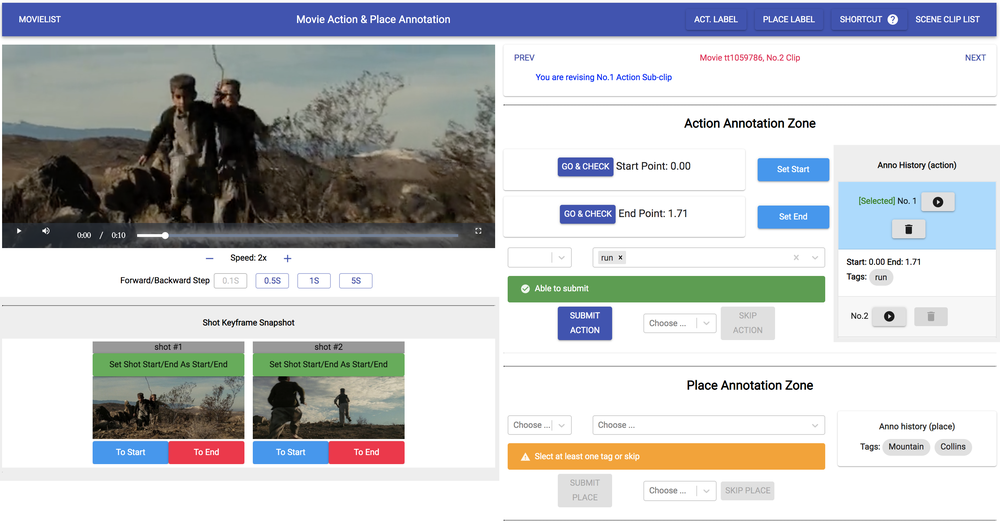}
	\caption{
		Annotation interface of the action and place tagging in MovieNet.
	}
	\label{fig:inter_mtag}
\end{figure}

\noindent\textbf{Annotation Quality.}
As mentioned above, the annotators can not only choose
pre-defined tags but also create tags as they will. 
Before the annotation procedure starts, we create a pre-defined list of action 
and place tags by the following steps: 
(1) We collect the tags from previous works. The action tags
are collected from datasets like AVA~\cite{gu2018ava}, Hollywood2~\cite{marszalek2009actions}, \etc and
the place tags are from datasets like Places~\cite{zhou2017places}, HVU~\cite{diba2019holistic}, Placepedia~\cite{huang2020placepedia}, \etc.
(2) We leverage GoogleNLP tools to detect verbs that stand for actions and nouns
that stand for location. Then we manually choose some
useful action and place tags into the list.
(3) We randomly annotate a few hours of videos to collect
tags before we ask the annotators to do so.

Besides, in case that there are uncovered tags, we
highly recommend the annotators to create tags by themselves.
During annotation stage, we gradually collect and merge
the created tags into our pre-defined list for improving
the annotation efficiency.

\noindent\textbf{Annotation Interface.}
We provide annotators with easy-to-use annotation
interface shown in Fig.~\ref{fig:inter_mtag}.
At the beginning of annotating tags, annotators are able
to get familiar with all the pre-defined tags by clicking
the button ``ACT. LABEL'' and ``PLACE LABEL'' in the menu bar.
Then they can carry out action and place
annotation at ``Action Annotation Zone'' and
``Place Annotation Zone'' respectively.
For the convenience of annotators and also for the consideration
of annotation quality, we provide a function of replaying
history action annotation that enable the annotators
to replay what they just labeled and refine the original annotations.
We also provide snapshot of shot keyframes to help the 
annotators quickly grasp the rough content of the current video.
To help annotate temporal action boundaries, we provide
workspace for annotators to set timestamps by moving forward or backward at a minimum stride of 0.1 seconds.
By the observation that shot boundaries are often action boundaries, we provide shortcuts to set action boundary
as shot boundary. The above strategies are beneficial for improving
annotation quality and efficiency.

\begin{table}[t]
	\centering
	\caption{Comparison of action and place tags with related datasets. For AVA, the action tags are annotated every second, hence the number of tags are larger than expected (see *). So for AVA, we show the statistics after merging the person instance into tracklet for fair comparison, resulting in 116K character tracklet (each tracklet is taken as a clip) and 360K action tags. }
	\label{tab:data-scene-tag}
	\begin{tabular}{c||c|c|c|c|c|c}
		\hline
		Dataset    & dura.(h) & action clip & action tag  & place clip & place tag & source \\ \hline
		Hollywood2~\cite{marszalek2009actions} & 21.7           &1.7K & 1.7K    &1.2K      & 1.2K   & movie      \\
		MovieGraphs~\cite{vicol2018moviegraphs} & 93.9    &7.6K    & 23.4K      &7.6K   & 7.6K   & movie      \\
		AVA~\cite{gu2018ava}                  & 107.5   &116K    & 360K(1.58M*)     &-    & -    & movie        \\ 
		SOA~\cite{Ray_2018_ECCV} & - & 308K & 484K & 173K & 223K & web video \\ 
		HVU~\cite{diba2019holistic} & - & 481K & 1.6M & 367K & 1.5M & web video \\ 
		MovieNet          & 214.2     &41.3K  & 45.0K  &13.7K   & 19.6K   & movie     \\ \hline
	\end{tabular}
\end{table}

\noindent\textbf{Dataset Comparison.}
Here we compare our annotated tags with other datasets with action and place tagging.
The comparison is shown in Tab.~\ref{tab:data-scene-tag}.
Note that for fair comparison, we merge the tags of AVA because they are annotated every second.
The de-duplication is done by merging the tags within the same character tracklet.
After de-duplication, the
number of tags is $360K$ in AVA. But most of them are
common actions like “stand”. There are $116K$ tracklet
with $426K$ bboxes in AVA. In a word, AVA is comparable
to MovieNet in spatial temporal action recognition,
but MovieNet can support much more research topics.

\subsection{Synopsis Alignment}
\label{subsec:syn_anno}
%
To support the movie segment retrieval task, we manually associate 
movie segments and synopsis paragraphs. 
In this section, we will present the following details about
the movie-synopsis alignment, including annotation interface and workflow.

\begin{figure}[t]
	\centering
	\includegraphics[width=0.9\linewidth]{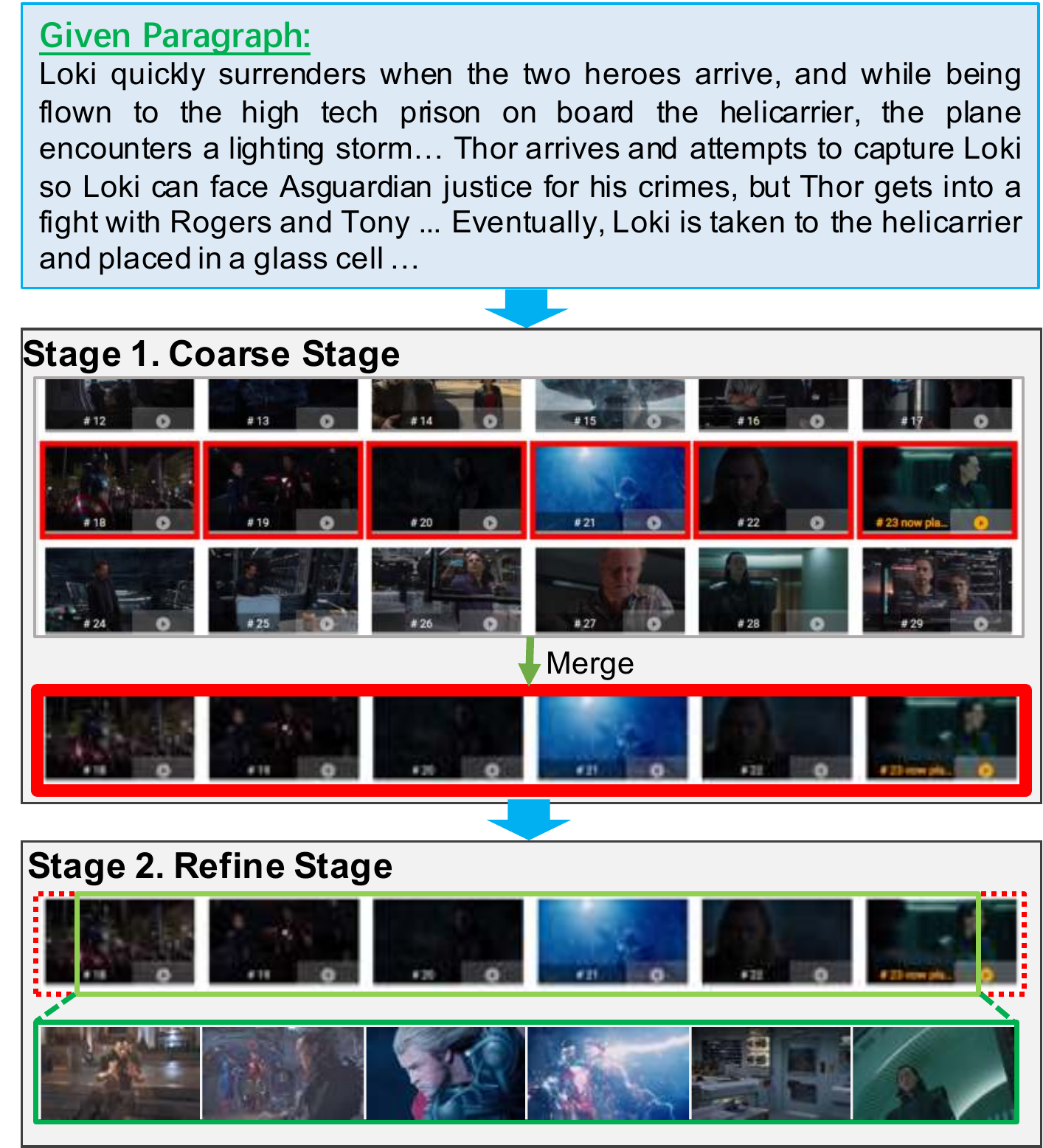}
	\caption{\small{Example of the annotating procedure for the movie
			\emph{The Avengers}. At the coarse stage, annotator chooses consecutive
			clips. At the refine stage, boundaries are refined.}
	}
	\label{fig:dataset_supp}
	\vspace{-10pt}
\end{figure}

%
\noindent\textbf{Annotation Workflow.}
After collecting synopses from IMDb, 
we further filtered out synopses with high quality, 
\ie those contain more than 50 sentences, for annotating.
Then we develop a coarse-to-fine procedure to effectively align
each paragraph to its corresponding segment.
(1) At the coarse stage, each movie is split into $N$ segments,
each lasting few minutes. Here we set $N=64$.
For each synopsis paragraph, we ask the annotators to
select $K$ consecutive clips that cover the content of the
whole synopsis paragraph.
(2) At the refine stage, the annotators are asked to refine the
boundaries that make the resulting segment better aligned
with the synopsis paragraph.

To ensure the quality of the annotation, we make the following
efforts.
(1) The synopsis paragraphs from the same movie will be assigned
to the same annotator. To ensure they are familiar with the movie, we provide the annotators with detailed overview of 
the movie, including character list, plot, reviews, \etc.
(2) Each synopsis paragraph is dispatched to $3$ annotators.
Then, we only keep those annotations with high consistency,
\ie, those with high temporal IoU among all the $3$ annotations.
Finally, $4208$ paragraph-segment pairs are obtained.

\begin{figure}[t]
	\centering
	\includegraphics[width=0.98\linewidth]{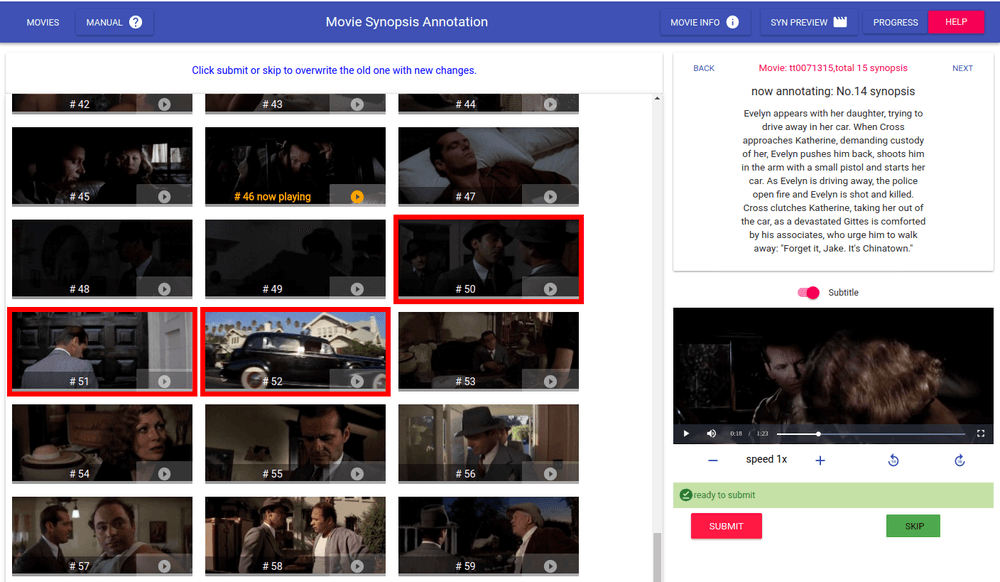}
	\caption{\small{User interface for the coarse stage. The left part display
			64 movie clips; the upper right panel shows the synopsis paragraph;
			the bottom right player plays the selected movie clip. 
		}
	}
	\label{fig:coarse}
\end{figure}

\begin{figure}[t]
	\centering
	\includegraphics[width=0.98\linewidth]{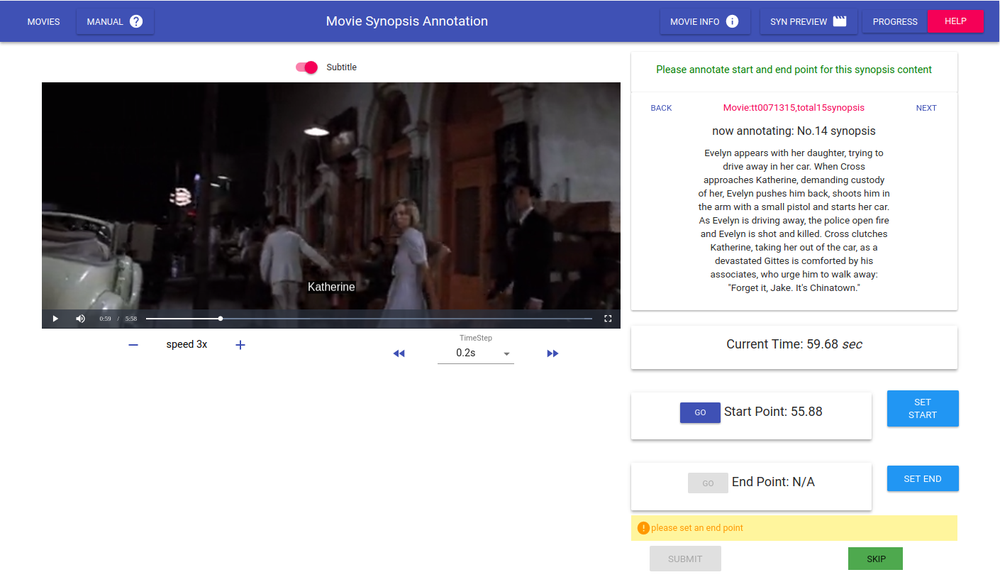}
	\caption{\small{User interface for the refine stage. The left part displays
			the merged movie segment; the upper right panel shows the synopsis paragraph;
			the bottom right buttons are for adjusting the boundaries.}
	}
	\label{fig:refine}
\end{figure}

\noindent\textbf{Annotation Interface.}
The movie-synopsis alignment is collected 
in a coarse-to-fine manner shown in Fig.~\ref{fig:dataset_supp}.
We develop an online interface to carry out these two stages.
The interface of the coarse stage is shown in Fig.~\ref{fig:coarse}.
At the beginning of annotating each movie, annotators are
required to browse the overview of this movie,
which is available at ``MOVIE INFO'' and ``SYN PREVIEW''
in the menu bar. Then annotators select a subsequence of N consecutive clips
that cover the corresponding synopsis description shown in the text panel.
After receiving the annotations, the back-end server will merge
the consecutive clips into a whole segment for refine stage.
As shown in Fig.~\ref{fig:refine}, at the refine stage, annotators adjust the temporal boundaries of the resultant segments. 
We allow annotators to set timestamps at the current playback location
as start or end timestamps.
To enable fine adjustment, users are able to control the video player 
by moving forward or backward at a minimum stride of 0.1 seconds.

\subsection{Trailer Alignment.}
\label{subsec:trailer_align}
To facilitate tasks like trailer generation, we provide an automatic process for matching 
a shot in a trailer to its movie counterpart. 
 The process is introduced below.
 
 Consider each shot in a trailer as a sequence of images $s^T=\{s_1^T, \dots, s_M^T\}$
where $M$ is the number of frames in this shot.
For a movie, we define it as a sequence of frame $s^M=\{s_1^M, \dots, s_N^T\}$ where
$N$ is the number of frame in the movie.
To locate the position of the trailer shot in the movie is to find the most 
similar sub-sequence of the movie with the shot sequence.
Let $Sim(s^T_i, s^M_j)$ denotes the similarity of i-th frame in trailer shot and j-th frame in the movie.
The solution is to find the sub-sequence in the movie that maximize the similarities:
\begin{equation}\label{eq:trailer_align}
\begin{aligned}
& j^* = \underset{j}{\text{argmax}}
& & \sum_{i=1}^{N} Sim(s^T_i, s^M_{j+i-1} )\\
& \text{s.t.}
& & j \leq N - M + 1.
\end{aligned}
\end{equation}

To obtain the similarity between two frames, we resort to features from low-level to high-level.
For each frame, we extract GIST feature~\cite{douze2009evaluation} and feature from $pool5$ layer of ResNet-50~\cite{he2016deep}
that pre-trained on ImageNet~\cite{russakovsky2015imagenet}.
We choose to use low-lwvel feature like GIST feature because
we observe that most of the frames from trailers are alike with the original ones in movies,
only with slightly changes in terms of color, size, lighting, boundary, \etc.
For some harder cases like cropped or carefully edited shots, we find high-level features like 
ResNet feature works.
Hence, the similarity can be obtained by
$$Sim(s_a, s_b) = cosine(f_a^{gist}, f_b^{gist}) + cosine(f_a^{imagenet}, f_b^{imagenet}).$$
where $f^{gist}$ stands for GIST feature while $f^{imagenet}$ stands for feature from ResNet.
 
Solving the optimization problem would result in the required alignment results. 
For those aligned shots with low optimized similarity score, we set it as misaligned shot.
By observation, those misaligned shots are mostly transition shots or shots that only exist in the trailer.
We then manually checked the alignment result to make sure the alignment results are accurate.
\setcounter{table}{0} 
\setcounter{figure}{0} 

\section{Experiments}
\label{sec:supp_experiment}
In this section, we introduce the detailed information for each benchmark.
Note that unless specified, the $1100$ movies are split according to
a ratio of $3:1:1$. The annotations are split according to the result of
the split of movies, hence there are no overlapping movies among
train, val and test sets for each task.

\subsection{Genre Classification}
\label{subsec:exp_genre}

\begin{table}[t]
	\centering
		\caption{Per-genre performance of genre classification.}
	\label{tab:genre-cls}
	\begin{tabular}{r|ccc}
		\hline
		& R@0.5 & P@0.5  & AP    \\ \hline
		Drama       & 79.42 & 71.16  & 79.95 \\ \hline
		Comedy      & 48.65 & 68.61  & 68.81 \\ \hline
		Thriller    & 14.50 & 64.98  & 49.80 \\ \hline
		Action      & 22.21 & 73.96  & 54.60 \\ \hline
		Romance     & 14.02 & 71.93  & 49.27 \\ \hline
		Horror      & 8.76  & 70.03  & 35.51 \\ \hline
		Crime       & 39.30 & 74.12  & 49.25 \\ \hline
		Documentary & 4.79  & 85.49  & 21.03 \\ \hline
		Adventure   & 24.72 & 75.24  & 53.06 \\ \hline
		Sci-Fi      & 14.51 & 81.35  & 44.14 \\ \hline
		Family      & 27.11 & 82.55  & 52.19 \\ \hline
		Fantasy     & 13.51 & 69.83  & 39.12 \\ \hline
		Mystery     & 7.76  & 76.42  & 39.70 \\ \hline
		Biography   & 0.04  & 100.00 & 9.13  \\ \hline
		Animation   & 74.09 & 93.16  & 86.45 \\ \hline
		History     & 12.52 & 82.90  & 34.41 \\ \hline
		Music       & 27.24 & 89.04  & 47.13 \\ \hline
		War         & 12.80 & 86.27  & 34.41 \\ \hline
		Sport       & 21.99 & 94.97  & 39.59 \\ \hline
		Musical     & 4.45  & 73.58  & 22.88 \\ \hline
		Western     & 51.93 & 88.89  & 73.99 \\ \hline
	\end{tabular}
\end{table}

Here we would introduce the benchmark setting, evaluation metrics, implementation details of the baseline models and show more experiment results.

\noindent \textbf{Benchmark Setting.}
There are total $28$ unique genres in MovieNet. Some rare
genres, e.g. Adult, are ignored. And we further remove non-visual
genres, e.g. News, to obtain a list with $21$ genres. For
image-based genre classification, we use the $3.9M$ photos
of MovieNet as our data source. Since not all types of photos
are related to the genres, e.g. publicity, here we only take
$4$ of them to build the benchmark for genre classification,
namely poster, still frame, product and production art, resulting
in $1.6M$ photos left. The $1.6M$ images are split into
train, validation and test set that contains $1.1M$, $160K$ and
$321K$ images respectively.
For video-based
genres classification, we take trailers for experiments. We sample
$32K$ trailers containing at least one of the $21$ genres.
They are split
as train, validation and test set contains $22.5K$, $3.2$K and
$6.4K$ videos respectively.

\noindent\textbf{Evaluation Metric.}
Genre classification is a multi-label classification problem. We use mAP, recall$@0.5$ and precision$@0.5$ as evaluation metrics. Here $0.5$ is the threshold to get the final prediction, which means that the final score (between $0$ and $1$) above $0.5$ would be set as positive while the others would be set as negative.

\noindent\textbf{Implementation Details.}
The models are trained with BCE Loss. The input size is set to $224\times224$ and we use SGD as optimizer.
For the video-based model, we get $8$ clips, each with $3$ frames, on training. At the inference stage, we would predict the score of all the clips and average them to get the final prediction.

\noindent\textbf{More Results.}
Here we show the per-genre results of the ResNet-50 model in Tab.~\ref{tab:genre-cls}. We can see that \emph{animation} achieve the highest accuracy. This is reasonable since the characteristic of animation is significant. And the AP of \emph{Biography} and \emph{Documentary}  is much lower since these genres are determined by higher semantic elements.

\subsection{Cinematic Style Analysis}
\label{subsec:exp_cs}

\begin{table}[t]
	\centering
		\caption{Baselines for MovieNet cinematic style prediction.}
	\label{tab:csp}
	\begin{tabular}{c|c|c|c}
		\hline
		Method & Backbone         & Scale-Accuracy & Move-Accuracy \\ \hline
		\multirow{3}{*}{TSN~\cite{TSN2016ECCV}} & ResNet18 & 79.31  & 68.02\\
		& ResNet34 & 82.73  & 69.91\\
		& ResNet50 & 84.08  & 70.46\\ \hline
		\multirow{3}{*}{I3D~\cite{carreira2017quo}} & ResNet18  & 76.79  & 78.45\\
		& ResNet34  & 82.10  & \textbf{82.17}\\
		& ResNet50  & 82.70  & 81.97\\ \hline
		TSN+R$^3$Net~\cite{deng2018r3net} & ResNet50 & \textbf{87.58} & 80.65 \\ \hline
	\end{tabular}
\end{table}

\noindent\textbf{Dataset Split.}
The MovieNet cinematic style prediction benchmark
contains $46K$ shots coming from $8K$ trailers where
$26K$ for training, $7K$ for validation and $13K$ for testing.

\noindent\textbf{Details of baseline models for cinematic style prediction.} We implement TSN~\cite{TSN2016ECCV} (3 segments) and I3D~\cite{carreira2017quo} with different backbones ResNet-18, ResNet-34, ResNet-50~\cite{he2016deep}.
The results are shown in the Tab.~\ref{tab:csp}.
We observe that 2D models achieve better results as the network becoming deeper both in terms of scale and movement classification.
Deeper 3D models are a bit of over fitting on movement since the performance drops a little when the network becomes deeper.

\noindent\textbf{Details of TSN+R$^3$Net.} As we point out in the paper, the subject is very important for cinematic style analysis. Thus a subject-based method is proposed to solve the problem. First, we adopt R$^3$Net~\cite{deng2018r3net} to get the saliency map of each frame in the shot. Then, each video clip passes through a two-branch classification network, one branch is for video clips, the other branch is for video saliency clips. For movement classification, the whole image and the background image are used as the inputs. For scale classification, the whole image and the subject image are used as the inputs of the two branches. The features from the two branches are concatenated and pass through a fully-connected layer to get the final prediction.

\noindent\textbf{Implementation Details.}
We take cross-entropy loss for
the classification. We train these models for 60 epochs with
mini-batch SGD, where the batch size is set to 128 and the
momentum is set to 0.9. The network weights are initialized with pretrained models from ImageNet~\cite{russakovsky2015imagenet}.
The initial learning rate is 0.001 and the learning rate will be divided
by 10 at the 20th and 40th epoch.

\subsection{Character Detection}
\label{subsec:exp_pdet}
\noindent\textbf{Dataset Split.}
We collect $1.3M$ bounding boxes from
$758K$ keyframe images. The $758K$ images are split into train and test
set with $692K$ and $66K$ respectively.

\noindent\textbf{Evaluation Metric}
Following one of the most popular benchmark for object detection -- COCO~\cite{lin2014microsoft}, we use the AP from $0.5$ to $0.95$ with a stride $0.05$, namely mAP, as our evaluation metric.

\noindent\textbf{Implementation Details.}
We take a Faster R-CNN with ResNet-50 as backbone, and train on three different datasets, COCO~\cite{lin2014microsoft}, CalTech~\cite{dollar2011pedestrian} and our MovieNet-PDet to show the large domain gap between movie and other data source. And then we also try more models on MovieNet, including a single-stage model, namely RetinaNet, and a more powerful model, \ie Cascade R-CNN~\cite{cai2018cascade} with ResNeXt-101~\cite{xie2017aggregated} backbone and feature pyramid~\cite{lin2017feature}.

\begin{table}[t]
	\begin{minipage}[t]{0.4\linewidth}\centering
		\caption{Character identification: comparison of MovieNet and related datasets that used as the training datasets
			in this benchmark.}
		\label{tab:data-person-id}
		\begin{tabular}{l|cc}
	\hline
	Dataset & ~~~ID~~~ & instance  \\ \hline
	Market~\cite{zheng2015scalable}   & 1,501       & 32K         \\
	CUHK03~\cite{li2014deepreid}   & 1,467       & 28K         \\
	CSM~\cite{huang2018person}   & 1,218       & 127K        \\ \hline
	MovieNet & \textbf{3,087}   & \textbf{1.1M}         \\ \hline
\end{tabular}
	\end{minipage}
	\hfill
	\begin{minipage}[t]{0.55\linewidth}\centering
				\caption{Performance of different methods for character identification in MovieNet.}
		\label{tab:exp-person-id_supp}
			\begin{tabular}{c|c|c|c}
	\hline
	Train Data          & cues        & Method      & ~~mAP~~ \\ \hline
	Market~\cite{zheng2015scalable}     & body   & r50-softmax & 4.62 \\ \hline
	CUHK03~\cite{li2014deepreid}       & body   & r50-softmax & 5.33 \\ \hline
	CSM~\cite{huang2018person}         & body  & r50-softmax & 26.21 \\ \hline
	\multirow{4}{*}{MovieNet} & body & r50-softmax & 32.81    \\ \cline{2-4}
	& body+face  & LP~\cite{zhu2002learning}  &  8.29  \\ \cline{2-4}
	& body+face  & two-step~\cite{loy2019wider}  &  63.95  \\ \cline{2-4}
	& body+face  & PPCC~\cite{huang2018person} & \textbf{75.95} \\ \hline
\end{tabular}
	\end{minipage}
\end{table}

\subsection{Character Identification}
\label{subsec:exp_pi}
\noindent \textbf{Dataset Split.}
We annotate
identities of more than $1.1M$ instances of $3K$ identities.
In the MovieNet cahracter identification benchmark.
They are split into train, val, test set with 
$2088$ identities with $639.9K$ instances, $821$ identities with $336.6K$ instances and $876$ identities with
$364.2K$ instances respectively.

\noindent\textbf{Benchmark Setting.}
The Character identification task is to search for all the instances
of a character in a movie with just one portrait.
To enable the task, we download a portrait from homepage
of each credited cast, which will serve as the query portraits
for the character identification tasks.

\noindent\textbf{Evaluation Metric.}
We use mAP that average the AP on each query as the evaluation protocol.

\noindent\textbf{Baseline Results.}
The results of character identification are shown in Tab.~\ref{tab:exp-person-id_supp}.
The character identification task is similar to conventional person ReID task,
however our dataset is much more challenging and larger than theirs.
So in order to show the domain gap, we train ResNet-50 with softmax loss
on three person/character identification dataset, namely, Market~\cite{zheng2015scalable}, CUHK03~\cite{li2014deepreid} and MovieNet shown in Tab.~\ref{tab:data-person-id}.
From the results, we see that due to the large domain gap, current ReID datasets cannot support the researches on character analysis in Movies.
We also adopt methods -- LP~\cite{zhu2002learning}, PPCC~\cite{huang2018person} for comparison.

\noindent\textbf{Implementation Details.}
The character identification task need to utilize both face feature and body feature. Here the face features are extracted by a ResNet-101 trained on MS1M~\cite{guo2016ms} and the body features are extracted by a ResNet-50 trained by MovieNet identity annotation or other ReID dataset. The cues in Tab.~\ref{tab:exp-person-id_supp} means that we retrieve the character by only the features mentioned by cues. The \emph{Two-Step} method means that we would first retrieve by face features, and then add some instances with high confidence to the query set, after which we would do set-to-set retrieval by body features. This is widely used in the WIDER Challenge~\cite{loy2019wider}. LP~\cite{zhu2002learning} means the naive label propagation method, which would be affected by noise and get a poor performance. And PPCC~\cite{huang2018person} improves LP by developing a competitive consensus scheme, which is the current state-of-the-art for character identification.

\begin{table}[t]
		\caption{Comparison of MovieNet scene segmentation with related datasets. }
	\centering
		\begin{tabular}{l|c|c|c}
			\hline
			Dataset & ~~\# Scene~~  & ~~Duration(hour)~~ & ~~Source~~         \\ 				\hline
			OVSD~\cite{rotman2017optimal}           & 300           & 10    & MiniFilm         \\
			BBC~\cite{baraldi2015deep}               & 670       & 9   & Documentary         \\ \hline
			MovieNet               & 42K  & 633                      & Movie        \\
			\hline
		\end{tabular}
	\label{tab:data-scene_supp}
\end{table}

\subsection{Scene Segmentation}
\label{subsec:exp_sseg}
\noindent\textbf{Dataset Split.}
The overall $42K$ scenes are split into train, val, test sets
with $25K$, $8.9K$ and $8.1K$ scene segments respectively.
There are no overlapped movies.
The comparison of our dataset with other related scene segmentation datasets
are shown in Tab.~\ref{tab:data-scene_supp}.

\noindent\textbf{Evaluation Metrics.}
We take two commonly used metrics:
(1) Average Precision (AP). Specifically in our experiment, it is the mean of AP of detected scene boundary for each movie.
(2) $M_{iou}$: a weighted sum of intersection of union of a detected scene boundary with respect to its distance to the closest
ground-truth scene boundary.

\noindent\textbf{Baseline Models.}
We reproduce Grouping~\cite{rotman2017optimal}  and Siamese~\cite{baraldi2015deep} according to their papers.
For our baseline multi-semantic LSTM (MS-LSTM), we extract audio, character, action and scene feature from each shot as follows,
\begin{itemize}
	\item \textbf{Audio feature.}
	We use tools from~\cite{xu2019recursive} that apply NaverNet~\cite{chung2019naver} pretrained on AVA-ActiveSpeaker dataset~\cite{roth2019ava} to separate speech and background sound, and stft~\cite{umesh1999fitting} to get their features respectively in a shot with 16K Hz sampling rate and 512 windowed signal length, and concatenate them to obtain {audio} features.
	\item \textbf{Character feature.}
	We firstly take the advantage of Faster-RCNN~\cite{NIPS2015_5638} pretrained on MovieNet character detection benchmark to detect character instances. And then we use a ResNet50 trained on MovieNet character identification benchmark to extract character features.
	\item \textbf{Action feature.}
	We utilize TSN~\cite{TSN2016ECCV} with AVA dataset~\cite{gu2018ava} pretraining to get \textit{action} features.
	\item \textbf{Place feature.}
	We take ResNet50~\cite{he2016deep} with Places dataset~\cite{zhou2018places} pretraining on key frame images of each shot to get {place} features.
\end{itemize}

\noindent\textbf{More Results.}
From Tab.~\ref{tab:sseg}, we observe that (1) Benefited from
large scale and high diversity, models trained on MovieNetSSeg achieve more than $40\%$ improvement in performance. Specifically, Grouping~\cite{rotman2017optimal} improves $98\%$ from 0.170 to 0.336, Siamese~\cite{baraldi2015deep} improves $34\%$ from 0.268 to 0.358, MS-LSTM improves $39\%$ from 0.334 to 0.465.
(2) Multiple semantic elements are important for scene segmentation, which highly raise the performance. Jointly using audio, character, action and place information surpass any single element.

\begin{table}[!t]
		\caption{Baseline results for scene segmentation.}
	\label{tab:sseg}
	\begin{center}
		{
			\begin{tabular}{c|c|ccc}
				\hline
				Training Data & Method
				& AP ($\uparrow$) & $M_{iou}$ ($\uparrow$)  \\
				\hline
				\multirow{2}{*}{OVSD~\cite{rotman2017optimal} }
				& Grouping~\cite{rotman2017optimal}   & 0.170                           & 0.301                           \\
				& MS-LSTM                                              & \textbf{0.313}                          & \textbf{0.387}            \\
				\hline
				\multirow{2}{*}{BBC~\cite{rotman2017optimal} }
				& Siamese~\cite{baraldi2015deep}  & {0.268} & {0.358} \\
				& MS-LSTM                                               &  \textbf{0.334}                               &     \textbf{0.379}  \\
				\hline
				\multirow{7}{*}{MovieNet-SSeg}
				& Grouping~\cite{rotman2017optimal}   & 0.336                           & 0.372                           \\
				& Siamese~\cite{baraldi2015deep}  & {0.358} & {0.396} \\
				& MS-LSTM (Audio only)                                             & 0.210                           & 0.341                           \\
				& MS-LSTM (Character only)                                              & 0.213                           & 0.348                           \\
				& MS-LSTM (Action only)                                            & 0.227                           & 0.368                           \\
				& MS-LSTM (Place only)                                             & 0.442                           & 0.421                           \\
				& MS-LSTM 				& \textbf{0.465} & \textbf{0.462} \\
				\hline
			\end{tabular}
		}
	\end{center}
\end{table}

\noindent\textbf{Implementation Details.}
We take cross entropy loss for
the binary classification. We train
these models for 30 epochs with SGD optimizer. The initial
learning rate is 0.01 and the learning rate will be divided by
10 at the 15th epoch.

\subsection{Action Recognition}
\label{subsec:exp_action}
We conduct experiments of action recognition on MovieNet action recognition benchmark. This task aims at predicting multiple action
tags of a given video.

\noindent\textbf{Dataset Split.}
The whole dataset is randomly split as train, val, test set
with $23747$, $7543$, $9969$ video clips respectively, without overlapping movies.

\noindent\textbf{Loss and Metrics.}
For training all models, we use binary cross-entropy loss
as the loss function.
For multi-label classification evaluation, we use
mAP as the evaluation protocol.

\noindent\textbf{Implementation of TSN.}
We adopt TSN~\cite{TSN2016ECCV} as one of our baseline models.
To be specific, the TSN2D model adopt ResNet50~\cite{he2016deep} as backbone.
We sample $3$ segments for each video and the consensus function
is simply \emph{Average}.
We set batch size as $32$ and dropout rate as $0.5$.
The model is trained using SGD for $100$ epochs with an initial learning rate $0.01$, momentum $0.9$ and weight decay $0.0005$.
The learning rate is divided by $10$ at the $60$ and $90$ epoch.

\noindent\textbf{Implementation of I3D.}
For I3D~\cite{carreira2017quo}, we adopt ResNet-I3D~\cite{carreira2017quo}
with depth $50$ as the baseline model.
The inflate style is set to $3\times1\times1$ and
the input length is $32$ with stride $2$.
We set batch size as $8$ and dropout rate as $0.5$.
The model is trained using SGD for $100$ epochs with an initial learning rate $0.01$, momentum $0.9$ and weight decay $0.0001$.
The learning rate is divided by $10$ at the $60$ and $90$ epoch.

\noindent\textbf{Implementation of SlowFast.}
For SlowFast Network~\cite{feichtenhofer2019slowfast}, the backbone is
I3D with ResNet50.
We set $\tau$ to $8$, $\alpha$ to $8$ and $\beta$ to $1/8$.
 The input length is $32$ with stride $2$.
 We set batch size as $8$ and dropout rate as $0.5$.
 The model is also trained $100$ epochs using  a half-period cosine schedule~\cite{loshchilov2016sgdr} of learning rate decaying with
 $n\text{-th}$ iteration learning rate as
 $0.5\eta[cos(\frac{n}{n_{max}}\pi)+1]$.
 $n_{max}$ is the max iteration number, $\eta$ is
 the basic learning rate set as $0.2$.
%
%

\begin{table}[t]
	\begin{minipage}[t]{0.52\linewidth}\centering
		\caption{Result of action classification.}
		\label{tab:tag_exp_action}
			\begin{tabular}{c|c}
			\hline
			 Method     & mAP   \\ \hline
			 TSN~\cite{TSN2016ECCV}     & 14.17 \\
			I3D~\cite{carreira2017quo}  & 20.69 \\
			SlowFast~\cite{feichtenhofer2019slowfast} & \textbf{23.52}\\ \hline
		\end{tabular}
	\end{minipage}
	\hfill
	\begin{minipage}[t]{0.46\linewidth}\centering
		\caption{Performance of different methods for place classification.}
		\label{tab:tag_exp_place}
	\begin{tabular}{c|c}
	\hline
	Method     & mAP   \\ \hline
	I3D~\cite{carreira2017quo}  &  7.66 \\
	TSN~\cite{TSN2016ECCV}   & \textbf{8.33} \\ \hline
\end{tabular}
	\end{minipage}
\end{table}

\begin{figure}[!b]
	\centering
	\includegraphics[width=\linewidth]{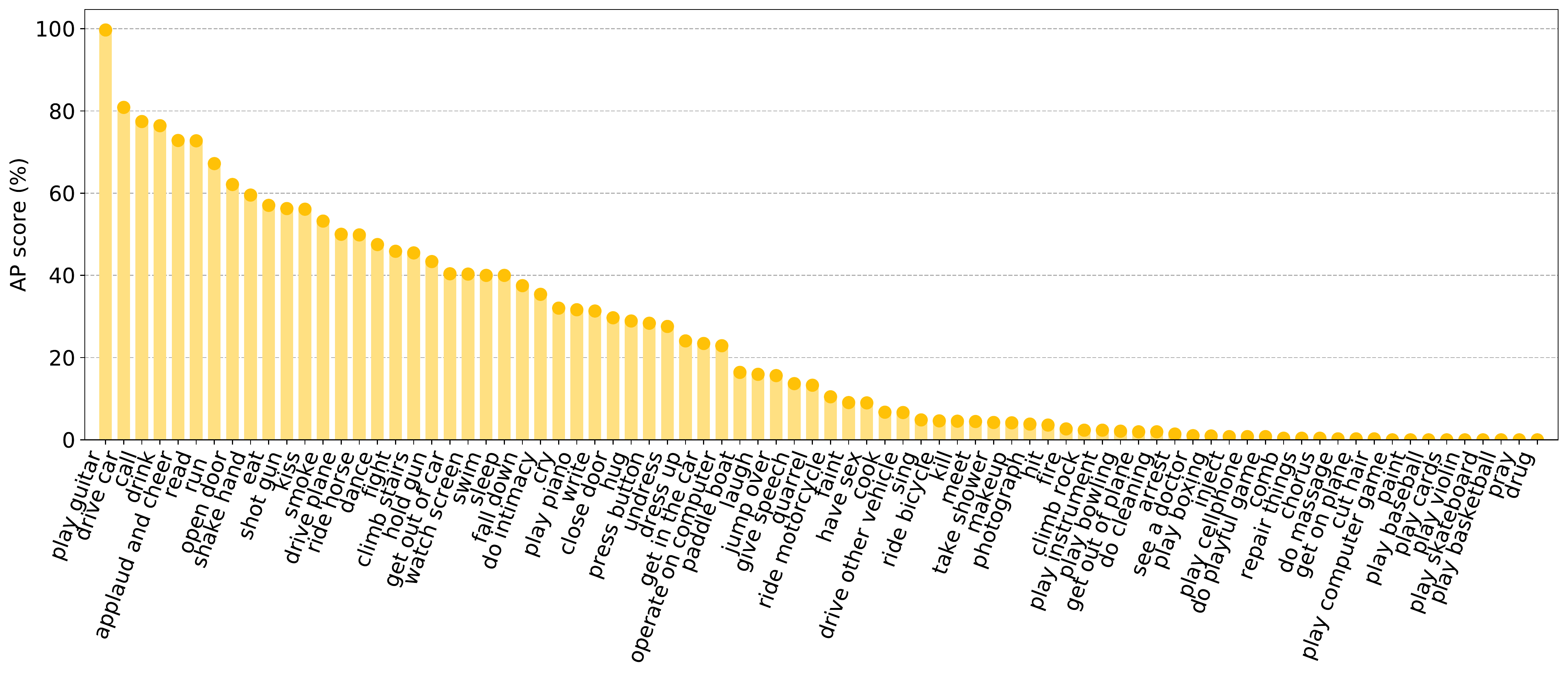}
	\caption{Per-class AP score of SlowFast Network for action recognition task on MovieNet (sorted according to
		the performance of each class).
	}
	\label{fig:action_ap}
\end{figure}

\noindent\textbf{Analysis.}
The experimental results are shown in Tab.~\ref{tab:tag_exp_action}.
We see that the performance of TSN is the lowest while SlowFast
is the best and outperform other baselines by a large margin.
To further analysis the performance, we show per-class
AP score of SlowFast Network in Fig.~\ref{fig:action_ap}.

\subsection{Place Recognition}
\label{subsec:exp_place}
We conduct experiments of place recognition on MovieNet place recognition benchmark. The task aims at predicting multiple place
tags of a given video.

\noindent\textbf{Dataset Split.}
The whole dataset is randomly split as train, val, test set
with  $8101$, $2624$, $2975$ clips respectively, without overlapping movies.

\noindent\textbf{Loss and Metrics.}
For training all models, we use binary cross-entropy loss
as the loss function.
For multi-label classification evaluation, we use
mAP as the evaluation protocol.

\noindent\textbf{Implementation of TSN.}
We adopt the same TSN structure as in MovieNet-Action,
except that we use $12$ segments here instead of $3$.
The training scheme is also the same as TSN in MovieNet-Action
except that the batch size is changed to $8$.

\noindent\textbf{Implementation of I3D.}
We use the same I3D model and the same training scheme of
I3D in MovieNet-Action.

\begin{figure}[t]
	\centering
	\includegraphics[width=\linewidth]{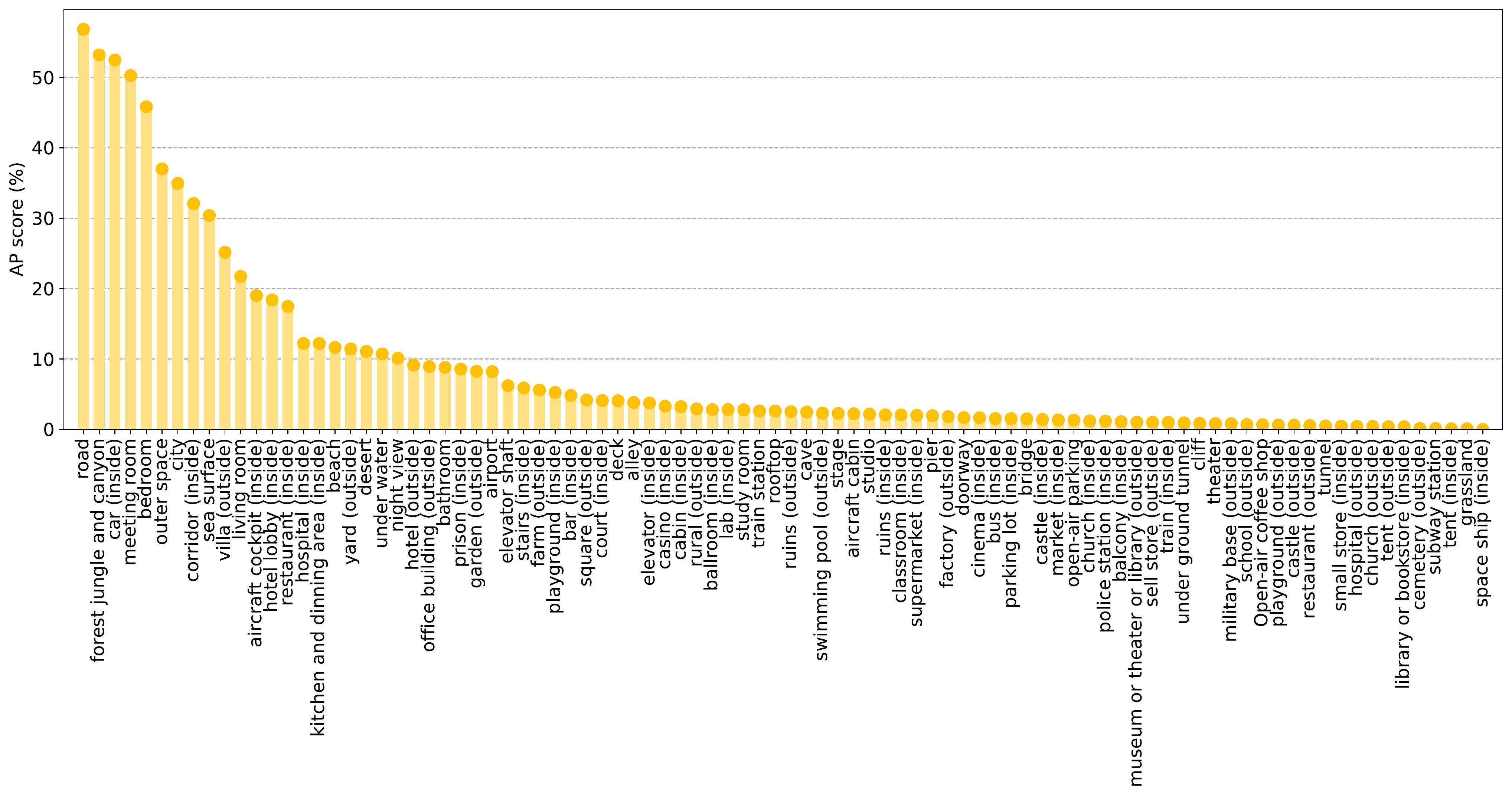}
	\caption{Per-class AP score of TSN for place recognition task on MovieNet (sorted according to
		the performance of each class).
	}
	\label{fig:place_ap}
\end{figure}

\noindent\textbf{Analysis.}
The experimental results are shown in Tab.~\ref{tab:tag_exp_place}.
The backbone weights are adopted from ImageNet pretrained model.
We see that the performance of TSN outperforms
I3D probably because 3d convolution is harder to learn
and the I3D model suffers over-fitting.
We do not use SlowFast Network as one of the baseline models
because SlowFast can not leverage the pretrain weight from ImageNet and our dataset is not large enough to support
training SlowFast from scratch.
To further analysis the performance, we show per-class
ap score of TSN Network in Fig.~\ref{fig:place_ap}.

\subsection{Story Understanding}
\label{subsec:exp_msr}
%
We conduct experiments of movie-synopsis retrieval
on MovieNet synopsis alignment dataset.
To be specific, the task is to search a relevant movie
segment given a synopsis paragraph as query.
The extended results are show in Tab.~\ref{tab:msa_exp_supp}.
and the details would be introduced below.

\begin{table}[t]
	\centering
	\caption{\small Results of movie segment retrieval. Here, G stands for global appearance feature, S for subtitle feature, A for action, P  for character and C for cinematic style.}
	\label{tab:msa_exp_supp}
	\begin{tabular}{l|cccc}
		\hline
		Method             & Recall@1   & Recall@5   & Recall@10  & MedR \\ \hline
		Random             & 0.11  & 0.54  & 1.09  & 460  \\
		G             & 3.16 & 11.43 & 18.72 & 66   \\
		G+S             & 3.37  & 13.17 & 22.74 & 56   \\
				G+P         & 12.76 & 42.98 & 53.97 & 8 \\
		G+S+A      & 5.22  & 13.28 & 20.35 & 52   \\
				S+A+P     & 12.62 & 27.21 & 49.40 & 11 \\
		G+S+A+P & 18.50 & 43.96 & 55.50 & 7    \\
		G+S+A+P+C & 18.72 & 44.94 & 56.37 & 7    \\ 
		MovieSynAssociation~\cite{Xiong_2019_ICCV} & \textbf{21.98} & \textbf{51.03} & \textbf{63.00} & \textbf{5}    \\ \hline
	\end{tabular}
\end{table}

%
\noindent\textbf{Dataset Split.}
After dataset split by movies, we obtain train, val, test set
with $2422$, $867$, $919$ samples respectively.

\noindent\textbf{Evaluation Metrics.}
For retrieval task, we adopt two metric to measure the performance, namely, Recall@K and MedR.
(1) Recall@K: the fraction of ground truth movie segments that have been ranked in top K;
(2) MedR: the median rank of ground truth movie segments.

\noindent\textbf{Baseline Models.}
We adopt VSE~\cite{frome2013devise} as the base model in our experiments.
In VSE model, we gradually add four kinds of nodes
into baseline model, namely, \emph{appearance}, \emph{subtitle}, \emph{action} and \emph{character}.
For each node, we extract a sequence of visual features from
movie segment and a sequence of text features from synopsis
paragraph. The detail of feature extraction
will be introduced in the next section.
In VSE, the input video features and
paragraph features are first transformed with two-layer MLPs.
For appearance, subtitle and action nodes, we
first obtain the embedding of segment and paragraph by taking the average of the output sequence features.
Then, the similarity score between segment and paragraph
is computed by applying cosine similarity between two embeddings.
For cast feature, we obtain the similarity score by applying \emph{Kuhn–Munkres (KM)} algorithm~\cite{kuhn1955hungarian}
between the output sequence features from segment and paragraph.
For training, we use the pairwise ranking loss with margin
$\alpha$ shown below,
\begin{multline}
\cL(S; \vtheta) = \sum_{i}\sum_{j\neq i} max(0, S(Q_j, P_i) - S(Q_i, P_i) + \alpha)\\
+ \sum_{i}\sum_{j\neq i} max(0, S(Q_i, P_j) - S(Q_i, P_i) + \alpha)
\end{multline}
where $S(Q_j, P_i)$ denotes the similarity score between $j^{th}$
segment $Q_j$ and $i^{th}$ paragraph $P_i$. $\vtheta$ denotes
the model parameters.

\noindent\textbf{Feature Extraction.}
The process of extracting input features for different modalities and different nodes are presented below.
\begin{itemize}
	\item \textbf{Appearance feature from movie segment.}
	Appearance feature consists of a sequence of features extracted
	from each shot.
	For each shot, we extract the feature from \emph{pool5} layer of ResNet-101~\cite{he2016deep}.

	\item \textbf{Appearance features from synopsis paragraph.}
	For paragraph, the appearance feature is represented as
	a sequence of Word2Vec~\cite{mikolov2013efficient} embeddings
	extracted from each sentence.

	\item \textbf{Subtitle feature from movie segment.} We also use Word2Vec embedding to extract features for subtitles in each shot.
	When adding to the model, we directly concatenate the subtitle
	feature of each shot to the appearance feature of each shot.

	\item \textbf{Action feature from movie segment.} The
	action features come from feature concatenated by TSN~\cite{TSN2016ECCV} pre-trained
	on AVA~\cite{gu2018ava} and on MovieNet action recognition benchmark. We extract the
	action feature on each shot when there are actors appear in
	this shot.

	\item \textbf{Action feature from synopsis paragraph.}
	We detect verbs using part-of-speech tagging provided by  GoogleNLP\footnote{https://cloud.google.com/natural-language/}. We select 1000 verbs with the highest frequency from the synopses corpus, and then retain those corresponding to visually observable actions, \eg \emph{run}.
	Action verbs are then represented by Word2Vec
	embeddings.

	\item \textbf{Character feature from movie segment.}
	We leverage the detector trained on MovieNet character detection benchmark to detect
	character instance in every shot.
	Then we use ResNet50 pretrained on PIPA~\cite{zhang2015beyond} to extract
	the body feature and face feature as representation.

	\item \textbf{Character feature from synopsis paragraph.}
	We detect all the named entities (\eg \emph{Jack}) using StanfordNer~\cite{finkel2005incorporating}. With the help of IMDb, we can
	retrieve a portrait for each named character and thus obtain
	facial and body features using ResNet50 pre-trained on
	PIPA. This allows character nodes to be matched to
	the character instances detected in the movie.

\end{itemize}

\noindent\textbf{Add Cinematic Style.}
%
As mentioned in the paper, cinematic style can help distinguish
whether a node is important in a particular shot.
Hence we design a module that take cinematic style and
the node itself as input and produce an attention on this
node.
For each element in a shot, we concatenate its feature and the probability of the cinematic style as input, then this feature
is passed through a MLP to produce a single attention score.
This score is later used as the weight of output embedding.

\noindent\textbf{Using Graph Formulation.}
We implement the algorithms in~\cite{Xiong_2019_ICCV}, both Event Flow Module and Character Interaction Module.
The results are the combination of the two modules. That being said, we leverage the two modules to model
spatial and temporal graph relations respectively. The difference between our implementation with theirs are the
feature we used are different (see the Feature Extraction section).

\noindent\textbf{Implementation Details.}
We train all the embedding networks using SGD with learning rate $0.001$. The batch size is set to $16$ and
the margin $\alpha$ in pair-wise ranking loss is set to $0.2$.

\setcounter{table}{0} 
\setcounter{figure}{0} 

\section{Toolbox}
\label{sec:tool}
In this section, we introduce the toolbox designed for MovieNet, it will
be released with the dataset and corresponding benchmark codes.
The toolbox are mainly comprised of the following parts:
\begin{itemize}
	\item \textbf{Crawlers.}
	The crawler for downloading metadata, subtitle and other useful data will be provided.
	\item \textbf{Preprocessing.}
	The preprocessing tools would provide functions that efficiently process multi-media resources.
	For example, extract audio waves, cut movies, resize movies.
	\item \textbf{Data generators.}
	The tools for generating the data, for example, shot detection will be included.
	Besides, we will also provide hany tool for users to align their own movies with ours, if needed.  
	\item \textbf{Data Parser.}
	The parsers are designed to easily access the required matadata or data.
	For example, to visualize the character bounding box or to read the genres of a movie.
\end{itemize}

%
%
\bibliographystyle{splncs04}
\bibliography{arxivbib}
\end{document}